\pdfoutput=1

\documentclass[11pt]{article}

\usepackage[preprint]{acl}

\usepackage{newtx}
\usepackage{latexsym}
\usepackage{graphicx}
\usepackage{hyperref}
\usepackage{listings}
\usepackage[T1]{fontenc}

\usepackage[utf8]{inputenc}

\usepackage{microtype}

\usepackage{inconsolata}

\usepackage{amsmath}
\usepackage{enumitem}
\usepackage{parallel}
\usepackage{subcaption}
\usepackage{algorithm, algpseudocode}
\newtheorem{theorem}{Theorem}[section]

\definecolor{codegreen}{rgb}{0,0.6,0}
\definecolor{codegray}{rgb}{0.5,0.5,0.5}
\definecolor{codepurple}{rgb}{0.58,0,0.82}
\definecolor{backcolour}{rgb}{0.95,0.95,0.92}
\lstdefinestyle{mystyle}{
    breakindent=0em,
    backgroundcolor=\color{backcolour},   
    commentstyle=\color{codegreen},
    keywordstyle=\color{magenta},
    numberstyle=\tiny\color{codegray},
    stringstyle=\color{codepurple},
    basicstyle=\ttfamily\scriptsize,
    breakatwhitespace=false,         
    breaklines=true,                 
    keepspaces=true,                 
    numbersep=5pt,                  
    showspaces=false,                
    showstringspaces=false,
    showtabs=false,                  
    tabsize=2,
    columns=fullflexible,
}

\setitemize{noitemsep,topsep=0pt,parsep=0pt,partopsep=1pt,leftmargin=6mm}
\newcommand{\comment}[1]{\textcolor{blue}{#1}}

\newcommand{\dontshow}[1]{}

\title{On the Empirical Complexity of Reasoning and Planning in LLMs}

\author{Liwei~Kang\thanks{Equal contribution, listed in alphabetical order.} \quad Zirui~Zhao\footnotemark[1] \quad David~Hsu \quad Wee~Sun~Lee \\
National University of Singapore\\
\texttt{\{kang,~ziruiz,~dyhsu,~leews\}@comp.nus.edu.sg}}

\begin{document}
\maketitle

\begin{abstract}
Chain-of-thought (CoT), tree-of-thought (ToT), and related techniques work surprisingly well in practice for some complex reasoning tasks with Large Language Models (LLMs), but why? This work seeks the underlying reasons by conducting experimental case studies and linking the performance benefits to well-established sample and computational complexity principles in machine learning. We experimented with 6 reasoning tasks, ranging from grade school math, air travel planning, \ldots, to Blocksworld. The results suggest that (i) both CoT and ToT benefit significantly from task decomposition, which breaks a complex reasoning task into a sequence of steps with low sample complexity and explicitly outlines the reasoning structure, and (ii) for computationally hard reasoning tasks, the more sophisticated tree structure of ToT outperforms the linear structure of CoT. These findings provide useful guidelines for the use of LLM in solving reasoning tasks in practice. 

\end{abstract}
\section{Introduction}
\dontshow{
\comment{
\begin{itemize}
    \item The empirical complexity can be measured by sample complexity and computational complexity
    \begin{itemize}
        \item The sample complexity can be measured by description length. LLM method with long DL is empirically worse in performance than shorter ones. [Travel, 24]
        \item LLM works well in solving problems with linear time complexity. [GSM, DP]
        \item Regarding hard problems (E.g., NP-complete), LLM cannot solve them well in polynomial time. External compositional reasoning methods such as tree search can boost the performance of LLM when designed properly. [24]
        \item Using LLM to linearise the tree search process would suffer from compounding errors and might not be better than CoT. [Travel]
    \end{itemize}
\end{itemize}
}}



Reasoning and planning tasks are often challenging due to their inherently multi-step processes. Compared with the \textbf{Direct} approach that prompts the model to provide immediate answers, large language models (LLMs) showed surprising results on reasoning problems when they were asked to explain their reasoning step-by-step through a \textbf{chain-of-thought (CoT)} \cite{cot, zerocot} before providing their answers. This was followed by improvements of the \textbf{tree-of-thought (ToT)} \cite{yao2023tree,xie2023self}, combining LLMs with a search algorithm to structure reasoning steps into a tree and selecting promising next steps by self-evaluation. Other variants include \textbf{CoT with self-consistency (CoT-SC)} \cite{cotsc}, which generates multiple CoTs and responds with the most common answer.

\begin{figure}[t]
    \centering

    \includegraphics[width=\linewidth]{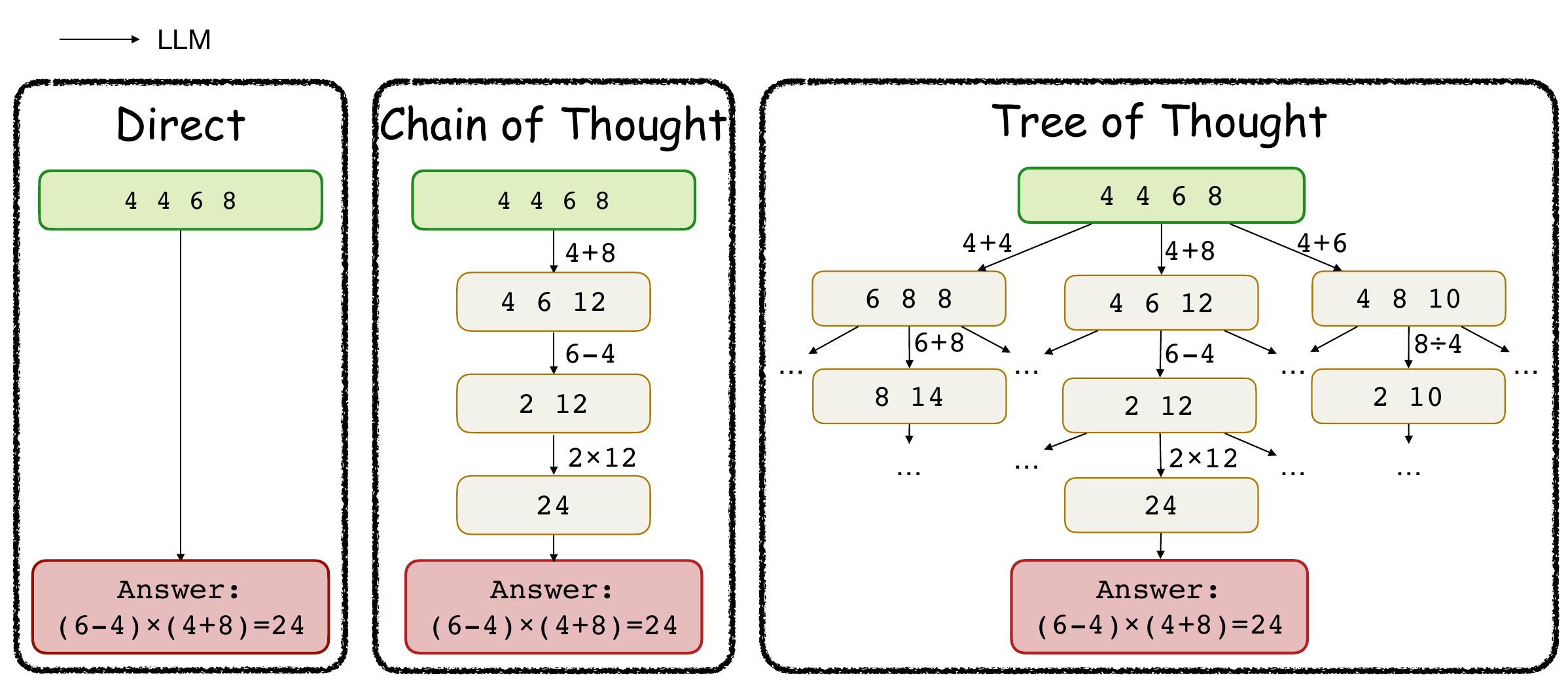}
    \caption{An illustration of LLM reasoning methods on the Game of 24. Given four poker cards, the player combines the cards using basic arithmetic operations, $(+, -, \times, \div)$, to reach the target number of 24.}
    
    
    \label{fig:examples}
\end{figure}

However, evidence shows that the relative performance of CoT, ToT, and their variants may vary from task to task. For example, CoT and its variants, such as self-consistency (CoT-SC) \cite{cotsc}, have been successful in solving grade school math problems (GSM). Our experiments show that ToT is only marginally better than CoT and even performs worse than CoT-SC on GSM when using the same token budget. However, in the Game of 24 shown in Fig \ref{fig:examples}, where four numbers need to be manipulated with arithmetic operations to obtain the number 24, using CoT to provide a solution with a short reasoning chain fails badly. In addition, CoT-SC did not show significant improvement while using more tokens at inference. In contrast, ToT significantly outperforms CoT and CoT-SC on Game of 24 \cite{yao2023tree}. These pieces of evidence raise an important question: what are the underlying factors that affect the effectiveness of different LLM reasoning methods in various tasks?


To answer this question, we investigate when and why CoT and ToT are effective in reasoning and planning problems from the viewpoint of sample complexity and computational complexity in machine learning. Sample complexity captures the amount of data needed for learning predictors, which we measure by description length \cite{shalev2014understanding}, the number of bits required to describe the learnable part of the predictor. The predictors take an input, e.g., a sequence of words, and produce a prediction, e.g. a label that may be used directly or as a component of a larger reasoning process \cite{cot}. Predictors with a small description length can be shown to require less training data, i.e. a small sample complexity \cite{shalev2014understanding}, to achieve low generalisation error. Computational complexity is relevant in two distinct ways in this paper: 1) in the amount of computation required for learning, e.g. finding the correct parameters in the predictor given the training data, and 2) in the amount of computation required for reasoning, e.g. finding the solution given a problem. Learning may become computationally intractable if the values of hidden variables are not observed during learning\footnote{Indirectly the values need to be inferred during learning.} \cite{aloise2009np,blum1988training}, motivating us to consider their presence during learning of CoT. For problems that are computationally hard to solve, e.g. NP-hard problems, it is unlikely that a small predictor producing a short CoT solution exists in the worst case, which motivates the use of more complex structures like search trees.

We focus on the \textbf{reasoning} and \textbf{planning} problems in the context of \textbf{natural language processing}. A reasoning problem entails deducing the answer to a question from provided evidence and applicable reasoning rules. It often requires applying various rules multiple times to connect different pieces of evidence and draw a conclusion. Planning, a special subset of reasoning, requires an action sequence to achieve a desired goal state from a current state. 
Planning often requires reasoning over a long time horizon, making it computationally harder to solve.

We empirically study these issues through six case studies: \textbf{grade school mathematics} \cite{cobbe2021training}, \textbf{multi-hop question answering} \cite{trivedi-etal-2022-musique}, a simple \textbf{dynamic programming (DP)} problem \cite{dziri2023faith}, \textbf{air travel planning} \cite{zhao2023large}, \textbf{Game of 24} \cite{yao2023tree} and \textbf{Blocksworld} \cite{valmeekam2023planning}. Grade school math, multi-hop question answering and DP problems we consider to have computationally efficient reasoning components. Air travel planning has two different efficient solutions that we compare. Finally, Game of 24 and Blocksworld appear to be computationally difficult.

We study the problems under different settings, including prompting general LLMs, fine-tuning with task-specific data, and in-context learning with task-specific examples. Our main findings are consistent over different settings. These findings suggest a few guiding principles for using LLM to solve reasoning and planning tasks in practice:
\begin{itemize}
    \item \emph{CoT and ToT enhances LLM reasoning when problem decomposition lowers the sample complexity.} In all six cases, decomposing problems with chain or tree structures reduces sample complexity and improves performance. 
    \item \emph{Explicitly annotating necessary information improves CoT performance.} In the DP problem and Blocksworld, explicitly demonstrating relevant variables improves learning. For tasks with small search trees like air travel planning, CoT that linearizes the search tree and explicitly describing intermediate computations outperforms CoT that directly predicts the next action.
    \item \emph{Tree structures help when generating short-chain solutions is computationally hard; CoT-SC is effective otherwise.} For tasks like Game of 24 and Blocksworld, where short-chain solutions are likely computationally hard to find, ToT works better. In tasks like GSM and multi-hop QA with efficient algorithms for finding next steps, CoT-SC performs better.
\end{itemize}

\section{Related Work}
LLMs have shown significant progress in tackling reasoning and planning problems. Initial studies \cite{cot, cotsc, zerocot, chen2022program, gao2023pal} unveiled various prompting techniques that enable LLMs to demonstrate reasoning processes step by step, thereby substantially boosting their reasoning abilities. This approach has been swiftly adapted to address everyday planning issues \cite{huang2022language, huang2022inner, ahn2022can, song2023llm, wang2023voyager, singh2023progprompt}. Subsequent research has integrated LLMs with diverse search algorithms, further enhancing their capability to solve complex reasoning and planning challenges \cite{zhang2023planning, yao2023tree, zhao2023large, xie2023self, ding2023everything, feng2023alphazero, hao2023reasoning, liu2023making}. 

Several works explore why CoT improves performance. \citet{feng2024towards} investigates how CoT enhances model capacity; \citet{prystawski2024think} and \citet{wang2024understanding} examine the training data distribution that enables CoT to excel; \citet{dziri2023faith} discusses CoT's limitations in compositional reasoning. Our work takes a different approach, we explore principles that guide LLM behavior across various reasoning frameworks, and provide insights into selecting appropriate strategies for different tasks. While \citet{zhao2023large} discusses the sample complexity of LLM planning, they overlook the computational implications.

\section{Analysis of LLM Reasoning Methods}


We aim to analyze LLM reasoning methods from the sample and computational complexity perspective in machine learning. Our goal is to understand the complexity of problems using simple representations. However, analyzing LLM learning with transformer architecture is challenging, and the effects of pre-training, which we do not control, are present throughout. Instead, we empirically observe whether our analysis reflects the practical behaviour of LLMs and whether the insights from our analysis are practically useful. Specifically, when our analysis suggests that a particular method is preferred, we examine whether it is indeed preferred empirically.

 
\subsection{Problem formulation}
 
The LLM reasoning approach, e.g., CoT and ToT, essentially uses a sequential decision approach to do the reasoning instead of other approaches, such as converting to a Boolean satisfiability problem (SAT) and solving it using SAT solvers. Thus, we formulate reasoning and planning problems using planning terminology. A planning problem can be defined using a state space $S$, an action space $A$, a transition function $T$, and a goal function $G$. 
The state space $S$ defines all possible states (e.g., the boolean values of assigned variables for logical inference or scalar values of assigned variables for math problems). The action space $A$ consists of possible equations or rules. Each state has applicable actions that can be executed. The transition function $T(s,a)$ specifies a new state $s'$ after applying action $a$ in state $s$. The goal function $G$ specifies whether the current state is a goal state.


The solution to a reasoning or planning problem is a sequence of actions in the action space that transforms the initial state into a goal state. In a CoT, we directly use the LLM as a \textbf{policy} to map the current state (as inferred by the LLM from the context) to the action, while in a ToT, the LLM is used to specify applicable actions in each state to construct a search tree. LLM is also used as a transition function in both methods.

\subsection{Decomposition and sample complexity}
 
\subsubsection{Description length (DL)}
Description length (DL), the number of bits required to describe the learnable part of the predictor, is used to analyse the sample complexity in machine learning. It is used in the minimum description length (MDL) principle, also known as Occam's Razor from the philosophy of science. The MDL principle suggests that a method with a shorter description length requires less training data and is preferred. MDL has been formalized in various ways. One formal statement (from section 7.3 of \citet{shalev2014understanding}) is provided here:
 
\begin{theorem}[Occam's Razor]
Let $\mathcal{H}$ be a hypothesis class and let $d\colon\mathcal{H} \rightarrow \{0, 1\}^*$ be a prefix-free description language for $\mathcal{H}$. Then, for every sample size, $m$, every confidence parameter, $\delta > 0$, and every probability distribution, $D$, with probability greater than $1-\delta$ over the choice of $S\sim D^m$ we have that,
$ \forall h\in\mathcal{H}, L_D(h)\leq L_S(h)+\sqrt{(|h|+\ln{(2/\delta)})/2m}$,
where $L_S(h)$ is the empirical loss of $h$ on the $S$, $L_D(h)$ is the expected loss of $h$, and $|h|$ is the length of $d(h)$.
\label{theorem:or_main}
\end{theorem}
According to Theorem \ref{theorem:or_main}, we can bound the expected loss of a solution $h$ by the description length $|h|$ and the training loss $L_S(h)$. Thus, when trained by the same amount of data, predictors with smaller DL have lower generalisation errors, indicating a lower sample complexity. 
 
\subsubsection{DL analysis of LLM reasoning methods}

For each reasoning method, we can analyze the description length of its policy and transition. 

\paragraph{Direct}
Direct answering method does not involve explicit steps of reasoning, all reasoning is performed internally in the neural network. This method may have a low sample complexity if the neural network architecture closely aligns with the reasoning algorithm \cite{xu2020can}. Analysing whether the algorithm fits the inductive bias of the network is complex, so we mostly explore a tabular representation for simplicity. In problems with $N$ variables, each taking $K$ values, direct answers require learning a table of size $K^N$, which exponentially increases with more variables. 

\paragraph{CoT}
With $|A|$ number of possible actions, each depending on $a_i$ variables, the description length for transition functions of these actions is proportional to $\sum_{i=1}^{|A|} K^{a_i}$. We also need a \textit{policy} function predicting action to select based on observations with its description length of $K^M$ if it depends on $M$ variables. If the policy depends only on whether the variables have been observed rather than their values, then a binary table of size $2^M$ is sufficient. CoT-SC runs CoT multiple times and responds with the most common answer -- this reduces the prediction variance and, consequently, the sample complexity at the cost of more computation.

\paragraph{ToT}
The complexity of transition functions in ToT is analysed similarly to CoT. Another component is the self-evaluation module, which evaluates whether the current state or selected action is promising to reach the final answer. It acts as a search heuristic: the highly promising branches will more likely be selected for the next search steps. This self-evaluation module is essentially a classifier. Similar to the policy for next-step prediction, the classifier can be described by a table with a size of $K^M$ if it depends on $M$ variables to determine the next actions. If the policy depends only on whether the variables have been observed rather than their values, then a binary table of size $2^M$ is sufficient. 

Our analysis is mostly done assuming a tabular representation. This is further complicated by two factors: the amount of pre-training of the LLMs, which we do not control, and possible failures in training, e.g. due to the presence of local minimums. Empirical observations in our case studies assess whether the transformer architecture used in LLM resembles table-filling behaviour for each case or successfully learns a better algorithm, taking into account the effects of pre-training.

\subsection{Reasoning structure and computational complexity}

The computational complexity of solving (versus learning) a reasoning or planning problem is a key factor in choosing between CoT and ToT. Some problems, e.g. NP-complete problems, have short solutions that are verifiable in polynomial time, but efficient policies to find such solutions are unlikely to exist. ToT, with its search algorithm, presents a viable approach for such hard problems by allowing the use of more computation during the search process. However, for simple problems with low computational complexity, a complex search algorithm may not produce additional improvement. 


\section{Case Studies\protect\footnote{See Appendix \ref{sec:experiment_detail} and \ref{sec:prompts} for experimental details and complete prompts.}}
\subsection{Grade School Maths}
GSM8K \cite{cobbe2021training} consists of grade school math problems in natural language. LLMs solve these problems effectively with CoT \cite{achiam2023gpt}. The dataset is widely used to evaluate the reasoning ability of language models. We analyze the dataset and find that it is well-suited for CoT, as most problems can be solved with a linear-time policy. We also empirically show that CoT is usually sufficient, and ToT offers little improvement. In a subset of 50 randomly selected problems, we found that 49 can be solved with a chain-style algorithm where each step involves selecting an equation with all but one variable known, allowing the remaining variable's value to be inferred. The remaining problem, solvable with simultaneous equations, is ignored in the rest of the study.
\subsubsection{Analysis}
\label{sec:gsm_analysis}

\paragraph{Direct}
Consider a problem with $N$ variables each can take $K$ values. A tabular representation requires a table of size $K^N$ and description length of $O(K^N\log K)$ for each question type, assuming answers also take $K$ possible values (thus $\log K$ bits to represent). The description length can be large (we manually analyze a subset of GSM8K problems and show statistics in Appendix \ref{sec:gsm_statistics}).


\paragraph{CoT}
In a math word problem, the action can be selecting the next applicable rule, and the transition would be deriving the next value of an intermediate variable based on the rule and the value of the input variables. Assuming $A$ different actions whose transition functions require $a_i$ variables, the total description length of the learnable transitions would be $O(\sum_{i=1}^{A} K^{a_i}\log K)$. This is more manageable than Direct (see Appendix \ref{sec:gsm_statistics} for details). As for the policy, we can select an equation where the values of all except one variable are known. There exists a linear time forward chaining algorithm to do that (Appendix~\ref{sec:gsm8k_fast_algo}), which translates to a relatively small policy that needs to be learned. The components of the decomposed problems are relatively simple and suggest that decomposition with CoT may be reasonable for this problem. 

\paragraph{ToT}
As discussed in CoT, there is a simple policy for deciding the next equation to solve. If the policy is learned reasonably well by the LLM, a search may give limited improvement.
\subsubsection{Experiments}
 
\begin{figure}
    \centering
    
    \begin{subfigure}[b]{0.4\linewidth}
        \includegraphics[width=\linewidth]{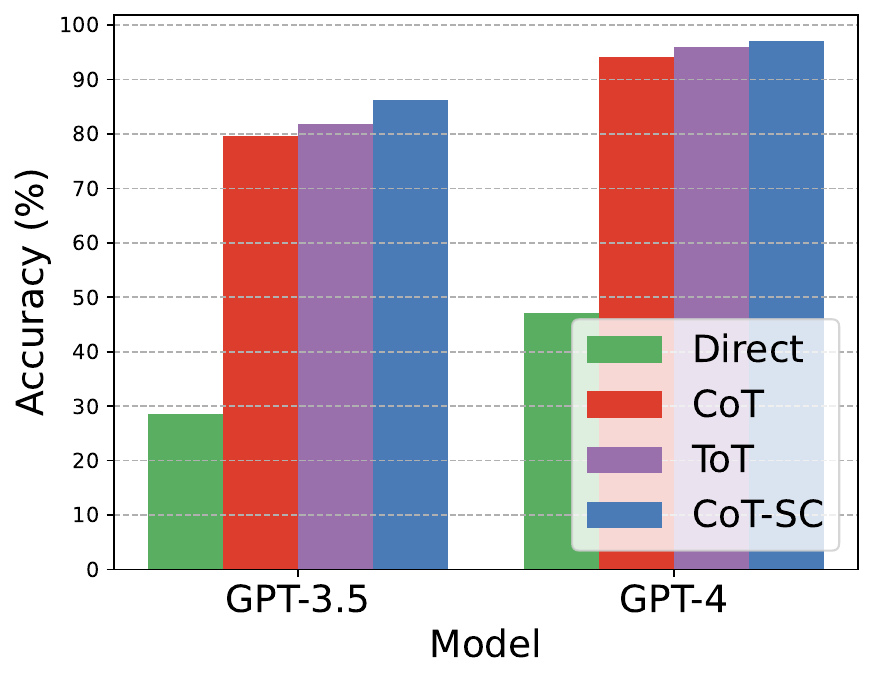}
        \caption{}
        \label{fig:sub1}
    \end{subfigure}
    \quad 
    \begin{subfigure}[b]{0.4\linewidth}
        \includegraphics[width=\linewidth]{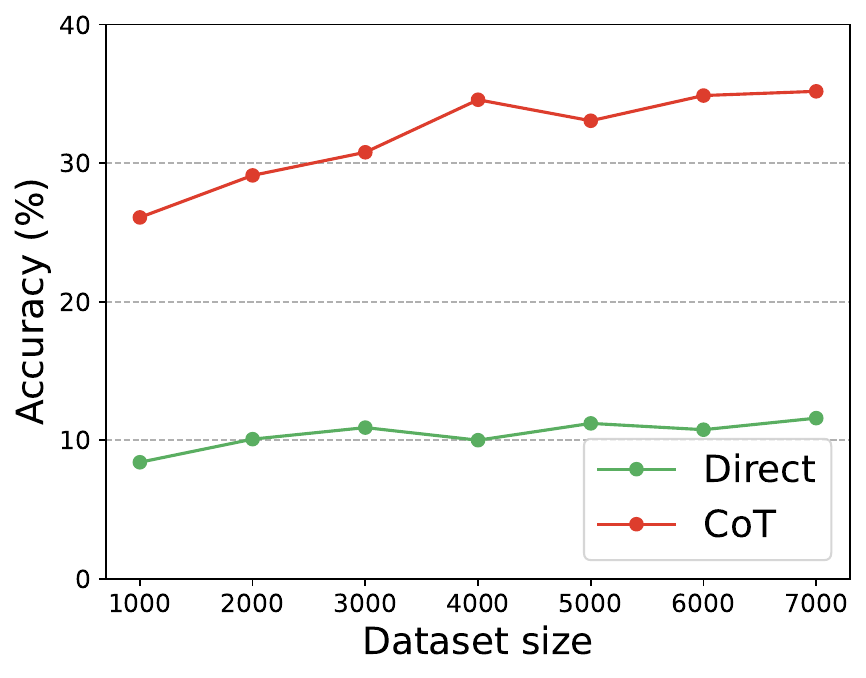} 
        \caption{}
        \label{fig:sub2}
    \end{subfigure}
    
    \vspace{8pt}
    \caption{(a) Results of GPT-3.5 and GPT-4 on GSM8K Test set; (b) Fine-tuning results on Llama2-7b}
    \label{fig:gsm8k_gpt}
\end{figure}

\paragraph{In-context Learning}
Our experimental results are shown in Fig.~\ref{fig:sub1}. We see that GPT-3.5 and GPT-4 give lower performance than CoT due to the higher sample complexity, which aligns with our analysis.

An unanalysed factor is that LLMs still need to learn to extract the equations from the question, learn the world knowledge that is not included in the question, and ground the variables' values from the previous observations. The LLMs, particularly GPT-4, do remarkably well on GSM8K, indicating that extraction and grounding may not be major hurdles for LLMs. Some errors are still present and are discussed in the Appendix \ref{sec:gsm_error}. 

For ToT, we run a beam search ToT, branching after each sentence on the choice of the next sentences suggested by the LLM. We prompt the same LLM to self-evaluate the quality of each proposed reasoning step. ToT yields similar performance to CoT, indicating the policy is reasonably well-learned. We also see that using the same token budget, CoT-SC performs better than ToT since it reduces the variance of the predictions. This suggests that when the problem does not inherently require search, spending the inference budget on CoT-SC may be more beneficial than ToT.



\paragraph{Fine-tuning}
\label{sec:word_problem_finetune}
The GPT experiments suggest that LLMs have difficulties learning to solve GSM-type questions directly. To check that, we perform fine-tuning experiments using the GSM8K training set and compare models fine-tuned with direct answer completion and CoT completion.





We fine-tune with varying sizes of subsets of the GSM8K training set and test performance on the GSM8K test set. The results are shown in Fig.~\ref{fig:sub2}. Note that each CoT example provides substantially more information than each Direct example, but CoT is substantially better even when Direct is provided with 7 times more training examples (Direct at 7k vs CoT at 1k). It suggests that the transformer in the LLM is behaving more like a tabular predictor and cannot learn to decompose the problem internally without being trained explicitly to do so.

\subsection{Multi-hop Question Answering}
We study the MusiQue dataset \cite{trivedi-etal-2022-musique}, where the task is to identify the answer given a question and a context of up to 20 paragraphs. The question may look like \textit{"Who did the hitman from The Hitman's Bodyguard play in Star Wars?"}, which can be solved by decomposing the question and answering each decomposed question to get the answer. 

\subsubsection{Analysis}
This question answering task requires both natural language processing and reasoning. We describe the reasoning process as follows: the paragraphs in the context provide a set of relation triplets, and the question can be translated to a logical expression which is a conjunction of some relation triplets, with several unknown entities in it, one of which is the target answer. The reasoning process would require finding the unknown entities in some order and to derive the final answer. The natural language processing part is common to different reasoning methods, while the complexity of reasoning can be different. 

\paragraph{Direct}
The set of relation triplets represented by the context can form a knowledge graph, with head and tail entities in the triplets being vertices and relations being edges. Consider a problem where there are $V$ vertices and $E$ edges in the context knowledge graph and question, each vertex and edge may take $K$ different values. To answer the question directly would require memorizing a table of size  $K^{(V+E)}\log K$, representing a mapping from possible configurations of the question and graph to an answer. Thus the description length of learning Direct in a tabular manner would be $O(K^{(V+E)}\log K)$. 

\paragraph{CoT}
Similar to the Grade School Maths problem, the reasoning process in this task can be seen as a sequence of actions and transitions. The action here would be to select a triplet in the question with unknown entities, and the transition is to infer the unknown entity. The unknown entity can be found by matching the known entity and relation from the question triplet to the appropriate triplet in the context. The same entity can be named differently in the question and the context triplets. A table of size $K^2$ can be learned to specify whether two entities match. Matching relations is similar. This suggests that learning the transition is manageable with reasonable complexity, although extracting the entity using natural language sentences adds additional complexity.
A policy for this problem can simply be selecting a triplet in the question that has two known element and one unknown element. As in the Grade School Maths problem, this can be done via forward chaining in time linear in the size of the knowledge graph. The small policy representation suggests that learning the policy would not have high sample complexity. 

\paragraph{ToT}
The description length of the transition functions of ToT is the same as CoT and as in the CoT case, performance of the transition likely depends more on the natural language processing component. 
In addition, all single hop questions in the dataset have one unique answer. Hence, there is no real need for branching if the transition is well learned, and search may offer little improvement.

\subsubsection{Experiments}
\begin{figure}
    \centering
    
    \begin{subfigure}[b]{0.45\linewidth}
        \includegraphics[width=\linewidth]{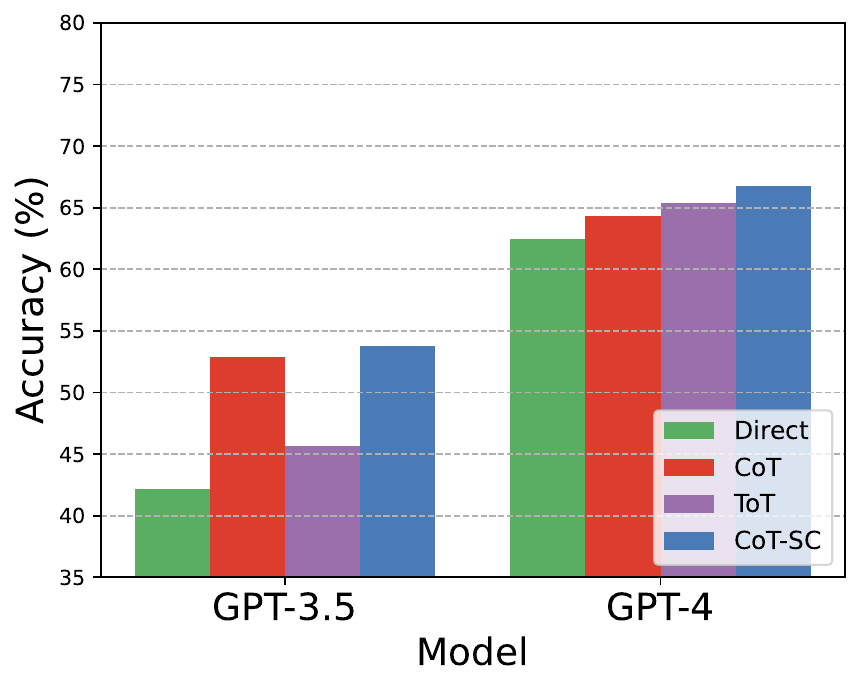}
        \caption{}
        \label{fig:musique_sub1}
    \end{subfigure}
    \quad 
    \begin{subfigure}[b]{0.45\linewidth}
        \includegraphics[width=\linewidth]{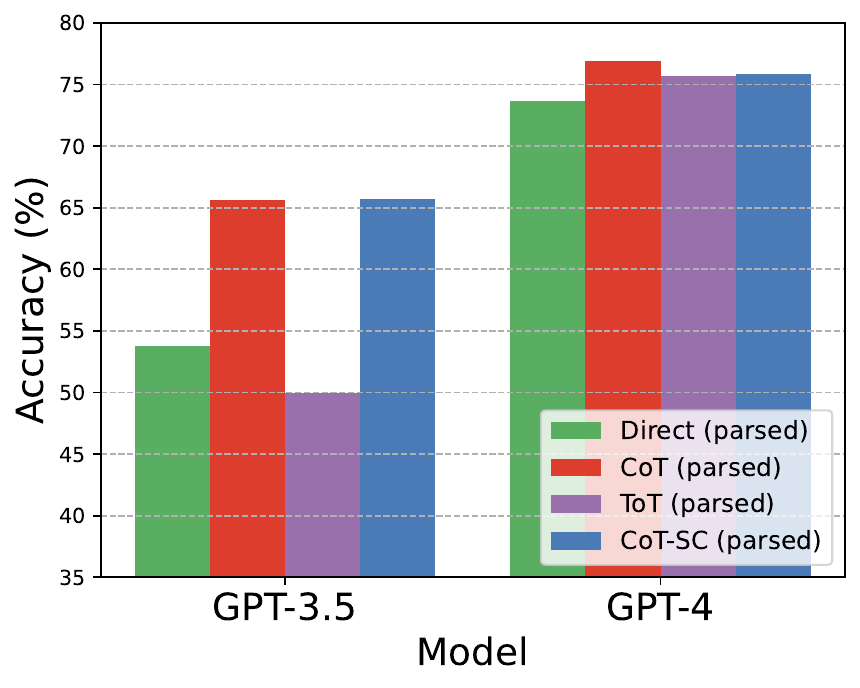} 
        \caption{}
        \label{fig:musique_sub2}
    \end{subfigure}
    
    \vspace{8pt}
    \caption{F1 score of GPT-3.5 and GPT-4 on MusiQue Dev set. (a) is using natural language context, (b) is using LLM parsed relation triplets as context.}
    \label{fig:musique}
\end{figure}

We are not able to fit each question which is paired with 20 paragraphs into the context window of Llama-2 for fine-tuning. Instead, we perform in-context learning studies with GPT models. \dontshow{\comment{Liwei, is the next sentence correct?} Yes.} We use 6-shot demonstrations consisting of the question and answer (which includes the reasoning steps for CoT) but did not include the contexts of the 6 questions in the demonstrations as the length would be too long. Results are shown in Figure \ref{fig:musique}. The experimental results align with our analysis. We make the following observations: 1) CoT consistently outperforms Direct but the gap between them is smaller when using GPT4. One possible explanation is that, the policy for reasoning, i.e. selecting the next triplet to infer its unknown entity, is relatively simple, and the more powerful GPT-4 may have been able to learn this directly during training. Extracting the correct entities from the context becomes a common bottleneck for the two methods. 2) ToT performs similarly to CoT when using GPT-4. This aligns with our analysis that ToT may give little improvement as there is no branching in the reasoning trajectories. Surprisingly, ToT is worse than CoT when using GPT-3.5. When prompted with an incomplete reasoning trajectory (i.e. some sub-questions are answered, some not asked yet), the GPT-3.5 model tends to give a final answer to the original question instead of generating the next sub-question\footnote{We further explored this issue with the following experiment: instead of generating the response all the way to the end (standard CoT), stop the generation when a newline is generated. And then use the original input together with previous generated text as input to prompt the model again to get a new line. This is equivalent to the CoT, only that we are stopping every line. This method for GPT-3.5 has f1 score 41.09 indicating that GPT3.5 is not behaving as expected.}.
3) CoT-SC is slightly better than ToT. While ToT does not improve over CoT, CoT-SC can still bring some improvement through variance reduction.



\paragraph{Comparison with GSM8K}
From our analysis, we see that math word problem (GSM8K) and multi-hop question answering (MusiQue) have simple policies. However, the in-context learning performance of GPTs on GSM8K is substantially higher than that of MusiQue. We believe there are two main reasons: 1) Retrieving the appropriate information from the context to answer a sub-question, i.e. the transition, is hard in MusiQue. After manual analysis of 20 failed examples, we found that 16 of them are due to retrieval mistake, where the correct sub-question is proposed, i.e. the policy is easy, but the proposed sub-answer is incorrect. To validate this hypothesis, we did an ablation study by replacing every natural language sentence in the context with a relation triplet automatically parsed by GPT-4 with appropriate prompting. We believe this should reduce the difficulty of retrieving information. Using the parsed context improved performance substantially as shown in Figure \ref{fig:musique_sub2}. Interestingly, CoT-SC no longer improves on CoT -- the variance in the problem may mostly be coming from the difficulty of retrieval and is probably substantially reduced in the parsed version. 2) Understanding the complex natural language question in MusiQue is hard. Some of the complex composed question in the dataset look like "How were the same people who the Somali Muslim Ajuran Empire declared independence from expelled from the natural boundary between Thailand and the country where Nam Theun is found?", which can be hard for the language model to understand. From our analysis, we see that it is easy to determine which incomplete relation triplet to infer next, however, understanding the natural language and translate it into a logical expression which is a conjunction of relation triplets can be difficult. In the 20 failed examples we analyzed, 4 of them are due to wrong decomposition, i.e. the sub-question proposed is not helpful in answering the composed question. 

ToT is not useful in this dataset likely because the sub-questions mostly have a single correct answer. Unlike GSM8K where each equation have only one correct answer, it is possible for sub-questions to have multiple acceptable answers that need to be refined with additional sub-questions. For datasets where this is common, ToT may be more effective. And if the knowledge graph representing the context is small, it may also be possible to linearize the search tree into a CoT of reasonable size, as demonstrated in the Air Travel Planning case study.


\subsection{Dynamic Programming}
We study another problem, the Maximum Weighted Independent Set problem (MWIS) \cite{KleinbergTardos2005}: \textit{Given a sequence of integers, find a subsequence with maximum sum such that no two elements in the subsequence are adjacent in the original sequence}. The problem can be solved in linear time using dynamic programming (see Appendix~\ref{sec:dp_algo}). MWIS was studied in \citet{dziri2023faith}, showing that LLMs trained on short sequences generalize poorly to longer sequences. In this paper, we focus on the amount of annotation provided in learning where only the answer is provided in Direct, whereas different levels of explicitness in annotation can be provided in CoT.

\subsubsection{Analysis}
\paragraph{Direct}
Consider a sequence with $N$ integers; each may take $K$ values. A tabular representation would have $K^N$ entries, where each entry needs $N$ bits to indicate the presence of the $N$ number in the subsequence, giving a description length of $O(NK^N)$.

\paragraph{CoT}
Using concepts in planning, we can see CoT as a combination of selecting which function to calculate next (policy), and derive the intermediate results based on the function and input variables (transition). There are a constant number of unique function that may take up to 3 variables (see Appendix~\ref{sec:mwis_prompts} for examples), so the description length of the transition would be $O(K^3\log K)$, which represents a mapping from input variables to the output value ($\log K$ bits to represent). To decide which function to apply next, we only need to know what variables have been calculated and what have not, so the policy can be represented by a table of size $O(N)$. Overall, the description length of CoT would be $O(K^3\log K+N)$, which appears manageable.

\subsubsection{Experiments}
\paragraph{In-context Learning}
In this section, we will compare prompting LLMs to answer the MWIS problem directly with prompting them to answer using CoT. We will also study two versions of CoT demonstrations and show that a more explicit demonstration can improve performance substantially.

Consider the following line from the CoT demonstration (see \ref{sec:implicit_vs_explicit} for the entire demonstration):\\
\textit{Implicit prompt} (from \citet{dziri2023faith}): \textit{... Since dp[0] != input[0] + dp[2] (6 != -4 + 5) ...}\\
\indent We can make it more explicit as follows:\\
\textit{Explicit prompt}: \textit{... Since dp[0]=6, input[0]=-4, dp[2]=5, input[0] + dp[2] = 1 != 6 = dp[0] ...}

Both prompts demonstrate steps to use DP to solve the MWIS problem, but in the Implicit prompt, when autoregressively generating the token "\textit{!=}", the values of \textit{dp[0]}, \textit{input[0]}, \textit{dp[2]}, and \textit{input[0]+dp[2]} are not explicitly stated in the immediate context and need to be inferred from all previous observations. 

\begin{figure}[htbp]
    \centering
    \includegraphics[width=0.81\linewidth]{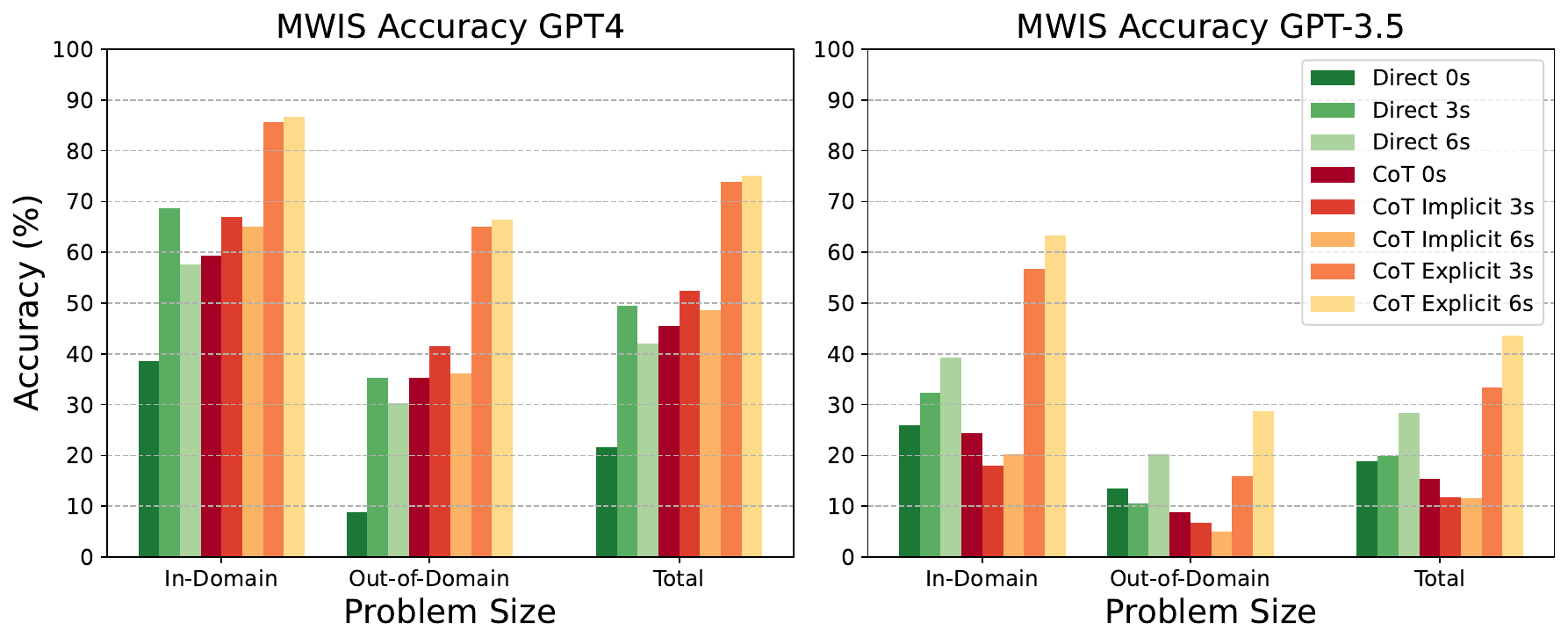}
    \caption{In-context learning results on MWIS. 3-shot prompts have one example each for sizes 4, 5, and 6, while 6-shot prompts have two examples each. "In-domain" refers sizes 4, 5, and 6, and "Out-of-Domain" refers to sizes from 6 to 10.}
    \label{fig:dp_incontext}
\end{figure}

As shown in Fig.~\ref{fig:dp_incontext}, making the demonstrations explicit provides more than 20\% improvement compared to the implicit demonstrations from \citet{dziri2023faith}. This is consistent with the learning problem becoming computationally easier if all relevant variables are explicit during learning. The sample complexity may also be smaller, as explicit demonstrations decompose the single reasoning steps into multiple simpler steps, effectively creating a small CoT. In contrast, we can view deciding between "\textit{!=}" and "\textit{==}" in the implicit demonstration as a function of all the previously observed variables. The tabular representation of such a function has a large description length which suggests that it would require a larger sample complexity to learn.

We observe that prompting LLM to directly give an answer performs comparably to the implicit CoT method (Fig.~\ref{fig:dp_incontext}). This suggests that while we prompt the LLM to "directly" give an answer, the underlying transformer model is not necessarily learning it by populating a table of size $K^N$ as it is unlikely to encounter a very large number of examples of the MWIS problem during pre-training. This suggests that the transformer used in the LLM may align well with the reasoning algorithm used here. We explore this further in fine-tuning experiments.

\paragraph{Fine-tuning}
We perform fine-tuning experiments to study both in- and out-of-domain performance.

To examine the generalizability of the fine-tuned model to OOD examples, we define two types of Domain: 
1) \textit{Problem size}: Fine-tune with problems of sizes 4, 5, and 6. Test with problems of size ranging from 4 to 10. All numbers in the input array are uniformly sampled from $[-100, 100]$ 2) \textit{Number range}: Fine-tune and test with problems of sizes 4, 5, and 6. For fine-tuning data, numbers in the input array are uniformly sampled from $[-100, 100]$, while for OOD test examples, numbers are uniformly sampled from $[-1000, 1000]$.


\begin{figure}[htbp]
    \centering
    \includegraphics[width=0.81\linewidth]{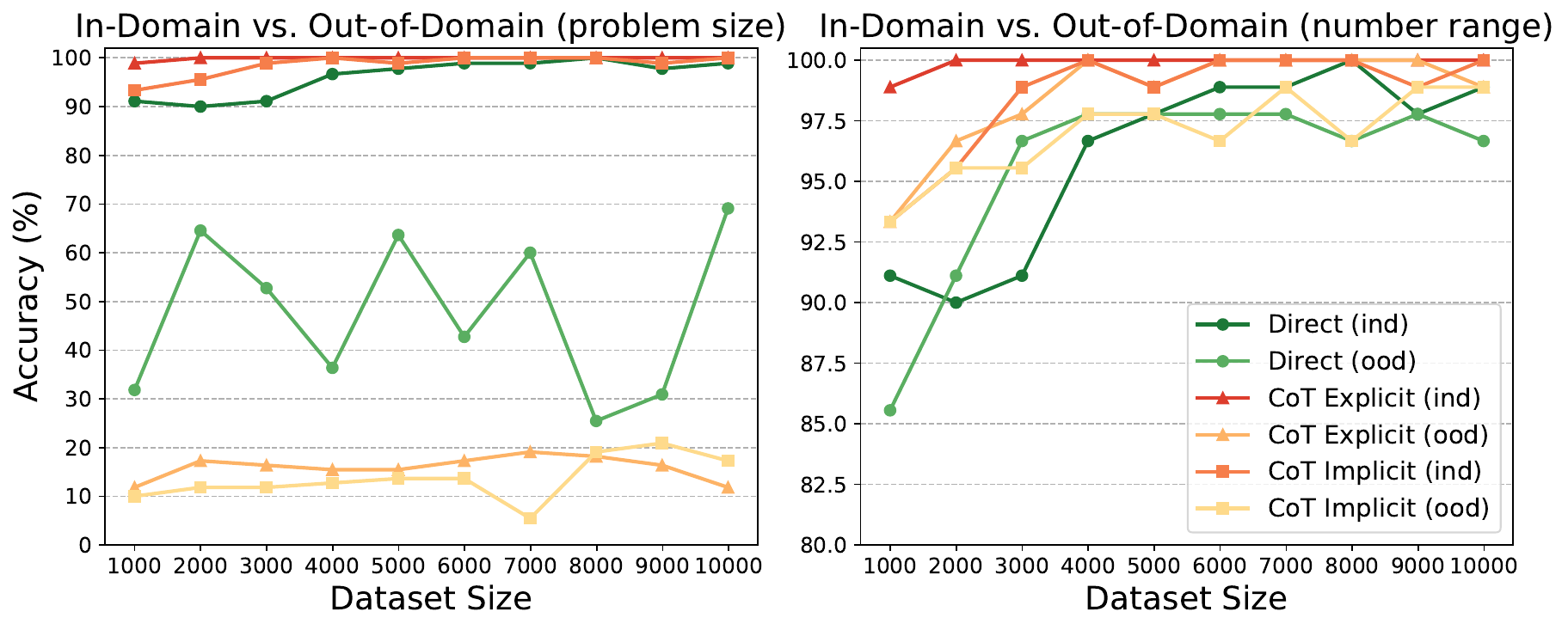}
    \caption{Results of fine-tuning Llama2-7B-chat on MWIS.}
    \label{fig:dp_finetune}
\end{figure}

\paragraph{Results}
For in-domain test examples, CoT Explicit outperforms CoT Implicit and Direct with the same number of training examples. Interestingly, with more fine-tuning data, Direct can achieve performance similar to CoT Explicit. This contrasts with math word problems, where Direct is not comparable to CoT even with 10 times more fine-tuning data. Training the transformer to directly approximate a DP algorithm seems easier than to compute the result of a multivariate equation in math word problems. However, it is unclear if this difficulty is due to computational complexity in learning or poor alignment of the transformer with solving the equation; see Appendix \ref{sec:inductive_bias} for more discussion.


As shown in \citet{dziri2023faith}, CoT struggles to generalize to reasoning lengths longer than the training data, performing worse than Direct, likely due to LLMs relying on pattern matching rather than compositional learning \cite{dziri2023faith, kharitonov2020they}. However, all methods generalize well to different ranges of numbers, possibly because the solution structure remains the same, making pattern matching less of an issue.

\subsection{Air Travel Planning}

Consider the problem of planning an air trip: \emph{given the starting city and destination, provide the flight route using direct flights between cities.} For example, \emph{What is the flight route from Singapore to New Orleans?} A valid answer might be \emph{Singapore-San Francisco-Houston-New Orleans}. This is a typical graph search problem where nodes are cities and edges are direct flights, and we aim to find a valid path that connects the start and target cities. To solve it, we can either use an LLM to predict the route directly or leverage the LLM's knowledge of the flight graph for a graph search. \citet{zhao2023large} studied this using prompting. In this paper, we go further and linearize the graph search algorithm into a CoT to study fine-tuning and learning of the graph search algorithm.

\subsubsection{Analysis}
Assume there are $N$ cities. We randomly select two cities as the current and target cities. We repeat the description length analysis in \citet{zhao2023large}, then extend it to a linearized ToT.

\paragraph{Direct \& CoT}
Generating the path directly is essentially the same as CoT, as we generate the cities on the path autoregressively. A concise representation of it is a table, where rows and columns are the current and goal cities, and each entry records the next city to fly to reach the goal. This table has $N^2$ entries, each taking $\log N$ bits to describe, resulting in a description length of $O(N^2\log N)$ bits.

\paragraph{ToT}
In ToT, LLM acts as the graph, i.e., predicts the direct flight from the current city, together with a hand-coded breadth-first search (BFS) algorithm to find the valid route. 
Assuming that the total number of edges grows proportionally to the number of cities, describing a sparse graph with $N$ nodes takes approximately $O(N\log N)$ bits, with $\log N$ bits to describe each city in the adjacency list. The graph describes the transition functions; thus, ToT can be described using $O(N\log N)$ bits if the other components are hand-coded. We can linearize the BFS algorithm into a CoT which is entirely generated by the LLM. Other than providing the adjacent cities to each city, the components include being maintaining a first-in-first-out queue, checking whether a city has been visited and recognizing the goal city. For a sparse graph as described, the runtime of BFS is $O(N)$, which translates to the existence of relatively small predictors for all the functions.


\subsubsection{Experiments}
\dontshow{
\comment{
\begin{itemize}
    \item In-context learning: GPT-3.5, GPT-4, Llama2-7b
    \begin{itemize}
        \item Compare CoT, ToT (BFS), Linearized BFS, Direct
        \item Error analysis, compounding errors in BFS and linearized BFS
    \end{itemize}
    \item Fine-tuning: Llama2-7b
    \begin{itemize}
        \item Compare the CoT, ToT (BFS), Direct, Linearized BFS
        \item Error analysis, compounding errors
    \end{itemize}
\end{itemize}
}
}


Since Direct and CoT are essentially the same, we compare CoT with ToT experimentally. For ToT, the LLM is used only in the expansion step of BFS, when it is queried to generate the neighbour of a city. In addition, we linearize the ToT process into a CoT by generating all the intermediate steps in the BFS computation in \textbf{ToT-linear}. 

We evaluate the settings of travelling between large cities (with a population of more than 5 million) and mid-sized cities (with a population between 1 million and 5 million). Details of the dataset can be found in Appendix \ref{sec:air_travel_planning_appendix}
.
\begin{figure}[htbp]
    \centering
    \includegraphics[width=0.81\linewidth]{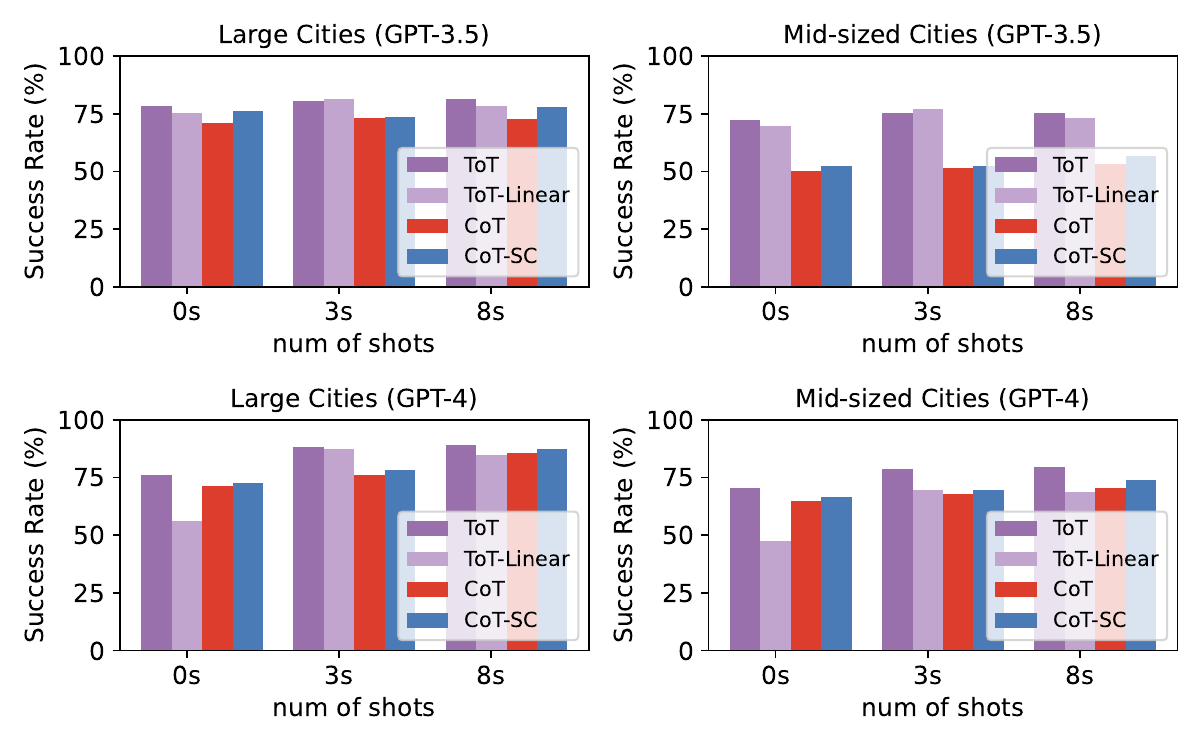}
    \caption{Results of GPT-3.5 and GPT-4 in air travel planning.}
    \vspace{0pt}
    \label{fig:travel-gpt}
\end{figure}

\paragraph{In-context learning} 
The result for in-context learning is shown in Fig.~\ref{fig:travel-gpt}. For GPT-3.5, ToT outperforms CoT slightly in large cities and substantially in mid-sized cities. This is consistent with the analysis where the description length of CoT and ToT are $O(N^2\log N)$ and $O(N\log N)$ respectively: the gap between CoT and ToT would be larger when $N$ is larger. Surprisingly, ToT-linear is comparable to ToT, even for zero-shot, where the steps in the BFS algorithm are briefly described in the prompt without any examples of its execution, indicating that there is some pre-training of the BFS algorithm in GPT-3.5. GPT-4 generally does better than GPT-3.5 for ToT and CoT, possibly because it has been trained with more data. Interestingly, GPT-4 does not do so well for ToT-linear, particularly for zero-shot, indicating that its pre-training for the BFS algorithm is possibly poorer than GPT-3.5.

\begin{figure}[htbp]
    \centering
    \includegraphics[width=0.75\linewidth]{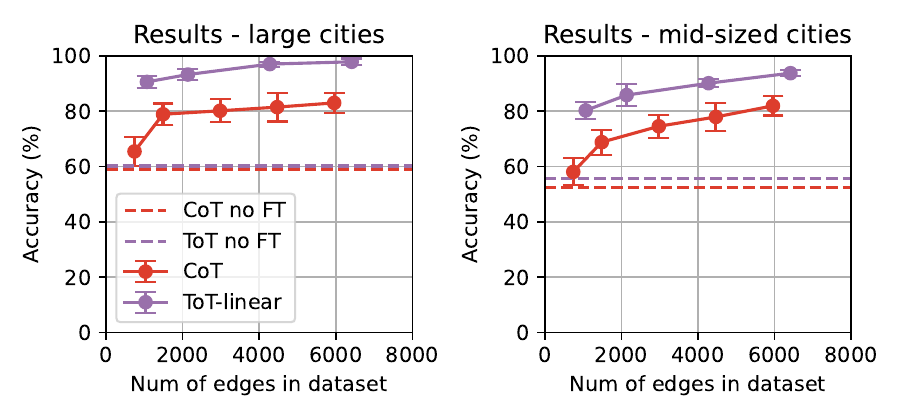}
    \caption{Results of fine-tuning Llama2-7b using different dataset sizes. The \emph{CoT no FT} and \emph{ToT no FT} means using the pre-trained Llama2-7b with CoT and ToT.}
    \label{fig:travel-llama}
\end{figure}

\paragraph{Fine-tuning}
We perform fine-tuning experiments which allows us to better control the amount of training data used in the experiments. 
The results of our fine-tuning experiments are in Fig.~\ref{fig:travel-llama}. Each ToT-linear example is longer than a CoT example; hence, we plot the results based on the number of edges observed in training. The results are consistent with the complexity analysis, with ToT-linear performing better than CoT. ToT-linear can also be viewed as a CoT where the intermediate computations are explicitly described instead of being left for the LLM to learn implicitly.

\subsection{Game of 24}
\dontshow{\comment{
\begin{enumerate}
    \item Analysis of complexity
    \item Experiment on CoT vs ToT
    \item Experiments on refining the transition decomposition, making explicit hidden steps in solution extraction.
    \item Discussion on what we learn from this case.
\end{enumerate}
}}

Many puzzles like Game of 24 are designed to be hard and unlikely to be efficiently solvable \cite{kendall2008survey}, although we are not aware of results on the computational complexity of the Game of 24\footnote{A modified version with $N$ rather than four numbers, arbitrary target number instead of 24, and only addition and multiplication with zero allowed is the same as subset-sum, an NP-complete problem. This suggests that similar puzzles are computationally difficult to solve.}. We use the Game of 24 shown in the introduction: given four numbers, the player must use basic arithmetic operations $(+, -, \times, \div)$ and all four numbers to reach 24. These types of puzzle games are often designed to be hard to solve . 

The results in \citet{yao2023tree}, obtained with in-context learning, show that CoT fails while ToT does substantially better. We extend the results by showing that CoT fails in fine-tuning as well, suggesting that the failure is likely due to the mismatch between the computational structure of CoT and the problem. We also consider the decomposition of the actions for in-context learning and show that the decomposition of complex actions into a sequence of simpler actions within a ToT can lead to substantial improvement in performance.

\subsubsection{Analysis}

We provide a general form of Game of 24 for analysis. Assume $N$ numbers are given, and each number can take $K$ different values. The goal is to use those numbers with arithmetic operations $(+, -, \times, \div)$ to reach $T$. For the standard Game of 24, $N=4$, $T=24$.

\paragraph{Direct}
Represented as a table, there are $K^N$ inputs. A solution is an expression consisting of the $N$ numbers together with $N-1$ operations and corresponding parentheses. Assuming $\log K$ bits to represent numbers, this can be represented using $O(N\log K)$ bits, giving a total table size of $O(K^NN\log K)$ bits.

\paragraph{CoT}
For CoT, the $N-1$ operations are produced in a step-by-step manner. Each step has $N(N-1)/2$ ways to select two numbers and 6 distinct operations (two orderings for $-$ and $\div$), giving $3N(N-1)$ possible actions. The transition of each operation can be represented with a table with $K^2$ entries using $O(K^2\log K)$ bits, although pre-training likely has learned these operations for small $K$. This gives a total description length of $O(N^2K^2\log K)$ if each action is learned using its own table. If we decompose the selection of two numbers and the arithmetic operation into two steps, then the total description length is $O(N^2+K^2\log K)$, and we consider this decomposition in our experiments (see Appendix \ref{lst:tot_decomp} for examples). Like other computationally difficult problems, there is no simple known policy for selecting the next action. A simple tabular policy would have $O(K^N)$ entries, each described using $O(\log N)$ bits.\\
\paragraph{ToT}
ToT uses the same actions as CoT but does not need a policy. Instead, we have a goal recognizer and an evaluation function that decides which nodes to expand. Verifying whether a solution is correct can be done in $O(N)$ time, hence a goal recognizer with a small representation exists. Difficult computational problems typically do not have a simple evaluation function for intermediate steps; a tabular evaluation function would have $O(K^N)$ entries. However, a ToT may use a larger computation budget to search a larger part of the search tree when the evaluation function is weaker, compared to CoT, where the next action is selected with a fixed learned policy.

\subsubsection{Experiments}

\begin{figure}[htbp]
    \centering
    \includegraphics[width=0.8\linewidth]{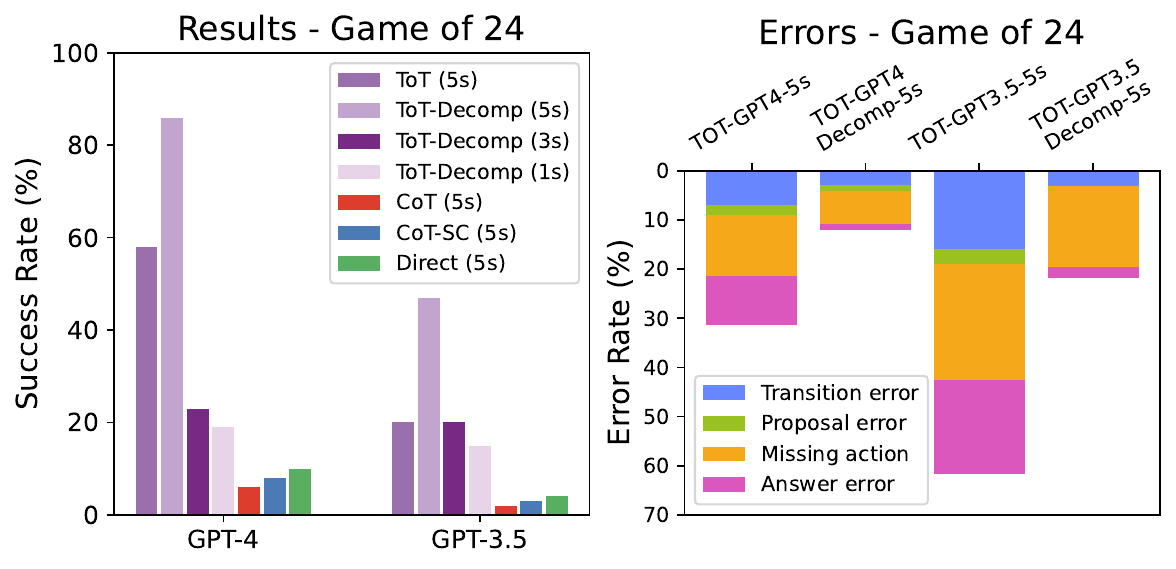}
    \caption{Results of Game of 24. 5s, 3s and 1s means 5, 3, and 1 examples in the prompt for few-shot in-context learning. }
    \label{fig:24_results_main}
\end{figure}

We perform fine-tuning with 1200 solution trajectories of Game of 24 to demonstrate the difficulty of learning a small chain solution. Both CoT and Direct failed in all test cases, indicating that moderate amounts of data are insufficient for learning in these settings. For in-context learning, the success rate of the 100 games is reported in Fig.~\ref{fig:24_results_main}.

For ToT, we use a beam search with a beam width of 5 and the same action and self-evaluation prompts as \citet{yao2023tree}. We also constructed a more decomposed version of ToT, ToT-Decomp, where we decompose the action into two steps: the selection of two numbers and the arithmetic operation. Also, ToT-Decomp uses a small CoT that provides the steps for constructing the final equation from the sequence of actions and states in the solution, whereas ToT directly generates the final equation from the action-state sequence.

From the results we see that ToT clearly outperforming CoT and Direct. Additionally, ToT-Decomp significangly outperforms ToT, demonstrating the advantages of decomposition even within the components of ToT. Error analysis (Fig.~\ref{fig:24_results_main}) categorizes the errors into four types: 1) transition error -- the next state (remaining numbers) is generated incorrectly; 2) proposal error -- the LLM does not generate the correct numbers in the action expression; 3) missing actions -- there are valid actions but are not proposed by the LLM; and 4) answer error -- the search is correct but the final expression is incorrect. ToT-Decomp notably reduces each error type compared to ToT.

\subsection{Blocksworld}

Blocksworld is a planning task motivated by robot manipulation. It has a table with blocks in different colours. Given a set of preferred constraints for the configuration of blocks, the agent needs to output a sequence of actions to rearrange the blocks into the target state that satisfies those constraints. The agent can only pick up the block if it has no blocks on its top and place it in an open space or stack on a top block. For example, with a starting state of blocks stacking in a specific order, the goal could be ``the red block is on top of the yellow block, and the blue block is at the bottom.'' The goal provides constraints of the goal state, and the agent should find a sequence of pick and place operations on the blocks to satisfy the constraints. It is a typical planning problem with constraint satisfaction in a finite domain, which is NP-complete to find the shortest plan \cite{gupta1992complexity}. 

In this section, we conduct an empirical study based on \citet{hao2024llm} with some new contributions: 1) We conduct finetuning experiments to verify the difficulty of learning the shortest plan solution for this problem. 2) We use a more realistic version of ToT where we replace the ground-truth action list in branches and perfect goal recognition in \citet{hao2024llm} with possibly unreliable LLM-based action proposal and goal recognition. 3) We show that further decomposing the direct action proposal which does not reveal the satisfied pre-conditions into a CoT that reveals the those pre-conditions before proposing the actions improves performance.
\subsubsection{Analysis}

\begin{figure}
    \centering
    \includegraphics[width=\linewidth]{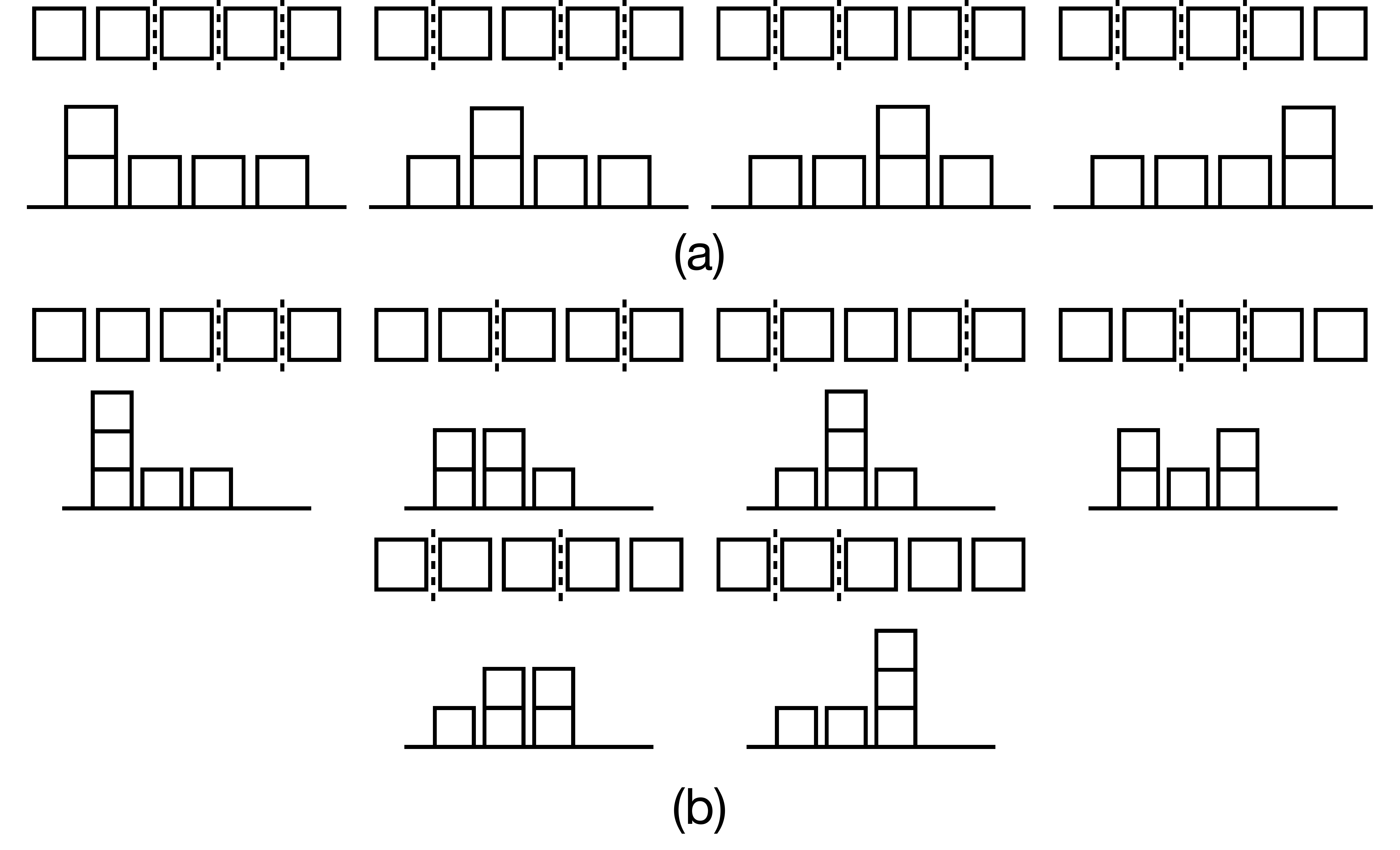}
    \caption{We take stacking 5 blocks in 4 stacks (shown in a) and 3 stacks (shown in b) as examples. (a) shows that stacking 5 blocks in 4 stacks is equivalent to choosing 3 out of 4 margins. (b) shows that stacking 5 blocks in 3 stacks is equivalent to choosing 2 out of 4 margins. }
    \label{fig:stacking example}
\end{figure}

We first upper bound the number of states. Assume there are $K$ blocks on the table. The blocks can be stacked arbitrarily on the table. Since the state is described in natural language, there must be a specific order in describing each stack of blocks. Therefore, the stacks of blocks in this state description are permutative. We first count the total number of stacking ways and their permutations, assuming the blocks are all in the same colours, i.e., without considering the order of blocks. For example, 5 blocks can be placed in 4 stacks, where one stack has two blocks while the other stacks have one block. In that case, there will be four stacking cases: (2, 1, 1, 1), (1, 2, 1, 1), (1, 1, 2, 1), and (1, 1, 1, 2). This is equivalent to choosing 3 out of 4 margins demonstrated in Figure \ref{fig:stacking example} (a). 5 blocks can also be placed in 3 stacks, a choice of 2 in 4 margins, which is shown in Figure \ref{fig:stacking example} (b). Thus, for $K$ blocks, there will be $\sum_{k=0}^{K-1} {K-1\choose k}$ is the total number of different stacking ways. Now, when the blocks are unique in colours, there will be $K!$ different ways of arranging the order of the blocks. Thus, the total number of states is $K!\sum_{k=0}^{K-1} {K-1\choose k}=K!2^{K-1}$, where $K!$ is the total number of permutation of all blocks, and $\sum_{k=0}^{K-1} {K-1\choose k}$ is the total number of different stacking ways. There are four types of actions available: pick up a block, unstack a block from on top of another block, put down a block, or stack a block on top of another block. The goal can be up to 3 pairwise spatial relation constraints. If the state satisfies the constraints, the task is considered a success. 

\paragraph{Direct} Learning to directly predict the sequence of actions can be represented as a table. The rows and columns of the table are the starting states and goal state constraints. Each entry records a sequence of actions. Assume the maximum number of actions is proportional to the number of blocks. The total size of the table is $O(K!K^32^{K-1}\log K)$ bits.

\paragraph{CoT} For CoT, the actions are generated step by step. In each step, the LLM generates the next action and predicts the next state, given the action and current state. The LLM needs to learn the precondition and effect of each action. The total number of grounded actions is $2K + 2K(K-1) = 2K^2$. Each action needs to learn its pre-conditions and effects. The state can be described using $O(K^2)$ propositions such as \texttt{On(A,B)} and \texttt{Clear(A)}. The pre-condition is a subset of these propositions that needs to be true. The effects consist of the add effect, which is a subset of propositions that will become true upon the execution of the action, and the delete effect, which is a subset of propositions that will become false. 
Assuming that the subsets in the pre-condition, add effects and delete effects have size at most $k$, we need to describe the choice of $k$ propositions from $O(K^2)$ possible propositions, which requires $O(k\log K)$ bits. With $O(K^2)$ actions, the total description length for describing the action preconditions and effects is $O(kK^2\log K)$ bits. In Blocksworld, the value of $k$ is a small constant and the complexity of the transitions seems reasonable.


Given the current state and goal constraints, we also require a policy to predict the next action. This can be represented as a look-up table whose rows and columns are the current state and goal constraints, and the entry is the next action. Describing one action requires $O(\log K)$ bits. Thus, the total size of the table is still $O(K!K^22^{K-1}\log K)$ bits, suggesting that a policy in tabular form may be difficult to learn. 

\paragraph{ToT} Unlike CoT, ToT does not require a policy for predicting the next correct actions, although it needs to learn the pre-condition and effects of each action. The LLM needs to behave as a world model that proposes valid actions, predicts the next state, and recognizes the goal. ToT uses the same actions as CoT, and our analysis has reasonable sample complexity. Verifying whether the solution is valid takes $O(K)$ time; hence, a small representation of the goal recognizer is possible. Similar to the game of 24, the self-evaluation (heuristic) could be complex, requiring $O(K!K^22^{K-1})$ entries. However, ToT spends more computational costs to conduct look-ahead searches; hence, we expect that its performance may be better than that of CoT. 

\subsubsection{Experiments}

We test the performance of Direct, CoT, CoT-SC, and ToT using four-block-stacking tasks. We test the in-context learning (ICL) performance of GPT-3.5 and GPT-4 and fine-tune the Llama-2-7b \cite{touvron2023llama} using Direct and CoT. We use the same testing dataset as \cite{hao2024llm}\footnote{\href{https://github.com/maitrix-org/llm-reasoners}{\texttt{https://github.com/maitrix-org/llm-reasoners}}}. As for the fine-tuning experiments, we generate 1000 trajectories for block stacking using the same distribution of the testing data. Our in-context learning experiment is adapted from \citet{hao2024llm}, but we have modified the reasoning methods: 1) The CoT in \citet{hao2024llm} is essentially Direct in our definition, while our CoT keeps tracking the state changes in each step. 2) Our ToT modified the RAP method in \citet{hao2024llm} to use LLM to propose all the actions in the branches, predict the next state after applying one action, and recognize the goal. In comparison, the RAP only use LLM for next-state prediction and uses the ground truth admissible actions. 3) We provide two versions of ToT: ToT and ToT-Decomp. ToT uses LLM to propose possible actions directly given the current state. ToT-Decomp uses a CoT prompt to propose possible actions. Instead of proposing possible actions directly from the current state, ToT-Decomp first explains the reasons that each proposed action is admissible (pre-conditions satisfied) before generating the possible action branches. For example, ToT-Decomp would explain that \texttt{Holding(x)} and \texttt{Clear(y)} are both be true before proposing \texttt{Stack(x,y)} as a possible action.  This can also be viewed as making the pre-conditions that are hidden in the direct prompt visible via CoT prompting. 

\begin{figure}[t]
    \centering
    \includegraphics[width=\linewidth]{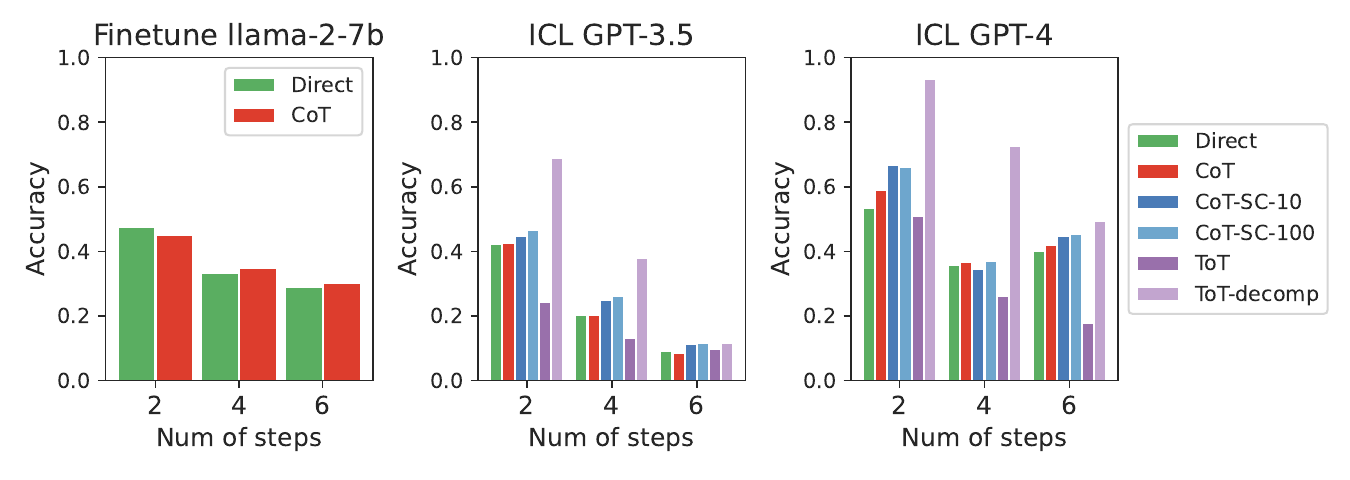}
    \caption{Results of the Blocksworld experiments. ICL stands for in-context learning. }
    \label{fig:bw-results}
\end{figure}


The experimental results are reported in the Figure \ref{fig:bw-results}. Both the Direct and CoT do not perform well in the fine-tuning experiments. It suggests that learning the short-chain solution for this planning problem is not sample-efficient. This is because finding the short-chain solution for this problem is likely computationally hard. It is also reflected by the description length analysis: Direct and CoT have similar, very large tabular description lengths, suggesting that their sample complexities are both very high. 

For the in-context learning (ICL), we also found that Direct and CoT have very similar performance in both experiments using GPT-3.5 and GPT-4. When using CoT-SC with 10 samples and 100 samples, the performance improved a bit in general, but not very much. CoT-SC achieves higher performance than CoT because it spends more on computation by sampling more reasoning trajectories and performing a majority vote, effectively decreasing the variance and improving the performance. Simple ToT performs even worse than Direct. This is because the LLM was not able to learn the pre-conditions directly from the small number of examples and the invalid actions compromised the entire performance. On the other hand, using ToT-Decomp achieved the best performance. This is because using a reasonable decomposition which makes the hidden pre-conditions of the actions visible produced actions that are almost all admissible, enabling the lookahead search of ToT to be effective. Even though the self-evaluation is inaccurate, ToT searches larger areas using the more accurate reward model and transitions, making it more effective than CoT and its variants.

\section{Conclusion}
This paper provides a detailed empirical study to understand the effectiveness of CoT and ToT reasoning in planning and reasoning tasks using notions of sample and computational complexity in machine learning. We view CoT and ToT as decomposition methods for the underlying problem and study the complexity of the component predictors in the decomposed problems. Our study finds that when the solution can be decomposed as a chain of reasoning steps where predicting the next step is not difficult, explicitly demonstrating the reasoning chain during learning can be helpful. Leaving out important variables for deciding the next reasoning step instead of making all relevant variables explicit in the demonstrations will make learning more difficult. Finally, when algorithmic analysis indicates that predicting the next reasoning step in a CoT is computationally hard, a ToT structure can be helpful.

\paragraph{Limitations}
The suggested methodology from this paper is to analyse the chain-of-thought as a decomposition of the problem and to analyse the complexity of its components. If learning the components has low sample complexity and the computational complexity of predicting the next reasoning step is low, then learning to solve the problem using a chain-of-thought would be reasonable. On the other hand, if the computational complexity of predicting the next reasoning step is high, it may be reasonable to consider learning the components and using a tree-of-thought to solve the problem. This oversimplifies various aspects of the problem. Even though the components have low sample complexity, it may be difficult to learn them in practice as the computational complexity of learning may be high, although this may be alleviated by overparameterization of the predictors used to learn the components. Another issue is out-of-domain generalization. As shown in the MWIS case study, generalization in-domain does not mean that the method will generalize out-of-domain, which may be further exacerbated by overparameterization. Further limitations may apply when doing in-context learning where very few examples are used. Performance may depend heavily on the pre-trained LLM used in this setting. Nonetheless, our case studies suggest that the proposed methodology may still be useful in the in-context learning setting. We would suggest using the guidelines proposed in this paper in a similar way that the Occam Razor principle in the philosophy of science is used. Occam's Razor suggests that simple explanations for a scientific phenomenon be preferred until shown otherwise by observations. The suggestions we proposed may not work all the time but should similarly be preferred until empirical observations suggest otherwise.

\paragraph{Ethics Statement}
This paper studies reasoning and planning in LLMs from a general perspective. While we do not focus on ethics issues, reasoning and planning techniques can potentially be useful in ensuring that AI agents behave ethically through the use of appropriate reward or goal functions that may possibly be learned from data. They may also be used in harmful ways in planning more sophisticated attacks against others.  Research on both the use of reasoning and planning for ensuring ethical AI agent behaviour and in mitigating the use of reasoning and planning in performing harmful attacks should be encouraged.

\bibliography{custom}

\clearpage
\appendix

\section{Experimental Details}
\label{sec:experiment_detail}
All prompting experiments are done with \texttt{gpt-3.5-\\turbo-1106} and \texttt{gpt-4-1106-preview}. All fine-tuning experiments are done with Llama2-7B-chat \cite{touvron2023llama} with LoRA $r=64$, $\alpha =16$ \cite{hu2021lora} applied to query and value matrices, and uses $batch\_size=1$ and gradient accumulation steps$=32$. \dontshow{The template word problem is fine-tuned for 10 epochs with a learning rate of $1e-3$.} MWIS and Game of 24 are fine-tuned for 5 epochs with a learning rate of $3e-4$. Travel planning is fine-tuned for 300 gradient optimization steps with a learning rate of $3e-4$. The fine-tuning data is wrapped in the template \textit{"<s> [INST] \{\{prompt\}\} [/INST] \{\{completion\}\} </s>"} and the loss is calculated on completion tokens.


\section{GSM8K}
\subsection{Statistics}
\label{sec:gsm_statistics}
In the GSM8K dataset, the variable values ($K$) are usually from $1$ to $6$ digits and the average number of variables per question ($N$) is $3.93$. The average number of variables ($a_i$) is $2.19$, the average number of reasoning steps is $3.17$, so each step is relatively simple and the number of steps are also small. We also find that the number of rules that need to be learned as world knowledge appears to be relatively small (see Appendix~\ref{sec:appendix_rules}). Overall, we see that the decomposed components in CoT are relatively simple, and suggest that decomposition with CoT may be reasonable for GSM problems.
\subsection{Common Rules in GSM8K}
\label{sec:appendix_rules}
We analyzed 50 problems from the GSM8K training set and identified a set of rules. The first five are general rules that can be inferred from the questions and are applicable to multiple problems. The last four are question-specific rules, involving commonsense knowledge that are not mentioned in the questions.
\begin{enumerate}
    \item Amount A = Amount B * multiplier
    \item Amount A = Amount B + difference
    \item Total = N\_unit * Amount per unit
    \item Total = Sum of components
    \item Current Amount = Initial Amount - Amount Given + Amount Received
    \item Question-specific (implicit): One hour = 60 Minutes
    \item Question-specific (implicit): one sandwich has two slices of bread
    \item Question-specific (implicit): face has two eyes
    \item Question-specific (implicit): 1 quarter = 25 cent; 1 dime = 10 cent; 1 nickel = 5 cent
\end{enumerate}

\subsection{An Efficient Algorithm for GSM8K}
\label{sec:gsm8k_fast_algo}
Based on our analysis of the GSM8K problems in \ref{sec:gsm_analysis}, we give a formulation of the GSM8K problems, and show that there exists an algorithm that has runtime linear to the total input length.

\paragraph{Problem Formulation}

\textbf{Input}: A set of $N$ variables $\{V_1, ..., V_N\}$, where the values of some variables are known (from natural language input), while some are unknown (to be inferred); A set of $M$ equations $\{R_1, ..., R_M\}$, where all equations have exactly one variable on LHS; A target variable $V_t$ whose value we want to know.

\textbf{Output}: The value of $V_t$.

The solvability of the problem ensures that for all variables, if the value is not given in the natural language question, will appear on the LHS of some equation.

\paragraph{An Efficient Algorithm}:

Inspired by \citet{dowling1984linear}, we design an algorithm whose runtime is linear to the size of the problem (total length of all equations).

We maintain a list \texttt{numvars}$[M]$ which stores the number of unsolved variables on RHS for each equation; a list \texttt{lhslist}$[M]$ which stores which variable is on LHS of an equation; a list \texttt{equationlist}$[N]$ which stores the index of the equations where the corresponding variable appears on RHS. We say an equation $R_i$ is ready to be processed if \texttt{numvars}$[i]=0$. We maintain a queue that will contain the equations that are ready to be processed, and it is initialized to contain the equations that are ready to process given the known variables from natural language input.

Then we loop over the queue. Let \texttt{equation1} be the current head of the queue and let \texttt{nextvar}$ = $\texttt{lhslist}$[$\texttt{equation1}$]$ be the variable on the LHS of of \texttt{equation1}. Pop the head of the queue, and iterate over $\texttt{equationlist}[\texttt{nextvar}]$, for every $\texttt{equation2}$ in it, reduce $\texttt{numvars}[\texttt{equation2}]$ by $1$, and if $\texttt{numvars}[\texttt{equation2}]$ becomes $0$, add \texttt{equation2} to the queue.

Loop until the queue is empty, we would have solved the values of all $N$ variables. Refer to Algorithm \ref{alg:forward_chaining_gsm8k} for a more concise representation of the algorithm. 

\begin{algorithm}
\caption{Forward Chaining Algorithm for GSM8K Problems}
\label{alg:forward_chaining_gsm8k}
\begin{algorithmic}[1]

\Require
\Statex $N$ variables $\{V_1, \ldots, V_N\}$, with known and unknown values.
\Statex $M$ equations $\{R_1, \ldots, R_M\}$, each with one variable on LHS.
\Statex Target variable $V_t$ to find the value of.

\Ensure
\Statex Value of $V_t$.

\State \texttt{numvars}$[1 \ldots M] \gets$ Number of unsolved variables on RHS of each $R_i$.
\State \texttt{lhslist}$[1 \ldots M] \gets$ Variable on LHS of each $R_i$.
\State \texttt{equationlist}$[1 \ldots N] \gets$ Indices of $R_i$ where $V_j$ appears on RHS.

\State Initialize queue \texttt{readyQueue} with indices of $R_i$ where \texttt{numvars}$[i] = 0$.

\While{\texttt{readyQueue} is not empty}
    \State $i \gets$ Dequeue from \texttt{readyQueue}.
    \State $V_j \gets$ \texttt{lhslist}$[i]$.
    \For{each $k$ in \texttt{equationlist}$[j]$}
        \State \texttt{numvars}$[k] \gets$ \texttt{numvars}$[k] - 1$.
        \If{\texttt{numvars}$[k] = 0$}
            \State Enqueue $k$ into \texttt{readyQueue}.
        \EndIf
    \EndFor
\EndWhile

\State \Return Value of $V_t$.

\end{algorithmic}
\end{algorithm}

\paragraph{Complexity of the Algorithm}
\texttt{numvars} and \texttt{lhslist} can be initialized in $O(L)$, where $L$ is the total length of all equations. When processing an equation, the decrement of \texttt{numvars} corresponds to the "\textit{deletion}" of occurrences of the variable in an equation, each variable in the equation is looked only once, thus processing all equations also runs in $O(L)$. Overall, the runtime of the algorithm is $O(L)$, i.e. linear to the total length of the equations. If we assume each variable appears only once on the LHS of equations, and each equation has a number of variables up to a constant $C$ (as we have seen in \ref{sec:gsm_analysis}, this is often true for GSM8K problems), the runtime of the algorithm would be $O(N)$, i.e. linear to the number of variables in the problem.

\subsection{GSM8K Errors}
\label{sec:gsm_error}
We observe three types of errors from GPT, namely grounding error, calculation error, and missing rule. We provide an example for each type.
\begin{lstlisting}[caption={Grounding Error}, label={lst:gsm_grounding_error}]
Question:
Annie was given a pack of crayons. The pack contained 21 crayons. She already had a box of 36 crayons in her locker. Her friend Bobby gave her half the amount she already had in her locker. She decided to give her sister Mary 1/3 of her total amount of crayons. How many crayons does she give to Mary?

GPT-3.5 response:
"Annie originally had 21 crayons in the pack and 36 crayons in her locker.
So she had a total of 21 + 36 = 57 crayons.
REDREDBobby gave her half the amount she already had, so he gave her 57 / 2 = 28.5 crayonsREDRED, but since you can't have half a crayon, let's round down to 28.
So she now has 57 + 28 = 85 crayons.
So 1/3 of her total amount of crayons is 85 / 3 = 28.33 crayons, but since you can't have a fraction of a crayon, let's round down to 28 crayons.
So she gives Mary 28 crayons.
So the answer is 28."
\end{lstlisting}

In Listing \ref{lst:gsm_grounding_error}, the problem involves using the rule \textit{"Amount A = Amount B * multiplier"}, which should be grounded as \textit{"Amount from Bobby = Amount in Locker * 1/2"}, not \textit{"Amount from Bobby = Amount Annie have in total * 1/2"}.

\begin{lstlisting}[caption={Calculation Error}, label={lst:gsm_calc_error}]
Question:
To get free delivery, Alice needs to spend a minimum of $35.00 online at her favorite grocery store.  In her cart she has 1.5 pounds of chicken at $6.00 per pound, 1 pack of lettuce for $3.00, cherry tomatoes for $2.50, 4 sweet potatoes at $0.75 each, 2 heads of broccoli for $2.00 each and a pound of Brussel sprouts for $2.50.  How much more does she need to spend in order to get free delivery?

GPT-3.5 response:
"Alice's total so far is:
REDRED(1.5 * 6) + 3 + 2.50 + (4 * 0.75) + (2 * 2) + 2.50
= 9 + 3 + 2.50 + 3 + 4 + 2 + 2.50
= 26.00REDRED

To get free delivery, she needs to spend at least $35.00.
So she needs to spend 35 - 26
= 9.00 more to get free delivery.

So the answer is 9.00."
\end{lstlisting}

In Listing \ref{lst:gsm_calc_error}, the calculation is wrong.

\begin{lstlisting}[caption={Missing Rule}, label={lst:gsm_missing_rule}]
Question:
A Ferris wheel can accommodate 70 people in 20 minutes.  If the Ferris wheel is open from 1:00 pm until 7:00 pm, how many people will get to ride?

GPT-3.5 response:
"The Ferris wheel can accommodate 70 people in 20 minutes.
REDREDThere are 6 periods of 20 minutes from 1:00 pm to 7:00 pm.REDRED
So in total, the Ferris wheel can accommodate 70 * 6 = 420 people.
So 420 people will get to ride the Ferris wheel from 1:00 pm until 7:00 pm.
So the answer is 420."
\end{lstlisting}
In Listing \ref{lst:gsm_missing_rule}, one commonsense knowledge (one hour has 60 minutes) not mentioned in the question is not properly used in the solution.

\section{Dynamic Programming}
\subsection{Algorithm for MWIS}
An algorithm that solves the Maximum Weighted Independent Set problem and runs in time $O(N)$ where $N$ is the number of numbers in the input is shown in Algorithm \ref{alg:dp_mwis}.
\label{sec:dp_algo}
\begin{algorithm}
\caption{Dynamic Programming Algorithm for the Maximum Weighted Independent Set problem}
\label{alg:dp_mwis}
\begin{algorithmic}[1]

\Require An array $arr$ of integers
\Ensure A sequence of decisions maximizing a certain criterion based on $arr$

\State $N \gets \text{length of } arr$
\State Initialize $dp[0 \ldots N-1]$ with zeros

\State $dp[N-1] \gets \max(arr[N-1], 0)$
\State $dp[N-2] \gets \max(arr[N-1], arr[N-2], 0)$

\For{$i \gets N-3$ \textbf{downto} $0$}
    \State $dp[i] \gets \max(dp[i+1], arr[i] + dp[i+2], 0)$
\EndFor

\State Initialize $result$ as an empty list
\State $can\_access\_next\_item \gets \textbf{true}$

\For{$i \gets 0$ \textbf{to} $N-3$}
    \If{$dp[i] = arr[i] + dp[i+2]$ \textbf{and} $can\_access\_next\_item$}
        \State Append $1$ to $result$
        \State $can\_access\_next\_item \gets \textbf{false}$
    \Else
        \State Append $2$ to $result$
        \State $can\_access\_next\_item \gets \textbf{true}$
    \EndIf
\EndFor

\If{$dp[N-2] = arr[N-2]$ \textbf{and} $can\_access\_next\_item$}
    \State Append $1$ to $result$
\Else
    \State Append $2$ to $result$
\EndIf

\If{$dp[N-1] = arr[N-1]$ \textbf{and} $can\_access\_next\_item$}
    \State Append $1$ to $result$
\Else
    \State Append $2$ to $result$
\EndIf

\State \Return $result$

\end{algorithmic}
\end{algorithm}


\section{Inductive bias of Transformers}
\label{sec:inductive_bias}

From the math word problem and dynamic programming fine-tuning experiments, we see that for some tasks (maximum weighted independent set), the transformer can learn to directly answer the problem efficiently, while for some other tasks (word problem), the direct answer is hard to learn.


We conduct two more experiments to study what might affect the performance of learning to direct answer other than sample complexity: 1) learn the \texttt{max} function, where the input is a list of integers, and the expected output is the maximum value in the input list. This requires only looping over the sequence once, and storing one intermediate value; 2) another dynamic programming problem called \texttt{rain water}\footnote{\href{https://leetcode.com/problems/trapping-rain-water/}{\texttt{https://leetcode.com/problems/trapping-rain-\\water/}}} that requires looping over the array three times and storing two one-dimensional arrays for memorization. These two problems are similar to MWIS as they all require looping over the input sequence and maintaining some internal variables during the iteration. We use them to study whether the difference between learning to directly answer the word problem and MWIS is related to the inductive bias of transformers. To eliminate the confounding part, the difficulty of language in the word problem, we perform a modified version of the problem, where we remove all natural language in the prompt, and use a fixed formula for ground-truth answer: $(v_1v_2+v_1v_3+v_1v_3/v_5+v_1v_2/v_4)v_7/v_6$. The input would look like \textit{"1, 6, 4, 3, 2, 14, 8"}, and the expected output for this example would be \textit{"8"} ($(1 \cdot 6 + 1 \cdot 4 + 1 \cdot 4/2 + 1 \cdot 6/3) \cdot 8/14=8$). We randomly sample the values of the variables, ensuring the answer value is integer to construct the dataset.

\begin{table}[htbp]
\centering
\begin{tabular}{lc}
\hline
Task       & Accuracy (\%) \\ \hline
MWP        &58.00\\
MWIS ($n\in[4,5,6]$) &98.89\\
MWIS ($n=200$)       &0.01\\
max ($n=30$)&99.50\\
rain water ($n=10$) &89.00\\ \hline
\end{tabular}

\caption{Fine-tuning results of different problems. MWP stands for the modified word problem where the input contains only 7 numbers. All tasks are fine-tuned with 10k direct answer examples and evaluated on in-domain examples.}
\label{tab:inductive_bias}
\end{table}

From the table, we see that MWIS, \texttt{max}, and \texttt{rain water} perform significantly better than MWP. This suggests that it might be easy for transformers to learn this loop type of problem when the problem size is small. However, when the problem size of MWIS is large ($n=200$), the model fails to generalize to unseen test examples. This aligns with previous findings \cite{weiss2021thinking, zhou2024understanding} that suggest that it would consume one transformer layer to approximate one iteration in an algorithm. And with a problem size of $200$, it can be hard for transformers to approximate the algorithm in a generalizable way, thus some other patterns in the training set may be exploited, leading to poor generalization.

\begin{figure}[htbp]
    \centering
    \includegraphics[width=0.8\linewidth]{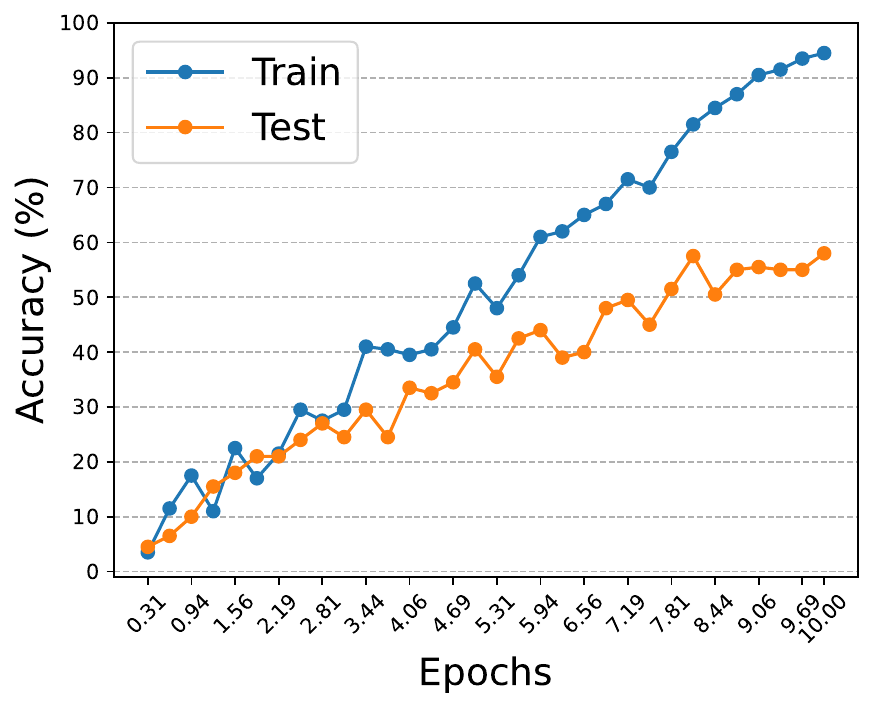}
    \caption{Results of fine-tuning word problem.}
    \label{fig:word_problem_computation}
\end{figure}

From Fig.~\ref{fig:word_problem_computation} we can see that for the modified word problem, the transformer can fit the training set reasonably well, while the test set performance peaks at $58.0\%$. This suggests that by learning to answer directly, the transformer is behaving similarly to learning by filling a table, instead of learning the underlying rational function, which supports our description length analysis.



\section{Air travel planning}
\label{sec:air_travel_planning_appendix}
We use the Kaggle World Cities\footnote{\href{https://www.kaggle.com/datasets/max-mind/world-cities-database}{\tt https://www.kaggle.com/datasets/max-mind/world\\-cities-database}} database data and sample 212 cities with more than 1 million populations. We sampled 58 large cities and 154 mid-sized cities. We use the Virtual Radar Server\footnote{\href{https://github.com/vradarserver/standing-data}{\texttt{https://github.com/vradarserver/standing-data}}} to get the real-time (Jan 13, 2024) flight data as the ground truth. 

\section{Game of 24}
We test the hard games indexed 901-1000 from 4nums.com. An output is considered correct if the expression evaluates to 24 and uses all input numbers once.

\section{Tables}

\begin{table}[htbp]
\centering
\begin{tabular}{lcc}
\hline
Method & GPT-3.5 & GPT-4 \\ \hline
Direct & 28.51   & 47.16 \\
CoT    & 79.53   & 94.09 \\
ToT    & \textbf{81.88}   & \textbf{96.00} \\ \hline
\end{tabular}
\caption{Figure \ref{fig:sub1}}
\end{table}

\begin{table}[htbp]
\centering
\begin{tabular}{ccc}
\hline
Dataset size & Direct & CoT   \\ \hline
1000         & 18.50  & 88.00 \\
2000         & 22.50  & 88.00 \\
3000         & 30.50  & 92.50 \\
4000         & 35.00  & 93.50 \\
5000         & 37.50  & 95.00 \\
6000         & 46.50  & 95.00 \\
7000         & 46.50  & 96.00 \\
8000         & 48.50  & 96.50 \\
9000         & 48.50  & \textbf{97.50} \\
10000        & \textbf{58.00}  & 96.50 \\ \hline
\end{tabular}
\caption{Figure \ref{fig:sub2}}
\end{table}

\begin{table}[htbp]
\centering
\begin{tabular}{lccc}
\hline
Method & InD   & OoD   & Total \\ \hline
Direct 0s                     & 38.67 & 8.75  & 21.57 \\
Direct 3s                     & 68.67 & 35.25 & 49.57 \\
Direct 6s                     & 57.67 & 30.25 & 42.00 \\
CoT 0s                        & 59.33 & 35.25 & 45.57 \\
CoT Implicit 3s               & 67.00 & 41.50 & 52.43 \\
CoT Implicit 6s               & 65.00 & 36.25 & 48.57 \\
CoT Explicit 3s               & 85.67 & 65.00 & 73.86 \\
CoT Explicit 6s               & \textbf{86.67} & \textbf{66.50} & \textbf{75.14} \\ \hline
\end{tabular}
\caption{Figure \ref{fig:dp_incontext} GPT-4}
\end{table}

\begin{table}[htbp]
\centering
\begin{tabular}{lccc}
\hline
Method & InD   & OoD   & Total \\ \hline
Direct 0s       & 26.00        & 13.50        & 18.86        \\
Direct 3s       & 32.33        & 10.50        & 19.86        \\
Direct 6s       & 39.33        & 20.25        & 28.43        \\
CoT 0s          & 24.33        & 8.75         & 15.43        \\
CoT Implicit 3s & 18.00        & 6.75         & 11.57        \\
CoT Implicit 6s & 20.33        & 5.00         & 11.57        \\
CoT Explicit 3s & 56.67        & 16.00        & 33.43        \\
CoT Explicit 6s & \textbf{63.33}        & \textbf{28.75}        & \textbf{43.57}        \\ \hline
\end{tabular}
\caption{Figure \ref{fig:dp_incontext} GPT-3.5}
\end{table}

\begin{table}[htbp]
\centering
\begin{tabular}{lcc}
\hline
Method & Large cities   & Mid-sized cities\\ \hline
CoT 0s       & 70.76        & 50.00       \\
CoT 3s       & 73.10        & 51.64       \\
CoT 8s       & 72.51        & 53.27       \\
CoT-SC 0s    & 72.51        & 55.47 \\
CoT-SC 3s    & 74.56        & 54.13 \\
CoT-SC 8s    & 75.43        & 56.85 \\
ToT-linear 0s & 75.43        & 69.67        \\
ToT-linear 3s & \textbf{81.29}        & 77.05        \\
ToT-linear 8s & 78.36        & 72.95        \\
ToT 0s & 78.36        & 72.13       \\
ToT 3s & 80.70       & \textbf{75.41}      \\
ToT 8s & \textbf{81.29}        & \textbf{75.41}      \\ \hline
\end{tabular}
\caption{Figure \ref{fig:travel-gpt} GPT-3.5 (Accuracy, \%)}
\end{table}
\begin{table}[htbp]
\centering
\begin{tabular}{lcc}
\hline
Method & Large cities   & Mid-sized cities\\ \hline
CoT 0s       & 71.35        & 64.75       \\
CoT 3s       & 76.02        & 68.03       \\
CoT 8s       & 85.38        & 70.49       \\
CoT-SC 0s    & 74.43        & 66.21 \\
CoT-SC 3s    & 79.59        & 69.67 \\
CoT-SC 8s    & 87.13        & 71.12 \\
ToT-linear 0s & 54.24        & 47.54        \\
ToT-linear 3s & 87.13        & 69.67        \\
ToT-linear 8s & 84.80        & 68.85       \\
ToT 0s & 76.02        & 70.49       \\
ToT 3s & 88.30        & 78.69       \\
ToT 8s & \textbf{88.89}        & \textbf{79.51}      \\ \hline
\end{tabular}
\caption{Figure \ref{fig:travel-gpt} GPT-4 (Accuracy, \%)}
\end{table}

\begin{table}[htbp]
\centering
\begin{tabular}{lcc}
\hline
Num of edges & Large cities   & Mid-sized cities\\ \hline
1069      & 90.64$\pm$2.21        & 80.32$\pm$3.21       \\
2138      & 93.30$\pm$2.02        & 85.87$\pm$3.92       \\
4277      & 97.07$\pm$0.94        & 90.16$\pm$1.45       \\
6415      & \textbf{97.90$\pm$1.20}        & \textbf{93.79$\pm$1.13}        \\ \hline
\end{tabular}
\caption{Figure \ref{fig:travel-llama} ToT-linear (Accuracy \% $\pm$ standard error)}
\end{table}

\begin{table}[htbp]
\centering
\begin{tabular}{lcc}
\hline
Num of edges & Large cities   & Mid-sized cities\\ \hline
744      & 65.50$\pm$5.22        & 58.10$\pm$4.91       \\
1489      & 78.94$\pm$3.90        & 68.85$\pm$4.56       \\
2979      & 80.19$\pm$4.12        & 74.59$\pm$4.11       \\
4468      & 81.52$\pm$5.23        & 77.97$\pm$5.10        \\ 
5958      & \textbf{83.04$\pm$3.54}        & \textbf{81.98$\pm$3.41}        \\\hline
\end{tabular}
\caption{Figure \ref{fig:travel-llama} CoT (Accuracy \% $\pm$ standard error)}
\end{table}

\begin{table}[htbp]
\centering
\begin{tabular}{lcc}
\hline
Method & GPT-4   & GPT-3.5\\ \hline
ToT 5s       & 58        & 20       \\
ToT-Decomp 5s & \textbf{86}        & \textbf{47}       \\
ToT-Decomp 3s & 23        & 20       \\
ToT-Decomp 1s & 19        & 15        \\
CoT 5s        & 6        & 2        \\
CoT-SC 5s     & 11       & 7 \\
Direct 5s     & 10        & 4       \\ \hline
\end{tabular}
\caption{Figure \ref{fig:24_results_main}, main results (Accuracy, \%). }
\end{table}

\begin{table}[htbp]
\centering\small
\begin{tabular}{lcc}
\hline
Method & Transition error & Proposal error \\ \hline
TOT-GPT4-5s &  7.12  &  2.04    \\
TOT-GPT4-Decomp-5s & \textbf{2.80}   & 1.44 \\
TOT-GPT3.5-5s & 16.62  &3.15 \\
TOT-GPT3.5-Decomp-5s & 3.06     & \textbf{0.30}   \\ \hline
Method & Missing action & Answer error \\ \hline
TOT-GPT4-5s & 12.44   & 10.04  \\
TOT-GPT4-Decomp-5s &  \textbf{6.63}    & \textbf{1.56}  \\
TOT-GPT3.5-5s & 23.63   & 19.03\\
TOT-GPT3.5-Decomp-5s & 16.60  & 2.28 \\ \hline
\end{tabular}
\caption{Figure \ref{fig:24_results_main}, main results (Error rate, \%). }
\end{table}

\begin{table}[htbp]
    \centering
\begin{tabular}{lcc}
\hline
        Method& \multicolumn{1}{l}{GPT-4} & \multicolumn{1}{l}{GPT-3.5} \\ \hline
Direct          & 62.5                      & 42.14                       \\
Direct (parsed) & 73.7                      & 53.74                       \\
CoT             & 64.33                     & 52.91                       \\
CoT (parsed)    & 76.89                     & 65.62                       \\
CoT-SC          & 66.73                     & 53.79                       \\
CoT-SC (parsed) & \textbf{75.89}                     & \textbf{65.67}                       \\
ToT             & 65.36                     & 45.68                       \\
ToT (parsed)    & 75.72                     & 49.98                       \\ \hline
\end{tabular}
    \caption{Figure~\ref{fig:musique}}
\end{table}

\begin{table}[htbp]
\centering
\begin{tabular}{lccc}
\hline
Method & 2-step & 4-step   & 6-step\\ \hline
Direct      & \textbf{47.6}  & 33.2       & 28.9       \\
CoT      & 45.2 & \textbf{34.9}        & \textbf{30.1}       \\
\hline
\end{tabular}
\caption{Figure \ref{fig:bw-results}a, results of finetuning Llama-2-7b (Accuracy, \%). }
\end{table}

\begin{table}[htbp]
\centering
\begin{tabular}{lccc}
\hline
Method & 2-step & 4-step   & 6-step\\ \hline
Direct      &  42.2 & 20.2        & 9.2       \\
CoT      & 42.7 & 20.2        & 8.4       \\
CoT-SC-10      & 44.7 & 25.0        & 11.2       \\
CoT-SC-100      & 46.7 & 26.2        & \textbf{11.8}       \\
ToT      & 22.4 & 13.1        & 9.8       \\
ToT-Decomp      & \textbf{68.9} & \textbf{38.1}        & \textbf{11.8}       \\ \hline
\end{tabular}
\caption{Figure \ref{fig:bw-results}b, results of GPT-3.5 for in-context learning (Accuracy, \%). }
\end{table}

\begin{table}[htbp]
\centering
\begin{tabular}{lccc}
\hline
Method & 2-step & 4-step   & 6-step\\ \hline
Direct      & 53.3 & 35.7        & 40.1       \\
CoT      &  58.9& 36.7        & 42.0       \\
CoT-SC-10      & 66.7 & 34.5        & 44.7       \\
CoT-SC-100      & 66.2 & 36.9        & 45.4       \\
ToT      & 51.1 & 26.2        & 11.7       \\
ToT-Decomp      & \textbf{93.3} & \textbf{72.6}        & \textbf{49.3}       \\ \hline
\end{tabular}
\caption{Figure \ref{fig:bw-results}c, results of GPT-4 for in-context learning (Accuracy, \%).}
\end{table}
\newpage
\section{Prompts}
\label{sec:prompts}
\subsection{GSM8K Prompts}
\begin{lstlisting}[caption={GSM8K Direct prompt}]
direct_8s = """Please answer a math word problem given the following exapmles. Respond only the answer, in the format "The answer is ###."
Example:
Question: There are 15 trees in the grove. Grove workers will plant trees in the grove today. After they are done, there will be 21 trees. How many trees did the grove workers plant today?
The answer is 6.

Question: If there are 3 cars in the parking lot and 2 more cars arrive, how many cars are in the parking lot?
The answer is 5.

Question: Leah had 32 chocolates and her sister had 10 more chocolates than her. If they ate 35, how many pieces do they have left in total?
The answer is 39.

Question: Jason had 20 lollipops. He gave Denny some lollipops. Now Jason has 12 lollipops. How many lollipops did Jason give to Denny?
The answer is 8.

Question: Shawn has five toys. For Christmas, he got two toys each from his mom and dad. How many toys does he have now?
The answer is 9.

Question: There were nine computers in the server room. Five more computers were installed each day, from monday to thursday. How many computers are now in the server room?
The answer is 29.

Question: Michael had 58 golf balls. On tuesday, he lost 23 golf balls. On wednesday, he lost 2 more. How many golf balls did he have at the end of wednesday?
The answer is 33.

Question: Olivia has $23. She bought five bagels for $3 each. How much money does she have left?
The answer is 8.

Question: {question}
The answer is 
"""
\end{lstlisting}

\begin{lstlisting}[caption={GSM8K CoT and ToT prompt}]
cot_8s = """Please answer a math word problem given the following example. Respond with reasoning steps, and end with the answer, in the format "So the answer is ###."
Example:
Let's think step by step.
Question: There are 15 trees in the grove. Grove workers will plant trees in the grove today. After they are done, there will be 21 trees. How many trees did the grove workers plant today?
Solution: There are 15 trees originally.
And there were 21 trees after some more were planted.
So 21 - 15 = 6 trees were planted.
So the answer is 6.

Let's think step by step.
Question: If there are 3 cars in the parking lot and 2 more cars arrive, how many cars are in the parking lot?
Solution: There are originally 3 cars.
And 2 more cars arrive.
So there are 3 + 2 = 5 cars now.
So the answer is 5.

Let's think step by step.
Question: Leah had 32 chocolates and her sister had 10 more chocolates than her. If they ate 35, how many pieces do they have left in total?
Solution: Originally, Leah had 32 chocolates.
And her sister had 10 more chocolates than her.
So her sister had 42 chocolates.
So in total they had 32 + 42 = 74 chocolates.
Then they ate 35 chocolates.
Therefore they had 74 - 35 = 39 chocolates left.
So the answer is 39.

Let's think step by step.
Question: Jason had 20 lollipops. He gave Denny some lollipops. Now Jason has 12 lollipops. How many lollipops did Jason give to Denny?
Solution: Jason started with 20 lollipops.
Then he had 12 after giving some to Denny.
So he gave Denny 20 - 12 = 8 lollipops.
So the answer is 8.

Let's think step by step.
Question: Shawn has five toys. For Christmas, he got two toys each from his mom and dad. How many toys does he have now?
Solution: Shawn started with 5 toys.
And he got 2 toys each from his mom and dad.
So he got 2 + 2 = 4 toys.
Therefore, he has 5 + 4 = 9 toys now.
So the answer is 9.

Let's think step by step.
Question: There were nine computers in the server room. Five more computers were installed each day, from monday to thursday. How many computers are now in the server room?
Solution: There were originally 9 computers.
And 5 more computers were added from onday to thursday.
There are 4 days between monday and thursday.
So 5 * 4 = 20 computers were added in total.
So there are 9 + 20 = 29 computers now.
So the answer is 29.

Let's think step by step.
Question: Michael had 58 golf balls. On tuesday, he lost 23 golf balls. On wednesday, he lost 2 more. How many golf balls did he have at the end of wednesday?
Solution: Michael started with 58 golf balls.
And he lost 23 golf balls on tuesday.
So after losing 23 on tuesday, he had 58 -23 = 35.
And then he lost 2 more golf balls on wednesday.
So after losing 2 more on wednesday, he had 35 - 2 = 33 golf balls.
So the answer is 33.

Let's think step by step.
Question: Olivia has $23. She bought five bagels for $3 each. How much money does she have left?
Solution: Olivia had 23 dollars.
And she bought 5 bagels.
And each bagel costs 3 dollars.
So she spent 5 * 3 = 15 dollars.
So she has 23 - 15 = 8 dollars left.
So the answer is 8.

Let's think step by step.
Question: {question}
Solution: 
"""
\end{lstlisting}

\begin{lstlisting}[caption=GSM8K ToT self-evaluation prompts]
evaluate_prompt = '''
Q: Julie climbed 15 steps up the giant slide. She climbed down 6 steps to talk to her friend, Maria. Then she climbed up 8 steps to get to the top. How many steps does the slide have?

A:
Julie climbed 15 steps up.
# Is the above step of reasoning:
# (A) Correct
# (B) Incorrect
# The above step of reasoning is (A)
Then she climbed down 6 steps.
# Is the above step of reasoning:
# (A) Correct
# (B) Incorrect
# The above step of reasoning is (A)
Then she climbed up 8 steps.
# Is the above step of reasoning:
# (A) Correct
# (B) Incorrect
# The above step of reasoning is (A)
So she climbed 15 + 8 = 23 steps.
# Is the above step of reasoning:
# (A) Correct
# (B) Incorrect
# The above step of reasoning is (B), because she also climbed down 6 steps, so she climbed 23 - 6 = 17 steps.
So the slide has 23 steps.
# Is the above step of reasoning:
# (A) Correct
# (B) Incorrect
# The above step of reasoning is (A), but the value of steps of slides is incorrect.
So the answer is 23.
# Is the above step of reasoning:
# (A) Correct
# (B) Incorrect
# The above step of reasoning is (A), but the value of steps of slides is incorrect.





Q: Suzanne read the first 15 pages of her book on Monday. She read 16 more pages than that on Tuesday. Then there were 18 pages left. How many pages are in Suzanne's book altogether?

A:
Suzanne read 15 pages on Monday.
# Is the above step of reasoning:
# (A) Correct
# (B) Incorrect
# The above step of reasoning is (A)
Then she read 16 more pages on Tuesday.
# Is the above step of reasoning:
# (A) Correct
# (B) Incorrect
# The above step of reasoning is (A)
So she read 15 + 16 = 31 pages in total.
# Is the above step of reasoning:
# (A) Correct
# (B) Incorrect
# The above step of reasoning is (B), because she read 16 more pages than that on Tuesday, so she read 15 + 16 = 31 pages on tuesday. So she read 15 + 31 = 46 pages in total.
Then there were 18 pages left.
# Is the above step of reasoning:
# (A) Correct
# (B) Incorrect
# The above step of reasoning is (A), but the value of total read pages of monday and tuesday is incorrect.
So the book had 31 + 18 = 49 pages.
# Is the above step of reasoning:
# (A) Correct
# (B) Incorrect
# The above step of reasoning is (A), but the value of total read pages of monday and tuesday is incorrect. So the book had 46 + 18 = 64 pages.
So the answer is 49.
# Is the above step of reasoning:
# (A) Correct
# (B) Incorrect
# The above step of reasoning is (A), but the value of total read pages of monday and tuesday is incorrect.





Q: Allison brought some CDs online. Each CD cost $7. There was an additional charge of $4 per order for shipping costs. The total bill came to $60. How many CDs did Allison buy? 

A:
Each CD cost 7 dollars.
# Is the above step of reasoning:
# (A) Correct
# (B) Incorrect
# The above step of reasoning is (A)
And there was an additional charge of 4 dollars.
# Is the above step of reasoning:
# (A) Correct
# (B) Incorrect
# The above step of reasoning is (A)
So the total cost of each CD is 7 + 4 = 11 dollars.
# Is the above step of reasoning:
# (A) Correct
# (B) Incorrect
# The above step of reasoning is (B), because each CD cose 7 dollars.
So 60 / 11 = 5.45.
# Is the above step of reasoning:
# (A) Correct
# (B) Incorrect
# The above step of reasoning is (B), because it cost 4 dollars for shipping costs. So the cost of CDs is 60 - 4 = 56 dollars. So Allison bought 56 / 7 = 8 CDs.
So the answer is 5.
# Is the above step of reasoning:
# (A) Correct
# (B) Incorrect
# The above step of reasoning is (A), but the value of number of CDs is incorrect.





Q: Luis and Cameron shared some stickers is the ratio 5:2. Luis received 15 more stickers than Cameron. How many stickers were there altogether?

A:
Let's say there were x stickers.
# Is the above step of reasoning:
# (A) Correct
# (B) Incorrect
# The above step of reasoning is (A)
Then Luis got 5x/7 and Cameron got 2x/7.
# Is the above step of reasoning:
# (A) Correct
# (B) Incorrect
# The above step of reasoning is (A)
Luis got 15 more than Cameron, so 5x/7 - 2x/7 = 15.
# Is the above step of reasoning:
# (A) Correct
# (B) Incorrect
# The above step of reasoning is (A)
So 3x/7 = 15.
# Is the above step of reasoning:
# (A) Correct
# (B) Incorrect
# The above step of reasoning is (A)
So x = 105.
# Is the above step of reasoning:
# (A) Correct
# (B) Incorrect
# The above step of reasoning is (B), because 3x/7 = 15. So x = 15 * 7 / 3 = 35. So there were 35 stickers.
So the answer is 105.
# Is the above step of reasoning:
# (A) Correct
# (B) Incorrect
# The above step of reasoning is (A), but the value of number of stickers is incorrect.





Q: Alexa has 92 cents in her pocket. She wants to buy 3 pencils at the school supply store. Each pencil costs 8 cents. How much money will Alexa have left?

A:
Alexa has 92 cents.
# Is the above step of reasoning:
# (A) Correct
# (B) Incorrect
# The above step of reasoning is (A)
And 3 pencils for 8 cents each will be 3 * 8 = 24 cents.
# Is the above step of reasoning:
# (A) Correct
# (B) Incorrect
# The above step of reasoning is (A)
So she has 92 - 24 = 68 cents left.
# Is the above step of reasoning:
# (A) Correct
# (B) Incorrect
# The above step of reasoning is (A)
So the answer is 68.
# Is the above step of reasoning:
# (A) Correct
# (B) Incorrect
# The above step of reasoning is (A)






Q: {input}

A: {output}
# Is the above step of reasoning:
# (A) Correct
# (B) Incorrect
# The above step of reasoning is '''

\end{lstlisting}

\subsection{MWIS Prompts}
\label{sec:mwis_prompts}
\begin{lstlisting}[caption={Direct prompts}]
direct_0s = """Given a sequence of integers, find a subsequence with the highest sum, such that no two numbers in the subsequence are adjacent in the original sequence.

To indicate the selected numbers, print an array with "1" for chosen numbers and "2" for unchosen ones. For instance, [1, 2, 2, 2, 2] implies selecting only the first number. If multiple solutions exist, select the lexicographically smallest.


{prompt}
"""

direct_3s = """Given a sequence of integers, find a subsequence with the highest sum, such that no two numbers in the subsequence are adjacent in the original sequence.

To indicate the selected numbers, print an array with "1" for chosen numbers and "2" for unchosen ones. For instance, [1, 2, 2, 2, 2] implies selecting only the first number. If multiple solutions exist, select the lexicographically smallest.


Let\'s solve input = [1, 1, -5, -1].
Answer: [1, 2, 2, 2]


Let\'s solve input = [3, 2, 1, -1, 2].
Answer: [1, 2, 1, 2, 1]


Let\'s solve input = [0, 4, -2, 3, -3, -1].
Answer: [2, 1, 2, 1, 2, 2]


{prompt}
"""

direct_6s = """Given a sequence of integers, find a subsequence with the highest sum, such that no two numbers in the subsequence are adjacent in the original sequence.

To indicate the selected numbers, print an array with "1" for chosen numbers and "2" for unchosen ones. For instance, [1, 2, 2, 2, 2] implies selecting only the first number. If multiple solutions exist, select the lexicographically smallest.


Let\'s solve input = [1, 1, -5, -1].
Answer: [1, 2, 2, 2]


Let\'s solve input = [3, 2, 1, -1, 2].
Answer: [1, 2, 1, 2, 1]


Let\'s solve input = [0, 4, -2, 3, -3, -1].
Answer: [2, 1, 2, 1, 2, 2]


Let\'s solve input = [-3, -4, 4, -1]
Answer: [2, 2, 1, 2]


Let\'s solve input = [3, 4, -3, -1, -4]
Answer: [2, 1, 2, 2, 2]


Let\'s solve input = [-4, 5, 0, 2, 3, -4]
Answer: [2, 1, 2, 2, 1, 2]


{prompt}
"""
\end{lstlisting}
\begin{lstlisting}[caption={CoT Implicit prompts}]
cot_implicit_3s = """Given a sequence of integers, find a subsequence with the highest sum, such that no two numbers in the subsequence are adjacent in the original sequence.

To indicate the selected numbers, print an array with "1" for chosen numbers and "2" for unchosen ones. For instance, [1, 2, 2, 2, 2] implies selecting only the first number. If multiple solutions exist, select the lexicographically smallest.


We will solve any task instance by using dynamic programming. We define dp[i] as the maximum sum of a subsequence that does not include adjacent elements, when considering only the elements of the input from the i-th position onwards.


Let\'s solve input = [1, 1, -5, -1].

dp[3] = max(input[3], 0) = max(-1, 0) = 0
dp[2] = max(input[2], input[3], 0) = max(-5, -1, 0) = 0
dp[1] = max(dp[2], input[1] + dp[3], 0) = max(0, 1 + 0, 0) = 1
dp[0] = max(dp[1], input[0] + dp[2], 0) = max(1, 1 + 0, 0) = 1

Finally, we reconstruct the lexicographically smallest subsequence that fulfills the task objective by selecting numbers as follows. We store the result on a list named "output".

Let can_use_next_item = True.
Since dp[0] == input[0] + dp[2] (1 == 1 + 0) and can_use_next_item == True, we store output[0] = 1. We update can_use_next_item = False.
Since dp[1] != input[1] + dp[3] (1 != 1 + 0) or can_use_next_item == False, we store output[1] = 2. We update can_use_next_item = True.
Since dp[2] != input[2] (0 != -5) or can_use_next_item == False, we store output[2] = 2. We update can_use_next_item = True.
Since dp[3] != input[3] (0 != -1) or can_use_next_item == False, we store output[3] = 2.

Reconstructing all together, output=[1, 2, 2, 2].


Let\'s solve input = [3, 2, 1, -1, 2].

dp[4] = max(input[4], 0) = max(2, 0) = 2
dp[3] = max(input[3], input[4], 0) = max(-1, 2, 0) = 2
dp[2] = max(dp[3], input[2] + dp[4], 0) = max(2, 1 + 2, 0) = 3
dp[1] = max(dp[2], input[1] + dp[3], 0) = max(3, 2 + 2, 0) = 4
dp[0] = max(dp[1], input[0] + dp[2], 0) = max(4, 3 + 3, 0) = 6

Finally, we reconstruct the lexicographically smallest subsequence that fulfills the task objective by selecting numbers as follows. We store the result on a list named "output".

Let can_use_next_item = True.
Since dp[0] == input[0] + dp[2] (6 == 3 + 3) and can_use_next_item == True, we store output[0] = 1. We update can_use_next_item = False.
Since dp[1] != input[1] + dp[3] (4 != 2 + 2) or can_use_next_item == False, we store output[1] = 2. We update can_use_next_item = True.
Since dp[2] == input[2] + dp[4] (3 == 1 + 2) and can_use_next_item == True, we store output[2] = 1. We update can_use_next_item = False.
Since dp[3] != input[3] (2 != -1) or can_use_next_item == False, we store output[3] = 2. We update can_use_next_item = True.
Since dp[4] == input[4] (2 == 2) and can_use_next_item == True, we store output[4] = 1.

Reconstructing all together, output=[1, 2, 1, 2, 1].


Let\'s solve input = [0, 4, -2, 3, -3, -1].

dp[5] = max(input[5], 0) = max(-1, 0) = 0
dp[4] = max(input[4], input[5], 0) = max(-3, -1, 0) = 0
dp[3] = max(dp[4], input[3] + dp[5], 0) = max(0, 3 + 0, 0) = 3
dp[2] = max(dp[3], input[2] + dp[4], 0) = max(3, -2 + 0, 0) = 3
dp[1] = max(dp[2], input[1] + dp[3], 0) = max(3, 4 + 3, 0) = 7
dp[0] = max(dp[1], input[0] + dp[2], 0) = max(7, 0 + 3, 0) = 7

Finally, we reconstruct the lexicographically smallest subsequence that fulfills the task objective by selecting numbers as follows. We store the result on a list named "output".

Let can_use_next_item = True.
Since dp[0] != input[0] + dp[2] (7 != 0 + 3) or can_use_next_item == False, we store output[0] = 2. We update can_use_next_item = True.
Since dp[1] == input[1] + dp[3] (7 == 4 + 3) and can_use_next_item == True, we store output[1] = 1. We update can_use_next_item = False.
Since dp[2] != input[2] + dp[4] (3 != -2 + 0) or can_use_next_item == False, we store output[2] = 2. We update can_use_next_item = True.
Since dp[3] == input[3] + dp[5] (3 == 3 + 0) and can_use_next_item == True, we store output[3] = 1. We update can_use_next_item = False.
Since dp[4] != input[4] (0 != -3) or can_use_next_item == False, we store output[4] = 2. We update can_use_next_item = True.
Since dp[5] != input[5] (0 != -1) or can_use_next_item == False, we store output[5] = 2.

Reconstructing all together, output=[2, 1, 2, 1, 2, 2].


{prompt}
"""

cot_implicit_6s = """Given a sequence of integers, find a subsequence with the highest sum, such that no two numbers in the subsequence are adjacent in the original sequence.

To indicate the selected numbers, print an array with "1" for chosen numbers and "2" for unchosen ones. For instance, [1, 2, 2, 2, 2] implies selecting only the first number. If multiple solutions exist, select the lexicographically smallest.


We will solve any task instance by using dynamic programming. We define dp[i] as the maximum sum of a subsequence that does not include adjacent elements, when considering only the elements of the input from the i-th position onwards.


Let\'s solve input = [1, 1, -5, -1].

dp[3] = max(input[3], 0) = max(-1, 0) = 0
dp[2] = max(input[2], input[3], 0) = max(-5, -1, 0) = 0
dp[1] = max(dp[2], input[1] + dp[3], 0) = max(0, 1 + 0, 0) = 1
dp[0] = max(dp[1], input[0] + dp[2], 0) = max(1, 1 + 0, 0) = 1

Finally, we reconstruct the lexicographically smallest subsequence that fulfills the task objective by selecting numbers as follows. We store the result on a list named "output".

Let can_use_next_item = True.
Since dp[0] == input[0] + dp[2] (1 == 1 + 0) and can_use_next_item == True, we store output[0] = 1. We update can_use_next_item = False.
Since dp[1] != input[1] + dp[3] (1 != 1 + 0) or can_use_next_item == False, we store output[1] = 2. We update can_use_next_item = True.
Since dp[2] != input[2] (0 != -5) or can_use_next_item == False, we store output[2] = 2. We update can_use_next_item = True.
Since dp[3] != input[3] (0 != -1) or can_use_next_item == False, we store output[3] = 2.

Reconstructing all together, output=[1, 2, 2, 2].


Let\'s solve input = [3, 2, 1, -1, 2].

dp[4] = max(input[4], 0) = max(2, 0) = 2
dp[3] = max(input[3], input[4], 0) = max(-1, 2, 0) = 2
dp[2] = max(dp[3], input[2] + dp[4], 0) = max(2, 1 + 2, 0) = 3
dp[1] = max(dp[2], input[1] + dp[3], 0) = max(3, 2 + 2, 0) = 4
dp[0] = max(dp[1], input[0] + dp[2], 0) = max(4, 3 + 3, 0) = 6

Finally, we reconstruct the lexicographically smallest subsequence that fulfills the task objective by selecting numbers as follows. We store the result on a list named "output".

Let can_use_next_item = True.
Since dp[0] == input[0] + dp[2] (6 == 3 + 3) and can_use_next_item == True, we store output[0] = 1. We update can_use_next_item = False.
Since dp[1] != input[1] + dp[3] (4 != 2 + 2) or can_use_next_item == False, we store output[1] = 2. We update can_use_next_item = True.
Since dp[2] == input[2] + dp[4] (3 == 1 + 2) and can_use_next_item == True, we store output[2] = 1. We update can_use_next_item = False.
Since dp[3] != input[3] (2 != -1) or can_use_next_item == False, we store output[3] = 2. We update can_use_next_item = True.
Since dp[4] == input[4] (2 == 2) and can_use_next_item == True, we store output[4] = 1.

Reconstructing all together, output=[1, 2, 1, 2, 1].


Let\'s solve input = [0, 4, -2, 3, -3, -1].

dp[5] = max(input[5], 0) = max(-1, 0) = 0
dp[4] = max(input[4], input[5], 0) = max(-3, -1, 0) = 0
dp[3] = max(dp[4], input[3] + dp[5], 0) = max(0, 3 + 0, 0) = 3
dp[2] = max(dp[3], input[2] + dp[4], 0) = max(3, -2 + 0, 0) = 3
dp[1] = max(dp[2], input[1] + dp[3], 0) = max(3, 4 + 3, 0) = 7
dp[0] = max(dp[1], input[0] + dp[2], 0) = max(7, 0 + 3, 0) = 7

Finally, we reconstruct the lexicographically smallest subsequence that fulfills the task objective by selecting numbers as follows. We store the result on a list named "output".

Let can_use_next_item = True.
Since dp[0] != input[0] + dp[2] (7 != 0 + 3) or can_use_next_item == False, we store output[0] = 2. We update can_use_next_item = True.
Since dp[1] == input[1] + dp[3] (7 == 4 + 3) and can_use_next_item == True, we store output[1] = 1. We update can_use_next_item = False.
Since dp[2] != input[2] + dp[4] (3 != -2 + 0) or can_use_next_item == False, we store output[2] = 2. We update can_use_next_item = True.
Since dp[3] == input[3] + dp[5] (3 == 3 + 0) and can_use_next_item == True, we store output[3] = 1. We update can_use_next_item = False.
Since dp[4] != input[4] (0 != -3) or can_use_next_item == False, we store output[4] = 2. We update can_use_next_item = True.
Since dp[5] != input[5] (0 != -1) or can_use_next_item == False, we store output[5] = 2.

Reconstructing all together, output=[2, 1, 2, 1, 2, 2].


Let\'s solve input = [-3, -4, 4, -1].

dp[3] = max(input[3], 0) = max(-1, 0) = 0
dp[2] = max(input[2], input[3], 0) = max(4, -1, 0) = 4
dp[1] = max(dp[2], input[1] + dp[3], 0) = max(4, -4 + 0, 0) = 4
dp[0] = max(dp[1], input[0] + dp[2], 0) = max(4, -3 + 4, 0) = 4

Finally, we reconstruct the lexicographically smallest subsequence that fulfills the task objective by selecting numbers as follows. We store the result on a list named "output".

Let can_use_next_item = True.
Since dp[0] != input[0] + dp[2] (4 != -3 + 4) or can_use_next_item == False, we store output[0] = 2. We update can_use_next_item = True.
Since dp[1] != input[1] + dp[3] (4 != -4 + 0) or can_use_next_item == False, we store output[1] = 2. We update can_use_next_item = True.
Since dp[2] == input[2] (4 == 4) and can_use_next_item == True, we store output[2] = 1. We update can_use_next_item = False.
Since dp[3] != input[3] (0 != -1) or can_use_next_item == False, we store output[3] = 2.

Reconstructing all together, output=[2, 2, 1, 2].


Let\'s solve input = [3, 4, -3, -1, -4].

dp[4] = max(input[4], 0) = max(-4, 0) = 0
dp[3] = max(input[3], input[4], 0) = max(-1, -4, 0) = 0
dp[2] = max(dp[3], input[2] + dp[4], 0) = max(0, -3 + 0, 0) = 0
dp[1] = max(dp[2], input[1] + dp[3], 0) = max(0, 4 + 0, 0) = 4
dp[0] = max(dp[1], input[0] + dp[2], 0) = max(4, 3 + 0, 0) = 4

Finally, we reconstruct the lexicographically smallest subsequence that fulfills the task objective by selecting numbers as follows. We store the result on a list named "output".

Let can_use_next_item = True.
Since dp[0] != input[0] + dp[2] (4 != 3 + 0) or can_use_next_item == False, we store output[0] = 2. We update can_use_next_item = True.
Since dp[1] == input[1] + dp[3] (4 == 4 + 0) and can_use_next_item == True, we store output[1] = 1. We update can_use_next_item = False.
Since dp[2] != input[2] + dp[4] (0 != -3 + 0) or can_use_next_item == False, we store output[2] = 2. We update can_use_next_item = True.
Since dp[3] != input[3] (0 != -1) or can_use_next_item == False, we store output[3] = 2. We update can_use_next_item = True.
Since dp[4] != input[4] (0 != -4) or can_use_next_item == False, we store output[4] = 2.

Reconstructing all together, output=[2, 1, 2, 2, 2].


Let\'s solve input = [-4, 5, 0, 2, 3, -4].

dp[5] = max(input[5], 0) = max(-4, 0) = 0
dp[4] = max(input[4], input[5], 0) = max(3, -4, 0) = 3
dp[3] = max(dp[4], input[3] + dp[5], 0) = max(3, 2 + 0, 0) = 3
dp[2] = max(dp[3], input[2] + dp[4], 0) = max(3, 0 + 3, 0) = 3
dp[1] = max(dp[2], input[1] + dp[3], 0) = max(3, 5 + 3, 0) = 8
dp[0] = max(dp[1], input[0] + dp[2], 0) = max(8, -4 + 3, 0) = 8

Finally, we reconstruct the lexicographically smallest subsequence that fulfills the task objective by selecting numbers as follows. We store the result on a list named "output".

Let can_use_next_item = True.
Since dp[0] != input[0] + dp[2] (8 != -4 + 3) or can_use_next_item == False, we store output[0] = 2. We update can_use_next_item = True.
Since dp[1] == input[1] + dp[3] (8 == 5 + 3) and can_use_next_item == True, we store output[1] = 1. We update can_use_next_item = False.
Since dp[2] != input[2] + dp[4] (3 != 0 + 3) or can_use_next_item == False, we store output[2] = 2. We update can_use_next_item = True.
Since dp[3] != input[3] + dp[5] (3 != 2 + 0) or can_use_next_item == False, we store output[3] = 2. We update can_use_next_item = True.
Since dp[4] == input[4] (3 == 3) and can_use_next_item == True, we store output[4] = 1. We update can_use_next_item = False.
Since dp[5] != input[5] (0 != -4) or can_use_next_item == False, we store output[5] = 2.

Reconstructing all together, output=[2, 1, 2, 2, 1, 2].


{prompt}
"""
\end{lstlisting}

\begin{lstlisting}[caption={CoT Explicit prompts}]
cot_explicit_3s = """Given a sequence of integers, find a subsequence with the highest sum, such that no two numbers in the subsequence are adjacent in the original sequence.

To indicate the selected numbers, print an array with "1" for chosen numbers and "2" for unchosen ones. For instance, [1, 2, 2, 2, 2] implies selecting only the first number. If multiple solutions exist, select the lexicographically smallest.


We will solve any task instance by using dynamic programming. We define dp[i] as the maximum sum of a subsequence that does not include adjacent elements, when considering only the elements of the input from the i-th position onwards.


Let\'s solve input = [1, 1, -5, -1].

There are 4 numbers in the input sequence, so we will use a list of size 4 to store the dynamic programming values. We initialize all values to 0.
dp[3] = max(input[3], 0) = max(-1, 0) = 0
dp[2] = max(input[2], input[3], 0) = max(-5, -1, 0) = 0
dp[1] = max(dp[2], input[1] + dp[3], 0) = max(0, 1 + 0, 0) = max(0, 1, 0) = 1
dp[0] = max(dp[1], input[0] + dp[2], 0) = max(1, 1 + 0, 0) = max(1, 1, 0) = 1

Finally, we reconstruct the lexicographically smallest subsequence that fulfills the task objective by selecting numbers as follows. We store the result on a list named "output".

Let can_use_next_item = True.
Since dp[0]=1, input[0]=1, dp[2]=0, input[0] + dp[2] = 1 == 1 = dp[0] and can_use_next_item == True, we store output[0] = 1. We update can_use_next_item = False.
Since can_use_next_item == False, we store output[1] = 2. We update can_use_next_item = True.
Since dp[2] = 0, input[2] = -5, dp[2] != input[2], we store output[2] = 2. We update can_use_next_item = True.
Since dp[3] = 0, input[3] = -1, dp[3] != input[3], we store output[3] = 2.

Reconstructing all together, output=[1, 2, 2, 2].



Let\'s solve input = [3, 2, 1, -1, 2].

There are 5 numbers in the input sequence, so we will use a list of size 5 to store the dynamic programming values. We initialize all values to 0.
dp[4] = max(input[4], 0) = max(2, 0) = 2
dp[3] = max(input[3], input[4], 0) = max(-1, 2, 0) = 2
dp[2] = max(dp[3], input[2] + dp[4], 0) = max(2, 1 + 2, 0) = max(2, 3, 0) = 3
dp[1] = max(dp[2], input[1] + dp[3], 0) = max(3, 2 + 2, 0) = max(3, 4, 0) = 4
dp[0] = max(dp[1], input[0] + dp[2], 0) = max(4, 3 + 3, 0) = max(4, 6, 0) = 6

Finally, we reconstruct the lexicographically smallest subsequence that fulfills the task objective by selecting numbers as follows. We store the result on a list named "output".

Let can_use_next_item = True.
Since dp[0]=6, input[0]=3, dp[2]=3, input[0] + dp[2] = 6 == 6 = dp[0] and can_use_next_item == True, we store output[0] = 1. We update can_use_next_item = False.
Since can_use_next_item == False, we store output[1] = 2. We update can_use_next_item = True.
Since dp[2]=3, input[2]=1, dp[4]=2, input[2] + dp[4] = 3 == 3 = dp[2] and can_use_next_item == True, we store output[2] = 1. We update can_use_next_item = False.
Since can_use_next_item == False, we store output[3] = 2. We update can_use_next_item = True.
Since dp[4] = 2, input[4] = 2, dp[4] == input[4] and can_use_next_item == True, we store output[4] = 1.

Reconstructing all together, output=[1, 2, 1, 2, 1].



Let\'s solve input = [0, 4, -2, 3, -3, -1].

There are 6 numbers in the input sequence, so we will use a list of size 6 to store the dynamic programming values. We initialize all values to 0.
dp[5] = max(input[5], 0) = max(-1, 0) = 0
dp[4] = max(input[4], input[5], 0) = max(-3, -1, 0) = 0
dp[3] = max(dp[4], input[3] + dp[5], 0) = max(0, 3 + 0, 0) = max(0, 3, 0) = 3
dp[2] = max(dp[3], input[2] + dp[4], 0) = max(3, -2 + 0, 0) = max(3, -2, 0) = 3
dp[1] = max(dp[2], input[1] + dp[3], 0) = max(3, 4 + 3, 0) = max(3, 7, 0) = 7
dp[0] = max(dp[1], input[0] + dp[2], 0) = max(7, 0 + 3, 0) = max(7, 3, 0) = 7

Finally, we reconstruct the lexicographically smallest subsequence that fulfills the task objective by selecting numbers as follows. We store the result on a list named "output".

Let can_use_next_item = True.
Since dp[0]=7, input[0]=0, dp[2]=3, input[0] + dp[2] = 3 != 7 = dp[0], we store output[0] = 2. We update can_use_next_item = True.
Since dp[1]=7, input[1]=4, dp[3]=3, input[1] + dp[3] = 7 == 7 = dp[1] and can_use_next_item == True, we store output[1] = 1. We update can_use_next_item = False.
Since can_use_next_item == False, we store output[2] = 2. We update can_use_next_item = True.
Since dp[3]=3, input[3]=3, dp[5]=0, input[3] + dp[5] = 3 == 3 = dp[3] and can_use_next_item == True, we store output[3] = 1. We update can_use_next_item = False.
Since can_use_next_item == False, we store output[4] = 2. We update can_use_next_item = True.
Since dp[5] = 0, input[5] = -1, dp[5] != input[5], we store output[5] = 2.

Reconstructing all together, output=[2, 1, 2, 1, 2, 2].


{prompt}
"""
cot_explicit_6s = """Given a sequence of integers, find a subsequence with the highest sum, such that no two numbers in the subsequence are adjacent in the original sequence.

To indicate the selected numbers, print an array with "1" for chosen numbers and "2" for unchosen ones. For instance, [1, 2, 2, 2, 2] implies selecting only the first number. If multiple solutions exist, select the lexicographically smallest.


We will solve any task instance by using dynamic programming. We define dp[i] as the maximum sum of a subsequence that does not include adjacent elements, when considering only the elements of the input from the i-th position onwards.


Let\'s solve input = [1, 1, -5, -1].

There are 4 numbers in the input sequence, so we will use a list of size 4 to store the dynamic programming values. We initialize all values to 0.
dp[3] = max(input[3], 0) = max(-1, 0) = 0
dp[2] = max(input[2], input[3], 0) = max(-5, -1, 0) = 0
dp[1] = max(dp[2], input[1] + dp[3], 0) = max(0, 1 + 0, 0) = max(0, 1, 0) = 1
dp[0] = max(dp[1], input[0] + dp[2], 0) = max(1, 1 + 0, 0) = max(1, 1, 0) = 1

Finally, we reconstruct the lexicographically smallest subsequence that fulfills the task objective by selecting numbers as follows. We store the result on a list named "output".

Let can_use_next_item = True.
Since dp[0]=1, input[0]=1, dp[2]=0, input[0] + dp[2] = 1 == 1 = dp[0] and can_use_next_item == True, we store output[0] = 1. We update can_use_next_item = False.
Since can_use_next_item == False, we store output[1] = 2. We update can_use_next_item = True.
Since dp[2] = 0, input[2] = -5, dp[2] != input[2], we store output[2] = 2. We update can_use_next_item = True.
Since dp[3] = 0, input[3] = -1, dp[3] != input[3], we store output[3] = 2.

Reconstructing all together, output=[1, 2, 2, 2].



Let\'s solve input = [3, 2, 1, -1, 2].

There are 5 numbers in the input sequence, so we will use a list of size 5 to store the dynamic programming values. We initialize all values to 0.
dp[4] = max(input[4], 0) = max(2, 0) = 2
dp[3] = max(input[3], input[4], 0) = max(-1, 2, 0) = 2
dp[2] = max(dp[3], input[2] + dp[4], 0) = max(2, 1 + 2, 0) = max(2, 3, 0) = 3
dp[1] = max(dp[2], input[1] + dp[3], 0) = max(3, 2 + 2, 0) = max(3, 4, 0) = 4
dp[0] = max(dp[1], input[0] + dp[2], 0) = max(4, 3 + 3, 0) = max(4, 6, 0) = 6

Finally, we reconstruct the lexicographically smallest subsequence that fulfills the task objective by selecting numbers as follows. We store the result on a list named "output".

Let can_use_next_item = True.
Since dp[0]=6, input[0]=3, dp[2]=3, input[0] + dp[2] = 6 == 6 = dp[0] and can_use_next_item == True, we store output[0] = 1. We update can_use_next_item = False.
Since can_use_next_item == False, we store output[1] = 2. We update can_use_next_item = True.
Since dp[2]=3, input[2]=1, dp[4]=2, input[2] + dp[4] = 3 == 3 = dp[2] and can_use_next_item == True, we store output[2] = 1. We update can_use_next_item = False.
Since can_use_next_item == False, we store output[3] = 2. We update can_use_next_item = True.
Since dp[4] = 2, input[4] = 2, dp[4] == input[4] and can_use_next_item == True, we store output[4] = 1.

Reconstructing all together, output=[1, 2, 1, 2, 1].



Let\'s solve input = [0, 4, -2, 3, -3, -1].

There are 6 numbers in the input sequence, so we will use a list of size 6 to store the dynamic programming values. We initialize all values to 0.
dp[5] = max(input[5], 0) = max(-1, 0) = 0
dp[4] = max(input[4], input[5], 0) = max(-3, -1, 0) = 0
dp[3] = max(dp[4], input[3] + dp[5], 0) = max(0, 3 + 0, 0) = max(0, 3, 0) = 3
dp[2] = max(dp[3], input[2] + dp[4], 0) = max(3, -2 + 0, 0) = max(3, -2, 0) = 3
dp[1] = max(dp[2], input[1] + dp[3], 0) = max(3, 4 + 3, 0) = max(3, 7, 0) = 7
dp[0] = max(dp[1], input[0] + dp[2], 0) = max(7, 0 + 3, 0) = max(7, 3, 0) = 7

Finally, we reconstruct the lexicographically smallest subsequence that fulfills the task objective by selecting numbers as follows. We store the result on a list named "output".

Let can_use_next_item = True.
Since dp[0]=7, input[0]=0, dp[2]=3, input[0] + dp[2] = 3 != 7 = dp[0], we store output[0] = 2. We update can_use_next_item = True.
Since dp[1]=7, input[1]=4, dp[3]=3, input[1] + dp[3] = 7 == 7 = dp[1] and can_use_next_item == True, we store output[1] = 1. We update can_use_next_item = False.
Since can_use_next_item == False, we store output[2] = 2. We update can_use_next_item = True.
Since dp[3]=3, input[3]=3, dp[5]=0, input[3] + dp[5] = 3 == 3 = dp[3] and can_use_next_item == True, we store output[3] = 1. We update can_use_next_item = False.
Since can_use_next_item == False, we store output[4] = 2. We update can_use_next_item = True.
Since dp[5] = 0, input[5] = -1, dp[5] != input[5], we store output[5] = 2.

Reconstructing all together, output=[2, 1, 2, 1, 2, 2].



Let\'s solve input = [-3, -4, 4, -1].

There are 4 numbers in the input sequence, so we will use a list of size 4 to store the dynamic programming values. We initialize all values to 0.
dp[3] = max(input[3], 0) = max(-1, 0) = 0
dp[2] = max(input[2], input[3], 0) = max(4, -1, 0) = 4
dp[1] = max(dp[2], input[1] + dp[3], 0) = max(4, -4 + 0, 0) = max(4, -4, 0) = 4
dp[0] = max(dp[1], input[0] + dp[2], 0) = max(4, -3 + 4, 0) = max(4, 1, 0) = 4

Finally, we reconstruct the lexicographically smallest subsequence that fulfills the task objective by selecting numbers as follows. We store the result on a list named "output".

Let can_use_next_item = True.
Since dp[0]=4, input[0]=-3, dp[2]=4, input[0] + dp[2] = 1 != 4 = dp[0], we store output[0] = 2. We update can_use_next_item = True.
Since dp[1]=4, input[1]=-4, dp[3]=0, input[1] + dp[3] = -4 != 4 = dp[1], we store output[1] = 2. We update can_use_next_item = True.
Since dp[2] = 4, input[2] = 4, dp[2] == input[2] and can_use_next_item == True, we store output[2] = 1. We update can_use_next_item = False.
Since can_use_next_item == False, we store output[3] = 2.

Reconstructing all together, output=[2, 2, 1, 2].



Let\'s solve input = [3, 4, -3, -1, -4].

There are 5 numbers in the input sequence, so we will use a list of size 5 to store the dynamic programming values. We initialize all values to 0.
dp[4] = max(input[4], 0) = max(-4, 0) = 0
dp[3] = max(input[3], input[4], 0) = max(-1, -4, 0) = 0
dp[2] = max(dp[3], input[2] + dp[4], 0) = max(0, -3 + 0, 0) = max(0, -3, 0) = 0
dp[1] = max(dp[2], input[1] + dp[3], 0) = max(0, 4 + 0, 0) = max(0, 4, 0) = 4
dp[0] = max(dp[1], input[0] + dp[2], 0) = max(4, 3 + 0, 0) = max(4, 3, 0) = 4

Finally, we reconstruct the lexicographically smallest subsequence that fulfills the task objective by selecting numbers as follows. We store the result on a list named "output".

Let can_use_next_item = True.
Since dp[0]=4, input[0]=3, dp[2]=0, input[0] + dp[2] = 3 != 4 = dp[0], we store output[0] = 2. We update can_use_next_item = True.
Since dp[1]=4, input[1]=4, dp[3]=0, input[1] + dp[3] = 4 == 4 = dp[1] and can_use_next_item == True, we store output[1] = 1. We update can_use_next_item = False.
Since can_use_next_item == False, we store output[2] = 2. We update can_use_next_item = True.
Since dp[3] = 0, input[3] = -1, dp[3] != input[3], we store output[3] = 2. We update can_use_next_item = True.
Since dp[4] = 0, input[4] = -4, dp[4] != input[4], we store output[4] = 2.

Reconstructing all together, output=[2, 1, 2, 2, 2].



Let\'s solve input = [-4, 5, 0, 2, 3, -4].

There are 6 numbers in the input sequence, so we will use a list of size 6 to store the dynamic programming values. We initialize all values to 0.
dp[5] = max(input[5], 0) = max(-4, 0) = 0
dp[4] = max(input[4], input[5], 0) = max(3, -4, 0) = 3
dp[3] = max(dp[4], input[3] + dp[5], 0) = max(3, 2 + 0, 0) = max(3, 2, 0) = 3
dp[2] = max(dp[3], input[2] + dp[4], 0) = max(3, 0 + 3, 0) = max(3, 3, 0) = 3
dp[1] = max(dp[2], input[1] + dp[3], 0) = max(3, 5 + 3, 0) = max(3, 8, 0) = 8
dp[0] = max(dp[1], input[0] + dp[2], 0) = max(8, -4 + 3, 0) = max(8, -1, 0) = 8

Finally, we reconstruct the lexicographically smallest subsequence that fulfills the task objective by selecting numbers as follows. We store the result on a list named "output".

Let can_use_next_item = True.
Since dp[0]=8, input[0]=-4, dp[2]=3, input[0] + dp[2] = -1 != 8 = dp[0], we store output[0] = 2. We update can_use_next_item = True.
Since dp[1]=8, input[1]=5, dp[3]=3, input[1] + dp[3] = 8 == 8 = dp[1] and can_use_next_item == True, we store output[1] = 1. We update can_use_next_item = False.
Since can_use_next_item == False, we store output[2] = 2. We update can_use_next_item = True.
Since dp[3]=3, input[3]=2, dp[5]=0, input[3] + dp[5] = 2 != 3 = dp[3], we store output[3] = 2. We update can_use_next_item = True.
Since dp[4] = 3, input[4] = 3, dp[4] == input[4] and can_use_next_item == True, we store output[4] = 1. We update can_use_next_item = False.
Since can_use_next_item == False, we store output[5] = 2.

Reconstructing all together, output=[2, 1, 2, 2, 1, 2].


{prompt}
"""
\end{lstlisting}
\subsection{Comparison between CoT Implicit and CoT Explicit}
\label{sec:implicit_vs_explicit}
\begin{lstlisting}[caption={Compare CoT Implicit with CoT Explicit}]
Given a sequence of integers, find a subsequence with the highest sum, such that no two numbers in the subsequence are adjacent in the original sequence.

To indicate the selected numbers, print an array with "1" for chosen numbers and "2" for unchosen ones. For instance, [1, 2, 2, 2, 2] implies selecting only the first number. If multiple solutions exist, select the lexicographically smallest.


We will solve any task instance by using dynamic programming. We define dp[i] as the maximum sum of a subsequence that does not include adjacent elements, when considering only the elements of the input from the i-th position onwards.

Let's solve input = [-45, 40, 77, -93, 25, -73].

###
BLUEBLUEExplicit: There are 6 numbers in the input sequence, so we will use a list of size 6 to store the dynamic programming values. We initialize all values to 0.BLUEBLUE
dp[5] = max(input[5], 0) = max(-73, 0) = 0
dp[4] = max(input[4], input[5], 0) = max(25, -73, 0) = 25
REDREDImplicit: dp[3] = max(dp[4], input[3] + dp[5], 0) = max(25, -93 + 0, 0) = 25REDRED
BLUEBLUEExplicit: dp[3] = max(dp[4], input[3] + dp[5], 0) = max(25, -93 + 0, 0) = max(25, -93, 0) = 25BLUEBLUE
REDREDImplicit: dp[2] = max(dp[3], input[2] + dp[4], 0) = max(25, 77 + 25, 0) = 102REDRED
BLUEBLUEExplicit: dp[2] = max(dp[3], input[2] + dp[4], 0) = max(25, 77 + 25, 0) = max(25, 102, 0) = 102BLUEBLUE
REDREDImplicit: dp[1] = max(dp[2], input[1] + dp[3], 0) = max(102, 40 + 25, 0) = 102REDRED
BLUEBLUEExplicit: dp[1] = max(dp[2], input[1] + dp[3], 0) = max(102, 40 + 25, 0) = max(102, 65, 0) = 102BLUEBLUE
REDREDImplicit: dp[0] = max(dp[1], input[0] + dp[2], 0) = max(102, -45 + 102, 0) = 102REDRED
BLUEBLUEExplicit: dp[0] = max(dp[1], input[0] + dp[2], 0) = max(102, -45 + 102, 0) = max(102, 57, 0) = 102BLUEBLUE

Finally, we reconstruct the lexicographically smallest subsequence that fulfills the task objective by selecting numbers as follows. We store the result on a list named "output".

Let can_use_next_item = True.
REDREDImplicit: Since dp[0] != input[0] + dp[2] (102 != -45 + 102) or can_use_next_item == False, we store output[0] = 2. We update can_use_next_item = True.REDRED
BLUEBLUEExplicit: Since dp[0]=102, input[0]=-45, dp[2]=102, input[0] + dp[2] = 57 != 102 = dp[0], we store output[0] = 2. We update can_use_next_item = True.BLUEBLUE
REDREDImplicit: Since dp[1] != input[1] + dp[3] (102 != 40 + 25) or can_use_next_item == False, we store output[1] = 2. We update can_use_next_item = True.REDRED
BLUEBLUEExplicit: Since dp[1]=102, input[1]=40, dp[3]=25, input[1] + dp[3] = 65 != 102 = dp[1], we store output[1] = 2. We update can_use_next_item = True.BLUEBLUE
REDREDImplicit: Since dp[2] == input[2] + dp[4] (102 == 77 + 25) and can_use_next_item == True, we store output[2] = 1. We update can_use_next_item = False.REDRED
BLUEBLUEExplicit: Since dp[2]=102, input[2]=77, dp[4]=25, input[2] + dp[4] = 102 == 102 = dp[2] and can_use_next_item == True, we store output[2] = 1. We update can_use_next_item = False.BLUEBLUE
REDREDImplicit: Since dp[3] != input[3] + dp[5] (25 != -93 + 0) or can_use_next_item == False, we store output[3] = 2. We update can_use_next_item = True.REDRED
BLUEBLUEExplicit: Since can_use_next_item == False, we store output[3] = 2. We update can_use_next_item = True.BLUEBLUE
REDREDImplicit: Since dp[4] == input[4] (25 == 25) and can_use_next_item == True, we store output[4] = 1. We update can_use_next_item = False.REDRED
BLUEBLUEExplicit: Since dp[4] = 25, input[4] = 25, dp[4] == input[4] and can_use_next_item == True, we store output[4] = 1. We update can_use_next_item = False.BLUEBLUE
REDREDImplicit: Since dp[5] != input[5] (0 != -73) or can_use_next_item == False, we store output[5] = 2.REDRED
BLUEBLUEExplicit: Since can_use_next_item == False, we store output[5] = 2.BLUEBLUE

Reconstructing all together, output=[2, 2, 1, 2, 1, 2].

\end{lstlisting}

\subsection{Travel planning prompts}\label{app:travel}

\begin{lstlisting}[caption=CoT prompts]
prompt_cot_zero_shot = """
The user will ask for a flight route between two cities. You need to generate a response with the route. Your response should be in the format "[city 1]-[city 2]-[city 3]-...-[city n]". If there is no solution, reply "Answer: None. "
Question: {input}
Answer: """

prompt_cot_1s = """
The user will ask for a flight route between two cities. You need to generate a response with the route. Your response should be in the format "Answer: [city 1]-[city 2]-[city 3]-...-[city n]". If there is no solution, reply "Answer: None. "
Question: What is the flight route from Dublin to Sydney?
Answer: Dublin-London-Sydney.
Question: {input}
Answer: """

prompt_cot_3s = """
The user will ask for a flight route between two cities. You need to generate a response with the route. Your response should be in the format "[city 1]-[city 2]-[city 3]-...-[city n]". If there is no solution, reply "Answer: None. "
Question: What is the flight route from Dublin to Sydney?
Answer: Dublin-London-Sydney.
Question: What is the flight route from New York to Amsterdam?
Answer: New York-London-Amsterdam.
Question: What is the flight route from Toronto to Sydney?
Answer: Toronto-San Francisco-Sydney.
Question: {input}
Answer: """

prompt_cot_8s = """
The user will ask for a flight route between two cities. You need to generate a response with the route. Your response should be in the format "[city 1]-[city 2]-[city 3]-...-[city n]". If there is no solution, reply "Answer: None. "
Question: What is the flight route from Dublin to Sydney?
Answer: Dublin-London-Sydney.
Question: What is the flight route from New York to Amsterdam?
Answer: New York-London-Amsterdam.
Question: What is the flight route from Toronto to Sydney?
Answer: Toronto-San Francisco-Sydney.
Question: What is the flight route from Astana to Rome?
Answer: Astana-Moscow-Rome.
Question: What is the flight route from Visakhapatnam to Odense?
Answer: Visakhapatnam-Hyderabad-Copenhagen-Odense.
Question: What is the flight route from Shanghai to Nanjing?
Answer: Shanghai-Nanjing.
Question: What is the flight route from Singapore to Taipei?
Answer: Singapore-Taipei.
Question: What is the flight route from Sydney to Istanbul?
Answer: Sydney-Singapore-Istanbul.
Question: {input}
Answer: """

\end{lstlisting}

\begin{lstlisting}[caption=ToT prompts]
prompt_tot_propose_zero_shot  = '''List a few possible cities to fly to from the current city via one direct flight. If the goal city can be reached via one direct flight from the current city, just answer the goal city. Format of your response is "Answer: [city 1], [city 2], [city 3], ... [city n]."
Question: {input}
'''

prompt_tot_propose_1s = '''List the a few possible cities to fly to from the current city via one direct flight. If the goal city can be reached via one direct flight from the current city, just answer the goal city. Format of your response is "Answer: [city 1], [city 2], [city 3], ... [city n]."
Question: You want to go to Sydney and you are at Dublin. Propose a few possible cities with direct flights to go to for the next step. 
Answer: London, Paris, Frankfurt, Amsterdam, Zurich.
Question: {input}
'''

prompt_tot_propose_3s = '''List the a few possible cities to fly to from the current city via one direct flight. If the goal city can be reached via one direct flight from the current city, just answer the goal city. Format of your response is "Answer: [city 1], [city 2], [city 3], ... [city n]."
Question: You want to go to Sydney and you are at Dublin. Propose a few possible cities with direct flights to go to for the next step. 
Answer: London, Paris, Mombai.
Question: You want to go to Nanjing and you are at Shanghai. Propose a few possible cities with direct flights to go to for the next step.
Answer: Nanjing.
Question: You want to go to Amsterdam and you are at New York. Propose a few possible cities with direct flights to go to for the next step. 
Answer: London, Paris, Frankfurt, Amsterdam.
Question: {input}
'''

prompt_tot_propose_8s = '''List the a few possible cities to fly to from the current city via one direct flight. If the goal city can be reached via one direct flight from the current city, just answer the goal city. Format of your response is "Answer: [city 1], [city 2], [city 3], ... [city n]."
Question: You want to go to Sydney and you are at Dublin. Propose a few possible cities with direct flights to go to for the next step. 
Answer: London, Paris, Mombai.
Question: You want to go to Amsterdam and you are at New York. Propose a few possible cities with direct flights to go to for the next step. 
Answer: London, Paris, Frankfurt.
Question: You want to go to Sydney and you are at Toronto. Propose a few possible cities with direct flights to go to for the next step.
Answer: San Francisco, Los Angeles, Vancouver.
Question: You want to go to Nanjing and you are at Shanghai. Propose a few possible cities with direct flights to go to for the next step.
Answer: Nanjing.
Question: You want to go to Rome and you are at Astana. Propose a few possible cities with direct flights to go to for the next step.
Answer: Moscow, Rome, Istanbul.
Question: You want to go to Odense and you are at Visakhapatnam. Propose a few possible cities with direct flights to go to for the next step.
Answer: Hyderabad, Copenhagen, Odense.
Question: You want to go to Taipei and you are at Singapore. Propose a few possible cities with direct flights to go to for the next step.
Answer: Taipei.
Question: You want to go to Istanbul and you are at Sydney. Propose a few possible cities with direct flights to go to for the next step.
Answer: Singapore, Dubai, Abu Dhabi.
Question: {input}
'''
\end{lstlisting}

\begin{lstlisting}[caption=ToT Linear prompts]
prompt_tot_linear_zero_shot = """The user will ask for a flight route between two cities. You need to generate a response with the route. 
You are simulating bfs process to find the route between two cities. In the beginning, you have a queue ['start city'] and an empty explored list []. You need to proceed with the following steps:
1. Take the first city in the queue as the current city. If the city is in the explored list, skip it. Otherwise, put the city into the explored list.
2. Propose the possible cities with direct flights to go to for the next step. Do not propose the explored cities and cities in the queue.
3. Put the cities into the queue.
Repeat steps 1-3 until the goal city is included in the queue.
Respond with reasoning steps, and end with the answer, in the format "Answer: [city 1]-[city 2]-[city 3]-...-[city n]"
Question: {input}
Let's think step by step.
"""


prompt_tot_linear_cot_1s = """The user will ask for a flight route between two cities. You need to generate a response with the route. 
You are simulating bfs process to find the route between two cities. In the beginning, you have a queue ['start city'], and you need to proceed the following steps:
1. Take the first city in the queue as the current city.
2. Propose the possible cities with direct flights to go to for the next step. Do not propose the explored cities and cities in the queue.
3. Put the cities into the queue.
Repeat steps 1-3 until the goal city is included in the queue.
Respond with reasoning steps, and end with the answer, in the format "Answer: [city 1]-[city 2]-[city 3]-...-[city n]"
Question: What is the flight route from Guatemala City to Guangzhou?
The queue is [Guatemala City]. Take the first path, Guatemala City, from the queue.
The current city is Guatemala City, which is not in the explored list. Thus, put the current city into the explored list. The explored list is [Guatemala City]
The current city is Guatemala City and the goal is Guangzhou. For the next step, the promising cities to go to are [New York, Los Angeles, Mexico City].
Puting those cities into the queue. The queue is [Guatemala City-New York, Guatemala City-Los Angeles, Guatemala City-Mexico City].
Take the first path, Guatemala City-New York, from the queue.
The current city is New York, which is not in the explored list. Thus, put the current city into the explored list. The explored list is [Guatemala City, New York]
The current city is New York and the goal is Guangzhou. For the next step, the promising cities to go to are [Helsinki, Guangzhou, Lahore].
The goal city is Guangzhou. Since Guangzhou is in the found, and the current selected path is Guatemala City-New York, the route is Guatemala City-New York-Guangzhou.
Answer: Guatemala City-New York-Guangzhou
Question: {input}
Let's think step by step.
"""

prompt_tot_linear_cot_2s = """The user will ask for a flight route between two cities. You need to generate a response with the route. 
You are simulating bfs process to find the route between two cities. In the beginning, you have a queue ['start city'], and you need to proceed the following steps:
1. Take the first city in the queue as the current city.
2. Propose the possible cities with direct flights to go to for the next step. Do not propose the explored cities and cities in the queue.
3. Put the cities into the queue.
Repeat steps 1-3 until the goal city is included in the queue.
Respond with reasoning steps, and end with the answer, in the format "Answer: [city 1]-[city 2]-[city 3]-...-[city n]"
Question: What is the flight route from Guatemala City to Guangzhou?
The queue is [Guatemala City]. Take the first path, Guatemala City, from the queue.
The current city is Guatemala City, which is not in the explored list. Thus, put the current city into the explored list. The explored list is [Guatemala City]
The current city is Guatemala City and the goal is Guangzhou. For the next step, the promising cities to go to are [New York, Los Angeles, Mexico City].
Puting those cities into the queue. The queue is [Guatemala City-New York, Guatemala City-Los Angeles, Guatemala City-Mexico City].
Take the first path, Guatemala City-New York, from the queue.
The current city is New York, which is not in the explored list. Thus, put the current city into the explored list. The explored list is [Guatemala City, New York]
The current city is New York and the goal is Guangzhou. For the next step, the promising cities to go to are [Helsinki, Guangzhou, Lahore].
The goal city is Guangzhou. Since Guangzhou is in the found, and the current selected path is Guatemala City-New York, the route is Guatemala City-New York-Guangzhou.
Answer: Guatemala City-New York-Guangzhou
Question: What is the flight route from Tegucigalpa to Helsinki?
The queue is [Tegucigalpa]. Take the first path, Tegucigalpa, from the queue.
The current city is Tegucigalpa, which is not in the explored list. Thus, put the current city into the explored list. The explored list is [Tegucigalpa]
The current city is Tegucigalpa and the goal is Helsinki. For the next step, the promising cities to go to are [Guatemala City, Miami].
Puting those cities into the queue. The queue is [Tegucigalpa-Guatemala City, Tegucigalpa-Miami].
Take the first path, Tegucigalpa-Guatemala City, from the queue.
The current city is Guatemala City, which is not in the explored list. Thus, put the current city into the explored list. The explored list is [Tegucigalpa, Guatemala City]
The current city is Guatemala City and the goal is Helsinki. For the next step, the promising cities to go to are [New York, Los Angeles, Mexico City].
Puting those cities into the queue. The queue is [Tegucigalpa-Miami, Tegucigalpa-Guatemala City-New York, Tegucigalpa-Guatemala City-Los Angeles, Tegucigalpa-Guatemala City-Mexico City].
Take the first path, Tegucigalpa-Miami, from the queue.
The current city is Miami, which is not in the explored list. Thus, put the current city into the explored list. The explored list is [Tegucigalpa, Guatemala City, Miami]
The current city is Miami and the goal is Helsinki. For the next step, the promising cities to go to are [Sao Paulo, Buenos Aires, Chicago].
Puting those cities into the queue. The queue is [Tegucigalpa-Guatemala City-New York, Tegucigalpa-Guatemala City-Los Angeles, Tegucigalpa-Guatemala City-Mexico City, Tegucigalpa-Miami-Sao Paulo, Tegucigalpa-Miami-Buenos Aires, Tegucigalpa-Miami-Chicago].
Take the first path, Tegucigalpa-Guatemala City-New York, from the queue.
The current city is New York, which is not in the explored list. Thus, put the current city into the explored list. The explored list is [Tegucigalpa, Guatemala City, Miami, New York]
The current city is New York and the goal is Helsinki. For the next step, the promising cities to go to are [Helsinki, Guangzhou, Lahore].
The goal city is Helsinki. Since Helsinki is in the found, and the current selected path is Tegucigalpa-Guatemala City-New York, the route is Tegucigalpa-Guatemala City-New York-Helsinki.
Answer: Tegucigalpa-Guatemala City-New York-Helsinki
Question: {input}
Let's think step by step.
"""

\end{lstlisting}

\subsection{Game of 24 prompts}\label{app:game24}

\begin{lstlisting}[caption=CoT prompts]
cot_prompt_1s = '''Use numbers and basic arithmetic operations (+ - * /) to obtain 24. Each step, you are only allowed to choose two of the remaining numbers to obtain a new number.
Please strictly follow the format of the example. Do not include unnecessary information in your output. Do not include serial numbers that are not in the example.
Input: 4 4 6 8
Steps:
4 + 8 = 12 (left: 4 6 12)
6 - 4 = 2 (left: 2 12)
2 * 12 = 24 (left: 24)
Answer: (6 - 4) * (4 + 8) = 24
Input: {input}
'''

cot_prompt_3s = '''Use numbers and basic arithmetic operations (+ - * /) to obtain 24. Each step, you are only allowed to choose two of the remaining numbers to obtain a new number.
Please strictly follow the format of the example. Do not include unnecessary information in your output. Do not include serial numbers that are not in the example.
Input: 4 4 6 8
Steps:
4 + 8 = 12 (left: 4 6 12)
6 - 4 = 2 (left: 2 12)
2 * 12 = 24 (left: 24)
Answer: (6 - 4) * (4 + 8) = 24
Input: 2 9 10 12
Steps:
12 * 2 = 24 (left: 9 10 24)
10 - 9 = 1 (left: 1 24)
24 * 1 = 24 (left: 24)
Answer: (12 * 2) * (10 - 9) = 24
Input: 4 9 10 13
Steps:
13 - 10 = 3 (left: 3 4 9)
9 - 3 = 6 (left: 4 6)
4 * 6 = 24 (left: 24)
Answer: 4 * (9 - (13 - 10)) = 24
Input: {input}
'''

cot_prompt_5s = '''Use numbers and basic arithmetic operations (+ - * /) to obtain 24. Each step, you are only allowed to choose two of the remaining numbers to obtain a new number.
Please strictly follow the format of the example. Do not include unnecessary information in your output. Do not include serial numbers that are not in the example.
Input: 4 4 6 8
Steps:
4 + 8 = 12 (left: 4 6 12)
6 - 4 = 2 (left: 2 12)
2 * 12 = 24 (left: 24)
Answer: (6 - 4) * (4 + 8) = 24
Input: 2 9 10 12
Steps:
12 * 2 = 24 (left: 9 10 24)
10 - 9 = 1 (left: 1 24)
24 * 1 = 24 (left: 24)
Answer: (12 * 2) * (10 - 9) = 24
Input: 4 9 10 13
Steps:
13 - 10 = 3 (left: 3 4 9)
9 - 3 = 6 (left: 4 6)
4 * 6 = 24 (left: 24)
Answer: 4 * (9 - (13 - 10)) = 24
Input: 1 4 8 8
Steps:
8 / 4 = 2 (left: 1 2 8)
1 + 2 = 3 (left: 3 8)
3 * 8 = 24 (left: 24)
Answer: (1 + 8 / 4) * 8 = 24
Input: 5 5 5 9
Steps:
5 + 5 = 10 (left: 5 9 10)
10 + 5 = 15 (left: 9 15)
15 + 9 = 24 (left: 24)
Answer: ((5 + 5) + 5) + 9 = 24
Input: {input}
'''
\end{lstlisting}

\begin{lstlisting}[caption=ToT prompts]
propose_prompt_1s = '''Use numbers and basic arithmetic operations (+ - * /) to propose possible next steps of operation. Each step, you are only allowed to choose two of the input numbers to obtain a new number.
Do not include serial numbers that are not in the example. Do not include unnecessary information in your output.
Input: 2 8 8 14
Possible next steps:
2 + 8 = 10 (left: 8 10 14)
8 / 2 = 4 (left: 4 8 14)
14 + 2 = 16 (left: 8 8 16)
2 * 8 = 16 (left: 8 14 16)
8 - 2 = 6 (left: 6 8 14)
14 - 8 = 6 (left: 2 6 8)
14 /  2 = 7 (left: 7 8 8)
14 - 2 = 12 (left: 8 8 12)
Input: {input}
Possible next steps:
'''

propose_prompt_3s = '''Use numbers and basic arithmetic operations (+ - * /) to propose possible next steps of operation. Each step, you are only allowed to choose two of the input numbers to obtain a new number.
Do not include serial numbers that are not in the example. Do not include unnecessary information in your output.
Input: 2 8 8 14
Possible next steps:
2 + 8 = 10 (left: 8 10 14)
8 / 2 = 4 (left: 4 8 14)
14 + 2 = 16 (left: 8 8 16)
2 * 8 = 16 (left: 8 14 16)
8 - 2 = 6 (left: 6 8 14)
14 - 8 = 6 (left: 2 6 8)
14 /  2 = 7 (left: 7 8 8)
14 - 2 = 12 (left: 8 8 12)
Input: 1 2 7 10
Possible next steps:
1 + 2 = 3 (left: 3 7 10)
2 + 7 = 9 (left: 1 9 10)
7 + 10 = 17 (left: 1 2 17)
1 * 2 = 2 (left: 2 7 10)
2 * 7 = 14 (left: 1 14 10)
7 * 10 = 70 (left: 1 2 70)
1 - 2 = -1 (left: -1 7 10)
2 - 7 = -5 (left: 1 -5 10)
7 - 10 = -3 (left: 1 2 -3)
1 / 2 = 0.5 (left: 0.5 7 10)
2 / 7 = 0.29 (left: 1 0.29 10)
7 / 10 = 0.7 (left: 1 2 0.7)
Input: 4 4 6 8
Possible next steps:
4 + 4 = 8 (left: 6 8 8)
4 + 6 = 10 (left: 8 10 8)
6 + 8 = 14 (left: 4 14 8)
4 * 4 = 16 (left: 6 8 16)
4 * 6 = 24 (left: 8 24 8)
6 * 8 = 48 (left: 4 48 8)
4 - 4 = 0 (left: 0 6 8)
4 - 6 = -2 (left: -2 8 8)
6 - 8 = -2 (left: 4 -2 8)
4 / 4 = 1 (left: 1 6 8)
4 / 6 = 0.67 (left: 8 0.67 8)
6 / 8 = 0.75 (left: 4 0.75 8)
Input: {input}
Possible next steps:
'''

propose_prompt_5s = '''Use numbers and basic arithmetic operations (+ - * /) to propose possible next steps of operation. Each step, you are only allowed to choose two of the input numbers to obtain a new number.
Do not include serial numbers that are not in the example. Do not include unnecessary information in your output.
Input: 2 8 8 14
Possible next steps:
2 + 8 = 10 (left: 8 10 14)
8 / 2 = 4 (left: 4 8 14)
14 + 2 = 16 (left: 8 8 16)
2 * 8 = 16 (left: 8 14 16)
8 - 2 = 6 (left: 6 8 14)
14 - 8 = 6 (left: 2 6 8)
14 /  2 = 7 (left: 7 8 8)
14 - 2 = 12 (left: 8 8 12)
Input: 1 2 7 10
Possible next steps:
1 + 2 = 3 (left: 3 7 10)
2 + 7 = 9 (left: 1 9 10)
7 + 10 = 17 (left: 1 2 17)
1 * 2 = 2 (left: 2 7 10)
2 * 7 = 14 (left: 1 14 10)
7 * 10 = 70 (left: 1 2 70)
1 - 2 = -1 (left: -1 7 10)
2 - 7 = -5 (left: 1 -5 10)
7 - 10 = -3 (left: 1 2 -3)
1 / 2 = 0.5 (left: 0.5 7 10)
2 / 7 = 0.29 (left: 1 0.29 10)
7 / 10 = 0.7 (left: 1 2 0.7)
Input: 4 4 6 8
Possible next steps:
4 + 4 = 8 (left: 6 8 8)
4 + 6 = 10 (left: 8 10 8)
6 + 8 = 14 (left: 4 14 8)
4 * 4 = 16 (left: 6 8 16)
4 * 6 = 24 (left: 8 24 8)
6 * 8 = 48 (left: 4 48 8)
4 - 4 = 0 (left: 0 6 8)
4 - 6 = -2 (left: -2 8 8)
6 - 8 = -2 (left: 4 -2 8)
4 / 4 = 1 (left: 1 6 8)
4 / 6 = 0.67 (left: 8 0.67 8)
6 / 8 = 0.75 (left: 4 0.75 8)
Input: 3 4 5 6
Possible next steps:
3 + 4 = 7 (left: 5 6 7)
4 + 5 = 9 (left: 6 9 7)
4 + 6 = 10 (left: 5 10 7)
5 + 6 = 11 (left: 4 11 7)
3 * 4 = 12 (left: 5 6 12)
4 * 5 = 20 (left: 6 20 7)
4 * 6 = 24 (left: 5 24 7)
5 * 6 = 30 (left: 4 30 7)
3 - 4 = -1 (left: -1 5 6)
4 - 5 = -1 (left: 6 -1 7)
4 - 6 = -2 (left: 5 -2 7)
5 - 6 = -1 (left: 4 -1 7)
3 / 4 = 0.75 (left: 0.75 5 6)
4 / 5 = 0.8 (left: 6 0.8 7)
4 / 6 = 0.67 (left: 5 0.67 7)
5 / 6 = 0.83 (left: 4 0.83 7)
Input: 2 4 6
Possible next steps:
2 + 4 = 6 (left: 6 6)
4 + 6 = 10 (left: 6 10)
2 * 4 = 8 (left: 6 8)
4 * 6 = 24 (left: 6 24)
2 - 4 = -2 (left: -2 6)
4 - 6 = -2 (left: 8 -2)
2 / 4 = 0.5 (left: 0.5 6)
4 / 6 = 0.67 (left: 8 0.67)
Input: {input}
Possible next steps:
'''
\end{lstlisting}
\subsubsection{ToT Decomp prompts}
\label{lst:tot_decomp}
\begin{lstlisting}[caption=ToT Decomp prompts]
select_prompt_1s = '''Select two numbers using the bracket. For example, (2) 8 8 (14) means select 2 and 14. Follow the format of the example.
Do not include serial numbers that are not in the example. Do not include unnecessary information in your output.
Input: 1 2 7 10
Select all combinations of two numbers using bracket. 
Output: 
(1) (2) 7 10
1 (2) (7) 10
1 2 (7) (10)
(1) 2 (7) 10
(1) 2 7 (10)
1 (2) 7 (10)
Input: {input}
Select all combinations of two numbers using bracket. 
Output:
'''


select_prompt_3s = '''Select two numbers using the bracket. For example, (2) 8 8 (14) means select 2 and 14. Follow the format of the example.
Do not include serial numbers that are not in the example. Do not include unnecessary information in your output.
Input: 7 8 9
Select all combinations of two numbers using bracket. 
Output: 
(7) (8) 9
7 (8) (9)
(7) 8 (9)
Input: 2.33 6
Output: 
(2.33) (6)
Input: 1 2 7 10
Select all combinations of two numbers using bracket. 
Output: 
(1) (2) 7 10
1 (2) (7) 10
1 2 (7) (10)
(1) 2 (7) 10
(1) 2 7 (10)
1 (2) 7 (10)
Input: {input}
Select all combinations of two numbers using bracket. 
Output:
'''


select_prompt_5s = '''Select two numbers using the bracket. For example, (2) 8 8 (14) means select 2 and 14. Follow the format of the example.
Do not include serial numbers that are not in the example. Do not include unnecessary information in your output.
Input: 7 8 9
Select all combinations of two numbers using bracket. 
Output: 
(7) (8) 9
7 (8) (9)
(7) 8 (9)
Input: 2.33 6
Output: 
(2.33) (6)
Input: 1 2 7 10
Select all combinations of two numbers using bracket. 
Output: 
(1) (2) 7 10
1 (2) (7) 10
1 2 (7) (10)
(1) 2 (7) 10
(1) 2 7 (10)
1 (2) 7 (10)
Input: 0.66 8 9
Select all combinations of two numbers using bracket. 
Output: 
(0.66) (8) 9
0.66 (8) (9)
(0.66) 8 (9)
Input: 2 8 8 14
Select all combinations of two numbers using bracket. 
Output: 
(2) (8) 8 14
2 (8) (8) 14
2 8 (8) (14)
(2) 8 (8) 14
(2) 8 8 (14)
2 (8) 8 (14)
Input: {input}
Select all combinations of two numbers using bracket. 
Output:
'''

propose_prompt_1s = '''Use the two numbers in the bracket and basic arithmetic operations to propose possible next steps. 
Then, remove the selected numbers by the new number. Use the format '(left: ...)' to present the remaining numbers.
Do not include serial numbers that are not in the example. Do not include unnecessary information in your output.
Input: (2) 8 8 (14)
2 + 14 = 16, replace 2 14 by 16 (left: 8 8 16)
2 * 14 = 28, replace 2 14 by 28 (left: 8 8 28)
2 / 14 = 0.14, replace 2 14 by 0.14 (left: 8 8 0.14)
14 / 2 = 7, replace 2 14 by 7 (left: 8 8 7)
14 - 2 = 12, replace 2 14 by 12 (left: 8 8 12)
2 - 14 = -12, replace 2 14 by -12 (left: 8 8 -12)
Input: {input}
'''

propose_prompt_3s = '''Use the two numbers in the bracket and basic arithmetic operations to propose possible next steps. 
Then, remove the selected numbers by the new number. Use the format '(left: ...)' to present the remaining numbers.
Do not include serial numbers that are not in the example. Do not include unnecessary information in your output.
Input: (2) 8 8 (14)
2 + 14 = 16, replace 2 14 by 16 (left: 8 8 16)
2 * 14 = 28, replace 2 14 by 28 (left: 8 8 28)
2 / 14 = 0.14, replace 2 14 by 0.14 (left: 8 8 0.14)
14 / 2 = 7, replace 2 14 by 7 (left: 8 8 7)
14 - 2 = 12, replace 2 14 by 12 (left: 8 8 12)
2 - 14 = -12, replace 2 14 by -12 (left: 8 8 -12)
Input: 1 (2) 7 (10)
2 + 7 = 9, replace 2 7 by 9 (left: 1 9 10)
2 * 7 = 14, replace 2 7 by 14 (left: 1 14 10)
2 / 7 = 0.29, replace 2 7 by 0.29 (left: 1 0.29 10)
7 / 2 = 3.5, replace 2 7 by 3.5 (left: 1 3.5 10)
7 - 2 = 5, replace 2 7 by 5 (left: 1 5 10)
2 - 7 = -5, replace 2 7 by -5 (left: 1 -5 10)
Input: (7) (8) 9
7 + 8 = 15, replace 7 8 by 15 (left: 15 9)
7 * 8 = 56, replace 7 8 by 56 (left: 56 9)
7 / 8 = 0.88, replace 7 8 by 0.88 (left: 0.88 9)
8 / 7 = 1.14, replace 7 8 by 1.14 (left: 1.14 9)
8 - 7 = 1, replace 7 8 by 1 (left: 1 9)
7 - 8 = -1, replace 7 8 by -1 (left: -1 9)
Input: {input}
'''

propose_prompt_5s = '''Use the two numbers in the bracket and basic arithmetic operations to propose possible next steps. 
Then, remove the selected numbers by the new number. Use the format '(left: ...)' to present the remaining numbers.
Do not include serial numbers that are not in the example. Do not include unnecessary information in your output.
Input: (2) 8 8 (14)
2 + 14 = 16, replace 2 14 by 16 (left: 8 8 16)
2 * 14 = 28, replace 2 14 by 28 (left: 8 8 28)
2 / 14 = 0.14, replace 2 14 by 0.14 (left: 8 8 0.14)
14 / 2 = 7, replace 2 14 by 7 (left: 8 8 7)
14 - 2 = 12, replace 2 14 by 12 (left: 8 8 12)
2 - 14 = -12, replace 2 14 by -12 (left: 8 8 -12)
Input: 1 (2) 7 (10)
2 + 7 = 9, replace 2 7 by 9 (left: 1 9 10)
2 * 7 = 14, replace 2 7 by 14 (left: 1 14 10)
2 / 7 = 0.29, replace 2 7 by 0.29 (left: 1 0.29 10)
7 / 2 = 3.5, replace 2 7 by 3.5 (left: 1 3.5 10)
7 - 2 = 5, replace 2 7 by 5 (left: 1 5 10)
2 - 7 = -5, replace 2 7 by -5 (left: 1 -5 10)
Input: (7) (8) 9
7 + 8 = 15, replace 7 8 by 15 (left: 15 9)
7 * 8 = 56, replace 7 8 by 56 (left: 56 9)
7 / 8 = 0.88, replace 7 8 by 0.88 (left: 0.88 9)
8 / 7 = 1.14, replace 7 8 by 1.14 (left: 1.14 9)
8 - 7 = 1, replace 7 8 by 1 (left: 1 9)
7 - 8 = -1, replace 7 8 by -1 (left: -1 9)
Input: (2.33) (6)
2.33 + 6 = 8.33, replace 2.33 6 by 8.33 (left: 8.33)
2.33 * 6 = 14, replace 2.33 6 by 14 (left: 14)
2.33 / 6 = 0.39, replace 2.33 6 by 0.39 (left: 0.39)
6 / 2.33 = 2.57, replace 2.33 6 by 2.57 (left: 2.57)
6 - 2.33 = 3.67, replace 2.33 6 by 3.67 (left: 3.67)
2.33 - 6 = -3.67, replace 2.33 6 by -3.67 (left: -3.67)
Input: 0.66 (8) (9) 
8 + 9 = 17, replace 8 9 by 17 (left: 0.66 17)
8 * 9 = 72, replace 8 9 by 72 (left: 0.66 72)
8 / 9 = 0.89, replace 8 9 by 0.89 (left: 0.66 0.89)
9 / 8 = 1.12, replace 8 9 by 1.12 (left: 0.66 1.12)
9 - 8 = 1, replace 8 9 by 1 (left: 0.66 1)
8 - 9 = -1, replace 8 9 by -1 (left: 0.66 -1)
Input: {input}
'''

assembly_prompt_1s = '''Use the previous steps of equations to form a final equation that obtains 24. Use 'Answer: ' to present your final answer.
Input: 4 4 6 8
Steps:
4 + 8 = 12 (left: 4 6 12)
6 - 4 = 2 (left: 2 12)
2 * 12 = 24 (left: 24)
Let's do it step by step:
f1 = 4 + 8 = 12. In this step, 4 and 8 are from the input.
f2 = 6 - 4 = 2. In this step, 6 and 4 are from the input.
f3 = 2 * 12 = 24. In this step, 2 is from f2, and 12 is from f1.
Thus, we replace 2 by f2: f3 = 2 * 12 = f2 * 12 = 24
Thus, we replace 12 by f1: f3 = 2 * 12 = f2 * f1 = 24
Since f1 = 4 + 8, we replace f1 by 4 + 8: f3 = 2 * 12 = f2 * (4 + 8) = 24
Since f2 = 6 - 4, we replace f2 by 6 - 4: f3 = 2 * 12 = (6 - 4) * (4 + 8) = 24
Answer: (6 - 4) * (4 + 8) = 24
Input: {input}Let's do it step by step:
f1 = '''

assembly_prompt_3s = '''Use the previous steps of equations to form a final equation that obtains 24. Use 'Answer: ' to present your final answer.
Input: 4 4 6 8
Steps:
4 + 8 = 12 (left: 4 6 12)
6 - 4 = 2 (left: 2 12)
2 * 12 = 24 (left: 24)
Let's do it step by step:
f1 = 4 + 8 = 12. In this step, 4 and 8 are from the input.
f2 = 6 - 4 = 2. In this step, 6 and 4 are from the input.
f3 = 2 * 12 = 24. In this step, 2 is from f2, and 12 is from f1.
Thus, we replace 2 by f2: f3 = 2 * 12 = f2 * 12 = 24
Thus, we replace 12 by f1: f3 = 2 * 12 = f2 * f1 = 24
Since f1 = 4 + 8, we replace f1 by 4 + 8: f3 = 2 * 12 = f2 * (4 + 8) = 24
Since f2 = 6 - 4, we replace f2 by 6 - 4: f3 = 2 * 12 = (6 - 4) * (4 + 8) = 24
Answer: (6 - 4) * (4 + 8) = 24
Input: 2 9 10 12
Steps:
12 * 2 = 24 (left: 9 10 24)
10 - 9 = 1 (left: 1 24)
24 * 1 = 24 (left: 24)
Let's do it step by step:
f1 = 12 * 2 = 24. In this step, 12 and 2 are from the input.
f2 = 10 - 9 = 1. In this step, 10 and 9 are from the input.
f3 = 24 * 1 = 24. In this step, 24 is from f1, and 1 is from f2.
Thus, we replace 24 by f1: f3 = 24 * 1 = f1 * 1 = 24
Thus, we replace 1 by f2: f3 = 24 * 1 = f1 * f2 = 24
Since f1 = 12 * 2, we replace f1 by 12 * 2: f3 = 24 * 1 = (12 * 2) * f2 = 24
Since f2 = 10 - 9, we replace f2 by 10 - 9: f3 = 24 * 1 = (12 * 2) * (10 - 9) = 24
Answer: (12 * 2) * (10 - 9) = 24
Input: 4 9 10 13
Steps:
13 - 10 = 3 (left: 3 4 9)
9 - 3 = 6 (left: 4 6)
4 * 6 = 24 (left: 24)
Let's do it step by step:
f1 = 13 - 10 = 3. In this step, 13 and 10 are from the input.
f2 = 9 - 3 = 6. In this step, 9 is from the input, and 3 is from f1.
Thus, we replace 3 by f1: f2 = 9 - 3 = 9 - f1 = 6
f3 = 4 * 6 = 24. In this step, 4 is from the input, and 6 is from f2.
Thus, we replace 6 by f2: f3 = 4 * 6 = 4 * f2 = 24
Since f2 = 9 - f1, we replace f2 by 9 - f1: f3 = 4 * 6 = 4 * (9 - f1) = 24
Since f1 = 13 - 10, we replace f1 by 13 - 10: f3 = 4 * 6 = 4 * (9 - (13 - 10)) = 24
Answer: 4 * (9 - (13 - 10)) = 24
Input: {input}Let's do it step by step:
f1 = '''

assembly_prompt_5s = '''Use the previous steps of equations to form a final equation that obtains 24. Use 'Answer: ' to present your final answer.
Input: 4 4 6 8
Steps:
4 + 8 = 12 (left: 4 6 12)
6 - 4 = 2 (left: 2 12)
2 * 12 = 24 (left: 24)
Let's do it step by step:
f1 = 4 + 8 = 12. In this step, 4 and 8 are from the input.
f2 = 6 - 4 = 2. In this step, 6 and 4 are from the input.
f3 = 2 * 12 = 24. In this step, 2 is from f2, and 12 is from f1.
Thus, we replace 2 by f2: f3 = 2 * 12 = f2 * 12 = 24
Thus, we replace 12 by f1: f3 = 2 * 12 = f2 * f1 = 24
Since f1 = 4 + 8, we replace f1 by 4 + 8: f3 = 2 * 12 = f2 * (4 + 8) = 24
Since f2 = 6 - 4, we replace f2 by 6 - 4: f3 = 2 * 12 = (6 - 4) * (4 + 8) = 24
Answer: (6 - 4) * (4 + 8) = 24
Input: 2 9 10 12
Steps:
12 * 2 = 24 (left: 9 10 24)
10 - 9 = 1 (left: 1 24)
24 * 1 = 24 (left: 24)
Let's do it step by step:
f1 = 12 * 2 = 24. In this step, 12 and 2 are from the input.
f2 = 10 - 9 = 1. In this step, 10 and 9 are from the input.
f3 = 24 * 1 = 24. In this step, 24 is from f1, and 1 is from f2.
Thus, we replace 24 by f1: f3 = 24 * 1 = f1 * 1 = 24
Thus, we replace 1 by f2: f3 = 24 * 1 = f1 * f2 = 24
Since f1 = 12 * 2, we replace f1 by 12 * 2: f3 = 24 * 1 = (12 * 2) * f2 = 24
Since f2 = 10 - 9, we replace f2 by 10 - 9: f3 = 24 * 1 = (12 * 2) * (10 - 9) = 24
Answer: (12 * 2) * (10 - 9) = 24
Input: 4 9 10 13
Steps:
13 - 10 = 3 (left: 3 4 9)
9 - 3 = 6 (left: 4 6)
4 * 6 = 24 (left: 24)
Let's do it step by step:
f1 = 13 - 10 = 3. In this step, 13 and 10 are from the input.
f2 = 9 - 3 = 6. In this step, 9 is from the input, and 3 is from f1.
Thus, we replace 3 by f1: f2 = 9 - 3 = 9 - f1 = 6
f3 = 4 * 6 = 24. In this step, 4 is from the input, and 6 is from f2.
Thus, we replace 6 by f2: f3 = 4 * 6 = 4 * f2 = 24
Since f2 = 9 - f1, we replace f2 by 9 - f1: f3 = 4 * 6 = 4 * (9 - f1) = 24
Since f1 = 13 - 10, we replace f1 by 13 - 10: f3 = 4 * 6 = 4 * (9 - (13 - 10)) = 24
Answer: 4 * (9 - (13 - 10)) = 24
Input: 1 4 8 8 
Steps:
8 / 4 = 2 (left: 1 2 8)
1 + 2 = 3 (left: 3 8)
3 * 8 = 24 (left: 24)
Let's do it step by step: 
f1 = 8 / 4 = 2. In this step, 8 and 4 are from the input.
f2 = 1 + 2 = 3. In this step, 2 is from f1, and 1 is from the input.
Thus, we replace 2 by f1: f2 = 1 + 2 = 1 + f1 = (1 + (8 / 4)) = 3
f3 = 3 * 8 = 24. In this step, 3 is from f2, and 8 is from the input.
Thus, we replace 3 by f2: f3 = 3 * 8 = f2 * 8 = 24
Since f2 = 1 + f1, we replace f2 by 1 + f1: f3 = 3 * 8 = (1 + f1) * 8 = 24
Since f1 = 8 / 4, we replace f1 by 8 / 4: (1 + f1) * 8 = (1 + (8 / 4)) * 8 = 24
Answer: (1 + (8 / 4)) * 8 = 24
Input: 5 5 5 9
Steps:
5 + 5 = 10 (left: 5 9 10)
10 + 5 = 15 (left: 9 15)
15 + 9 = 24 (left: 24)
Let's do it step by step:
f1 = 5 + 5 = 10. In this step, 5 and 5 are from the input.
f2 = 10 + 5 = 15. In this step, 10 is from f1, and 5 is from the input.
Thus, we replace 10 by f1: f2 = 10 + 5 = f1 + 5 = 15
f3 = 15 + 9 = 24. In this step, 15 is from f2, and 9 is from the input.
Thus, we replace 15 by f2: f3 = 15 + 9 = f2 + 9 = 24
Since f2 = f1 + 5, we replace f2 by f1 + 5: f3 = 15 + 9 = (f1 + 5) + 9 = 24
Since f1 = 5 + 5, we replace f1 by 5 + 5: f3 = 15 + 9 = ((5 + 5) + 5) + 9 = 24
Answer: ((5 + 5) + 5) + 9 = 24
Input: {input}Let's do it step by step:
f1 = '''
\end{lstlisting}

\begin{lstlisting}[caption=MusiQue prompts]
direct = """Please answer a question given some paragraphs as context. Respond only the answer, in the format "The answer is ###."

Below are some examples (contexts are omitted):

Question: What is the extreme low temperature of the city where WNJN-FM is located?
The answer is -9 (*@°@*)F.

Question: When did muslim armies invade the country Al-Mahabah is located and the country Kleicha originates?
The answer is in 634.

Question: When did hurricane Sandy his the city where The Dealer's performer was born?
The answer is October 28, 2012.

Question: What is the enrollment of undergraduates at the university attended by the entrepreneur owning the gold spike in the location holding PollyGrind Film Festival?
The answer is 7,200.

Question: When did the nation that seized the country where Al-Berka is located from the empire that declined following the Crimean War join the Allies in WW2?
The answer is September 1943.

Question: An institution like a German Fachhochschule is known by what term in (*@Éric@*) Losfeld's birth country and the country where painters remained focused on textures and surfaces.
The answer is hogeschool.

Now, answer a question given the following paragraphs as context:

{ctx}

Question: {question}
The answer is"""

cot = """Please answer a question given some paragraphs as context. Respond by decomposing the question into subquestions, and end with the format "The answer is ###."

Below are some examples (contexts are omitted):
Question: What is the extreme low temperature of the city where WNJN-FM is located?
The question can be decomposed into following subquestions:
What city is WNJN-FM located? Answer: Atlantic City
What is the extreme low temperature of Atlantic City ? Answer: -9 (*@°F@*)
The answer is -9 (*@°F@*).

Question: When did muslim armies invade the country Al-Mahabah is located and the country Kleicha originates?
The question can be decomposed into following subquestions:
Which was the country for Kleicha? Answer: Iraq
Which country is Al-Mahabah in? Answer: Syria
When did muslim armies invade Syria and Iraq? Answer: in 634
The answer is in 634.

Question: When did hurricane Sandy his the city where The Dealer's performer was born?
The question can be decomposed into following subquestions:
Who is The Dealers' performer? Answer: Mal Waldron
Where is Mal Waldron's place of birth? Answer: New York City
When did hurricane sandy hit New York City? Answer: October 28, 2012
The answer is October 28, 2012.

Question: What is the enrollment of undergraduates at the university attended by the entrepreneur owning the gold spike in the location holding PollyGrind Film Festival?
The question can be decomposed into following subquestions:
What is the location of PollyGrind Film Festival? Answer: Las Vegas
Who owns the gold spike in Las Vegas? Answer: Tony Hsieh
Tony Hsieh is educated at where? Answer: Harvard
What is the enrollment of undergraduates at Harvard ? Answer: 7,200
The answer is 7,200.

Question: When did the nation that seized the country where Al-Berka is located from the empire that declined following the Crimean War join the Allies in WW2?
The question can be decomposed into following subquestions:
Which country is Al-Berka in? Answer: Libya
What empire declined after the Crimean War? Answer: the Ottoman Empire
Which european state seized Libya from the Ottoman Empire in 1911? Answer: Italy
When did Italy join the allies in ww2? Answer: September 1943
The answer is September 1943.

Question: An institution like a German Fachhochschule is known by what term in (*@Éric@*) Losfeld's birth country and the country where painters remained focused on textures and surfaces.
The question can be decomposed into following subquestions:
Where is (*@Éric@*) Losfeld's place of birth? Answer: Mouscron
Arrondissement of Mouscron >> country? Answer: Belgium
Where was the focus of paintings on textures and surfaces? Answer: the Netherlands
What term is used in Belgium and the the Netherlands to refer to an institution like a German Fachhochschule? Answer: hogeschool
The answer is hogeschool.

Now, answer a question given the following paragraphs as context:

{ctx}

Question: {question}
The question can be decomposed into following subquestions:"""

direct_triplet = """Please answer a question given some relation triplets as context. Respond only the answer, in the format "The answer is ###."

Below are some examples:

Relation triplets:
(WNJN-FM, located in, Atlantic City)
(The extreme low temperature of Atlantic City, is, -9 (*@°F@*))
Question: What is the extreme low temperature of the city where WNJN-FM is located?
The answer is -9 (*@°F@*).

Relation triplets:
(Kleicha, country, Iraq)
(Al-Mahabah, country, Syria)
(Muslim armies, invade, Syria and Iraq)
(Syria and Iraq, invaded in, 634)
Question: When did muslim armies invade the country Al-Mahabah is located and the country Kleicha originates?
The answer is in 634.

Relation triplets:
(The Dealers, performer, Mal Waldron)
(Mal Waldron, place of birth, New York City)
(Hurricane Sandy, hit New York City, October 28, 2012)
Question: When did hurricane Sandy his the city where The Dealer's performer was born?
The answer is October 28, 2012.

Relation triplets:
(PollyGrind Film Festival, location, Las Vegas)
(Gold Spike, owned by, Tony Hsieh)
(Tony Hsieh, educated at, Harvard)
(Harvard, undergraduate enrollment, 7,200)
Question: What is the enrollment of undergraduates at the university attended by the entrepreneur owning the gold spike in the location holding PollyGrind Film Festival?
The answer is 7,200.

Relation triplets:
(Al-Berka, country, Libya))
(Ottoman Empire, declined after, Crimean War)
(The European state, seized Libya from the Ottoman Empire in 1911, Italy)
(Italy, join the allies in WW2, September 1943)
Question: When did the nation that seized the country where Al-Berka is located from the empire that declined following the Crimean War join the Allies in WW2?
The answer is September 1943.

Relation triplets:
((*@Éric@*) Losfeld, place of birth, Mouscron)
(The Collegian, owned by, Houston Baptist University)
(Houston Baptist University, founded, 1960)
(Hertfordshire, located in, East of England)
(Jan (*@Šindel@*), birthplace, Hradec (*@Králové@*))
(Arrondissement of Mouscron, country, Belgium)
(The focus of paintings on textures and surfaces, located in, the Netherlands)
(Belgium and the Netherlands, refer to, hogeschool)
Question: An institution like a German Fachhochschule is known by what term in (*@Éric@*) Losfeld's birth country and the country where painters remained focused on textures and surfaces.
The answer is hogeschool.

Now, answer a question given the following relation triplets as context:

{ctx}

Question: {question}
The answer is"""

cot_triplet = """Please answer a question given some relation triplets as context. Respond by decomposing the question into subquestions, and end with the format "The answer is ###."

Below are some examples:
Relation treiplets:
(WNJN-FM, located in, Atlantic City)
(The extreme low temperature of Atlantic City, is, -9 (*@°F@*))
Question: What is the extreme low temperature of the city where WNJN-FM is located?
The question can be decomposed into following subquestions:
What city is WNJN-FM located? Answer: Atlantic City
What is the extreme low temperature of Atlantic City ? Answer: -9 (*@°F@*)
The answer is -9 (*@°F@*).


Relation treiplets:
(Kleicha, country, Iraq)
(Al-Mahabah, country, Syria)
(Muslim armies, invade, Syria and Iraq)
(Syria and Iraq, invaded in, 634)
Question: When did muslim armies invade the country Al-Mahabah is located and the country Kleicha originates?
The question can be decomposed into following subquestions:
Which was the country for Kleicha? Answer: Iraq
Which country is Al-Mahabah in? Answer: Syria
When did muslim armies invade Syria and Iraq? Answer: in 634
The answer is in 634.


Relation treiplets:
(The Dealers, performer, Mal Waldron)
(Mal Waldron, place of birth, New York City)
(Hurricane Sandy, hit New York City, October 28, 2012)
Question: When did hurricane Sandy his the city where The Dealer's performer was born?
The question can be decomposed into following subquestions:
Who is The Dealers' performer? Answer: Mal Waldron
Where is Mal Waldron's place of birth? Answer: New York City
When did hurricane sandy hit New York City? Answer: October 28, 2012
The answer is October 28, 2012.


Relation treiplets:
(PollyGrind Film Festival, location, Las Vegas)
(Gold Spike, owned by, Tony Hsieh)
(Tony Hsieh, educated at, Harvard)
(Harvard, undergraduate enrollment, 7,200)
Question: What is the enrollment of undergraduates at the university attended by the entrepreneur owning the gold spike in the location holding PollyGrind Film Festival?
The question can be decomposed into following subquestions:
What is the location of PollyGrind Film Festival? Answer: Las Vegas
Who owns the gold spike in Las Vegas? Answer: Tony Hsieh
Tony Hsieh is educated at where? Answer: Harvard
What is the enrollment of undergraduates at Harvard ? Answer: 7,200
The answer is 7,200.


Relation treiplets:
(Al-Berka, country, Libya))
(Ottoman Empire, declined after, Crimean War)
(The European state, seized Libya from the Ottoman Empire in 1911, Italy)
(Italy, join the allies in WW2, September 1943)
Question: When did the nation that seized the country where Al-Berka is located from the empire that declined following the Crimean War join the Allies in WW2?
The question can be decomposed into following subquestions:
Which country is Al-Berka in? Answer: Libya
What empire declined after the Crimean War? Answer: the Ottoman Empire
Which european state seized Libya from the Ottoman Empire in 1911? Answer: Italy
When did Italy join the allies in ww2? Answer: September 1943
The answer is September 1943.


Relation treiplets:
((*@Éric@*) Losfeld, place of birth, Mouscron)
(The Collegian, owned by, Houston Baptist University)
(Houston Baptist University, founded, 1960)
(Hertfordshire, located in, East of England)
(Jan (*@Šindel@*), birthplace, Hradec (*@Králové@*))
(Arrondissement of Mouscron, country, Belgium)
(The focus of paintings on textures and surfaces, located in, the Netherlands)
(Belgium and the Netherlands, refer to, hogeschool)
Question: An institution like a German Fachhochschule is known by what term in (*@Éric@*) Losfeld's birth country and the country where painters remained focused on textures and surfaces.
The question can be decomposed into following subquestions:
Where is (*@Éric@*) Losfeld's place of birth? Answer: Mouscron
Arrondissement of Mouscron >> country? Answer: Belgium
Where was the focus of paintings on textures and surfaces? Answer: the Netherlands
What term is used in Belgium and the the Netherlands to refer to an institution like a German Fachhochschule? Answer: hogeschool
The answer is hogeschool.


Now, answer a question given the following relation triplets as context:

{ctx}

Question: {question}
The question can be decomposed into following subquestions:"""

tot = """Please answer a question given some paragraphs as context. Respond by decomposing the question into subquestions, and end with the format "The answer to the original question is ###." All questions can find their answers in the provided paragraphs.

Below are some examples to demonstrate the desired answering format (contexts are omitted for brevity):
Question: What is the extreme low temperature of the city where WNJN-FM is located?
The question can be decomposed into following subquestions:
sub question: What city is WNJN-FM located?
sub answer: Atlantic City
sub question: What is the extreme low temperature of Atlantic City ?
sub answer: -9 (*@°F@*)
The answer to the original question is -9 (*@°F@*).

Question: When did muslim armies invade the country Al-Mahabah is located and the country Kleicha originates?
The question can be decomposed into following subquestions:
sub question: Which was the country for Kleicha?
sub answer: Iraq
sub question: Which country is Al-Mahabah in?
sub answer: Syria
sub question: When did muslim armies invade Syria and Iraq?
sub answer: in 634
The answer to the original question is in 634.

Question: When did hurricane Sandy his the city where The Dealer's performer was born?
The question can be decomposed into following subquestions:
sub question: Who is The Dealers' performer?
sub answer: Mal Waldron
sub question: Where is Mal Waldron's place of birth?
sub answer: New York City
sub question: When did hurricane sandy hit New York City?
sub answer: October 28, 2012
The answer to the original question is October 28, 2012.

Question: What is the enrollment of undergraduates at the university attended by the entrepreneur owning the gold spike in the location holding PollyGrind Film Festival?
The question can be decomposed into following subquestions:
sub question: What is the location of PollyGrind Film Festival?
sub answer: Las Vegas
sub question: Who owns the gold spike in Las Vegas?
sub answer: Tony Hsieh
sub question: Tony Hsieh is educated at where?
sub answer: Harvard
sub question: What is the enrollment of undergraduates at Harvard ?
sub answer: 7,200
The answer to the original question is 7,200.

Question: When did the nation that seized the country where Al-Berka is located from the empire that declined following the Crimean War join the Allies in WW2?
The question can be decomposed into following subquestions:
sub question: Which country is Al-Berka in?
sub answer: Libya
sub question: What empire declined after the Crimean War?
sub answer: the Ottoman Empire
sub question: Which european state seized Libya from the Ottoman Empire in 1911?
sub answer: Italy
sub question: When did Italy join the allies in ww2?
sub answer: September 1943
The answer to the original question is September 1943.

Question: An institution like a German Fachhochschule is known by what term in (*@Éric@*) Losfeld's birth country and the country where painters remained focused on textures and surfaces.
The question can be decomposed into following subquestions:
sub question: Where is (*@Éric@*) Losfeld's place of birth?
sub answer: Mouscron
sub question: Arrondissement of Mouscron >> country?
sub answer: Belgium
sub question: Where was the focus of paintings on textures and surfaces?
sub answer: the Netherlands
sub question: What term is used in Belgium and the the Netherlands to refer to an institution like a German Fachhochschule?
sub answer: hogeschool
The answer to the original question is hogeschool.

Now, answer a question given the following paragraphs as context. Remember to decompose the question into subquestions, and it is guaranteed that the answer can be found in the provided paragraphs:

{ctx}

Now answer the following question. Remember to decompose the question into subquestions, and it is guaranteed that the answer can be found in the provided paragraphs.

Question: {question}
The question can be decomposed into following subquestions:"""

tot_triplet = """Please answer a question given some paragraphs as context. Respond by decomposing the question into subquestions, and end with the format "The answer to the original question is ###." All questions can find their answers in the provided paragraphs.

Below are some examples to demonstrate the desired answering format (contexts are omitted for brevity):
Question: What is the extreme low temperature of the city where WNJN-FM is located?
The question can be decomposed into following subquestions:
sub question: What city is WNJN-FM located?
sub answer: Atlantic City
sub question: What is the extreme low temperature of Atlantic City ?
sub answer: -9 (*@°F@*)
The answer to the original question is -9 (*@°F@*).

Question: When did muslim armies invade the country Al-Mahabah is located and the country Kleicha originates?
The question can be decomposed into following subquestions:
sub question: Which was the country for Kleicha?
sub answer: Iraq
sub question: Which country is Al-Mahabah in?
sub answer: Syria
sub question: When did muslim armies invade Syria and Iraq?
sub answer: in 634
The answer to the original question is in 634.

Question: When did hurricane Sandy his the city where The Dealer's performer was born?
The question can be decomposed into following subquestions:
sub question: Who is The Dealers' performer?
sub answer: Mal Waldron
sub question: Where is Mal Waldron's place of birth?
sub answer: New York City
sub question: When did hurricane sandy hit New York City?
sub answer: October 28, 2012
The answer to the original question is October 28, 2012.

Question: What is the enrollment of undergraduates at the university attended by the entrepreneur owning the gold spike in the location holding PollyGrind Film Festival?
The question can be decomposed into following subquestions:
sub question: What is the location of PollyGrind Film Festival?
sub answer: Las Vegas
sub question: Who owns the gold spike in Las Vegas?
sub answer: Tony Hsieh
sub question: Tony Hsieh is educated at where?
sub answer: Harvard
sub question: What is the enrollment of undergraduates at Harvard ?
sub answer: 7,200
The answer to the original question is 7,200.

Question: When did the nation that seized the country where Al-Berka is located from the empire that declined following the Crimean War join the Allies in WW2?
The question can be decomposed into following subquestions:
sub question: Which country is Al-Berka in?
sub answer: Libya
sub question: What empire declined after the Crimean War?
sub answer: the Ottoman Empire
sub question: Which european state seized Libya from the Ottoman Empire in 1911?
sub answer: Italy
sub question: When did Italy join the allies in ww2?
sub answer: September 1943
The answer to the original question is September 1943.

Question: An institution like a German Fachhochschule is known by what term in (*@Éric@*) Losfeld's birth country and the country where painters remained focused on textures and surfaces.
The question can be decomposed into following subquestions:
sub question: Where is (*@Éric@*) Losfeld's place of birth?
sub answer: Mouscron
sub question: Arrondissement of Mouscron >> country?
sub answer: Belgium
sub question: Where was the focus of paintings on textures and surfaces?
sub answer: the Netherlands
sub question: What term is used in Belgium and the the Netherlands to refer to an institution like a German Fachhochschule?
sub answer: hogeschool
The answer to the original question is hogeschool.

Now, answer a question given the following paragraphs as context. Remember to decompose the question into subquestions, and it is guaranteed that the answer can be found in the provided paragraphs:

{triplet_ctx}

Now answer the following question. Remember to decompose the question into subquestions, and it is guaranteed that the answer can be found in the provided paragraphs.

Question: {question}
The question can be decomposed into following subquestions:"""

tot_self_eval = """
The following are candidates to answer a multi-hop question. Some of them are complete reasoning trajectories while other may be intermediate. Please sort these candidates based on how likely they will lead to a correct solution. You should sort based on the quality, instead of length, i.e. complete reasoning may or may not be correct. You should return a comma separated list, and use #1 to indicate the first candidate, #2 to indicate the second candidate, and so on. Note that the multi-hop question can be answered by decomposing them into subquestions and answering them one by one. A good candidate should be on the right track of decomposing question, and make no mistakes, and a good final answer should be as concise as possible.

Below is an example:

Candidate 1:
Question: Where was the film The Beach filmed in the country where Pao Sarasin was born?
The question can be decomposed into following subquestions:
sub question: Where was Pao Sarasin born?
sub answer: Bangkok, Thailand
sub question: Where was the film The Beach filmed?
sub answer: The film The Beach was filmed on the Thai island Koh Phi Phi.
The answer to the original question is Thailand.

Candidate 2:
Question: Where was the film The Beach filmed in the country where Pao Sarasin was born?
The question can be decomposed into following subquestions:
sub question: Where was Pao Sarasin born?
sub answer: Bangkok, Thailand
sub question: Where was the film The Beach filmed?
sub answer: Koh Phi Phi, Thailand
The answer to the original question is Koh Phi Phi, Thailand.

Candidate 3:
Question: Where was the film The Beach filmed in the country where Pao Sarasin was born?
The question can be decomposed into following subquestions:
sub question: Where was Pao Sarasin born?
sub answer: Bangkok, Thailand
sub question: Where was the film The Beach filmed?
sub answer: The film The Beach was filmed on the Thai island Koh Phi Phi, in Thailand.
The answer to the original question is The film The Beach was filmed on the Thai island Koh Phi Phi, in Thailand.

Candidate 4:
Question: Where was the film The Beach filmed in the country where Pao Sarasin was born?
The question can be decomposed into following subquestions:
sub question: What is the country of birth of Pao Sarasin?
sub answer: Thailand
sub question: Where was the film The Beach filmed?
sub answer: The Thai island Koh Phi Phi
The answer to the original question is The Thai island Koh Phi Phi.

Candidate 5:
Question: Where was the film The Beach filmed in the country where Pao Sarasin was born?
The question can be decomposed into following subquestions:
sub question: What is the country of birth of Pao Sarasin?
sub answer: Thailand
sub question: Where was the film The Beach filmed?
sub answer: The film The Beach was filmed on the Thai island Koh Phi Phi.
The answer to the original question is The film The Beach was filmed on the Thai island Koh Phi Phi.

Now pick the most likely candidate to answer the original question.
Answer: #4, #5, #3, #2, #1

Now pick from the following candidates:

{candidates}

Now pick the most likely candidate to answer the original question.
Answer: 
"""
\end{lstlisting}

\begin{lstlisting}[caption=Blocksworld prompts]
    {
    "example_pool": [
        {
            "init": "the red block is clear, the orange block is clear, the hand is empty, the orange block is on top of the blue block, the red block is on the table and the blue block is on the table",
            "goal": "the blue block is on top of the orange block",
            "plan": "\nunstack the orange block from on top of the blue block\nput down the orange block\npick up the blue block\nstack the blue block on top of the orange block\n[PLAN END]\n",
            "states": [
                "the red block is clear, the blue block is clear, the orange block is clear, the hand is empty, the red block is on the table, the blue block is on the table and the orange block is on the table",
                "the red block is clear, the blue block is clear, the hand is empty, the blue block is on top of the orange block, the red block is on the table and the orange block is on the table"
            ]
        },
        {
            "init": "the blue block is clear, the orange block is clear, the hand is empty, the red block is on top of the yellow block, the orange block is on top of the red block, the blue block is on the table and the yellow block is on the table",
            "goal": "the blue block is on top of the yellow block and the orange block is on top of the blue block",
            "plan": "\nunstack the orange block from on top of the red block\nput down the orange block\nunstack the red block from on top of the yellow block\nput down the red block\npick up the blue block\nstack the blue block on top of the yellow block\npick up the orange block\nstack the orange block on top of the blue block\n[PLAN END]\n",
            "states": [
                "the red block is clear, the blue block is clear, the orange block is clear, the hand is empty, the red block is on top of the yellow block, the blue block is on the table, the orange block is on the table and the yellow block is on the table",
                "the red block is clear, the blue block is clear, the orange block is clear, the yellow block is clear, the hand is empty, the red block is on the table, the blue block is on the table, the orange block is on the table and the yellow block is on the table",
                "the red block is clear, the blue block is clear, the orange block is clear, the hand is empty, the blue block is on top of the yellow block, the red block is on the table, the orange block is on the table and the yellow block is on the table",
                "the red block is clear, the orange block is clear, the hand is empty, the blue block is on top of the yellow block, the orange block is on top of the blue block, the red block is on the table and the yellow block is on the table"
            ]
        },
        {
            "init": "the red block is clear, the yellow block is clear, the hand is empty, the red block is on top of the blue block, the blue block is on top of the orange block, the orange block is on the table and the yellow block is on the table",
            "goal": "the blue block is on top of the orange block and the yellow block is on top of the red block",
            "plan": "\npick up the yellow block\nstack the yellow block on top of the red block\n[PLAN END]\n",
            "states": [
                "the yellow block is clear, the hand is empty, the red block is on top of the blue block, the blue block is on top of the orange block, the yellow block is on top of the red block and the orange block is on the table"
            ]
        },
        {
            "init": "the blue block is clear, the yellow block is clear, the hand is empty, the red block is on top of the orange block, the blue block is on top of the red block, the orange block is on the table and the yellow block is on the table",
            "goal": "the blue block is on top of the red block and the yellow block is on top of the blue block",
            "plan": "\npick up the yellow block\nstack the yellow block on top of the blue block\n[PLAN END]\n",
            "states": [
                "the yellow block is clear, the hand is empty, the red block is on top of the orange block, the blue block is on top of the red block, the yellow block is on top of the blue block and the orange block is on the table"
            ]
        },
        {
            "init": "the blue block is clear, the orange block is clear, the hand is empty, the blue block is on top of the yellow block, the yellow block is on top of the red block, the red block is on the table and the orange block is on the table",
            "goal": "the blue block is on top of the red block and the orange block is on top of the yellow block",
            "plan": "\nunstack the blue block from on top of the yellow block\nstack the blue block on top of the orange block\nunstack the yellow block from on top of the red block\nput down the yellow block\nunstack the blue block from on top of the orange block\nstack the blue block on top of the red block\npick up the orange block\nstack the orange block on top of the yellow block\n[PLAN END]\n",
            "states": [
                "the blue block is clear, the yellow block is clear, the hand is empty, the blue block is on top of the orange block, the yellow block is on top of the red block, the red block is on the table and the orange block is on the table",
                "the red block is clear, the blue block is clear, the yellow block is clear, the hand is empty, the blue block is on top of the orange block, the red block is on the table, the orange block is on the table and the yellow block is on the table",
                "the blue block is clear, the orange block is clear, the yellow block is clear, the hand is empty, the blue block is on top of the red block, the red block is on the table, the orange block is on the table and the yellow block is on the table",
                "the blue block is clear, the orange block is clear, the hand is empty, the blue block is on top of the red block, the orange block is on top of the yellow block, the red block is on the table and the yellow block is on the table"
            ]
        },
        {
            "init": "the blue block is clear, the orange block is clear, the hand is empty, the blue block is on top of the red block, the red block is on the table and the orange block is on the table",
            "goal": "the red block is on top of the orange block and the orange block is on top of the blue block",
            "plan": "\nunstack the blue block from on top of the red block\nput down the blue block\npick up the orange block\nstack the orange block on top of the blue block\npick up the red block\nstack the red block on top of the orange block\n[PLAN END]\n",
            "states": [
                "the red block is clear, the blue block is clear, the orange block is clear, the hand is empty, the red block is on the table, the blue block is on the table and the orange block is on the table",
                "the red block is clear, the orange block is clear, the hand is empty, the orange block is on top of the blue block, the red block is on the table and the blue block is on the table",
                "the red block is clear, the hand is empty, the red block is on top of the orange block, the orange block is on top of the blue block and the blue block is on the table"
            ]
        },
        {
            "init": "the red block is clear, the yellow block is clear, the hand is empty, the red block is on top of the orange block, the orange block is on top of the blue block, the blue block is on the table and the yellow block is on the table",
            "goal": "the red block is on top of the yellow block, the blue block is on top of the orange block and the yellow block is on top of the blue block",
            "plan": "\nunstack the red block from on top of the orange block\nput down the red block\nunstack the orange block from on top of the blue block\nput down the orange block\npick up the blue block\nstack the blue block on top of the orange block\npick up the yellow block\nstack the yellow block on top of the blue block\npick up the red block\nstack the red block on top of the yellow block\n[PLAN END]\n",
            "states": [
                "the red block is clear, the orange block is clear, the yellow block is clear, the hand is empty, the orange block is on top of the blue block, the red block is on the table, the blue block is on the table and the yellow block is on the table",
                "the red block is clear, the blue block is clear, the orange block is clear, the yellow block is clear, the hand is empty, the red block is on the table, the blue block is on the table, the orange block is on the table and the yellow block is on the table",
                "the red block is clear, the blue block is clear, the yellow block is clear, the hand is empty, the blue block is on top of the orange block, the red block is on the table, the orange block is on the table and the yellow block is on the table",
                "the red block is clear, the yellow block is clear, the hand is empty, the blue block is on top of the orange block, the yellow block is on top of the blue block, the red block is on the table and the orange block is on the table",
                "the red block is clear, the hand is empty, the red block is on top of the yellow block, the blue block is on top of the orange block, the yellow block is on top of the blue block and the orange block is on the table"
            ]
        },
        {
            "init": "the red block is clear, the blue block is clear, the hand is empty, the red block is on top of the orange block, the blue block is on the table and the orange block is on the table",
            "goal": "the red block is on top of the blue block and the blue block is on top of the orange block",
            "plan": "\nunstack the red block from on top of the orange block\nput down the red block\npick up the blue block\nstack the blue block on top of the orange block\npick up the red block\nstack the red block on top of the blue block\n[PLAN END]\n",
            "states": [
                "the red block is clear, the blue block is clear, the orange block is clear, the hand is empty, the red block is on the table, the blue block is on the table and the orange block is on the table",
                "the red block is clear, the blue block is clear, the hand is empty, the blue block is on top of the orange block, the red block is on the table and the orange block is on the table",
                "the red block is clear, the hand is empty, the red block is on top of the blue block, the blue block is on top of the orange block and the orange block is on the table"
            ]
        },
        {
            "init": "the blue block is clear, the yellow block is clear, the hand is empty, the blue block is on top of the orange block, the yellow block is on top of the red block, the red block is on the table and the orange block is on the table",
            "goal": "the blue block is on top of the red block and the yellow block is on top of the orange block",
            "plan": "\nunstack the blue block from on top of the orange block\nput down the blue block\nunstack the yellow block from on top of the red block\nstack the yellow block on top of the orange block\npick up the blue block\nstack the blue block on top of the red block\n[PLAN END]\n",
            "states": [
                "the blue block is clear, the orange block is clear, the yellow block is clear, the hand is empty, the yellow block is on top of the red block, the red block is on the table, the blue block is on the table and the orange block is on the table",
                "the red block is clear, the blue block is clear, the yellow block is clear, the hand is empty, the yellow block is on top of the orange block, the red block is on the table, the blue block is on the table and the orange block is on the table",
                "the blue block is clear, the yellow block is clear, the hand is empty, the blue block is on top of the red block, the yellow block is on top of the orange block, the red block is on the table and the orange block is on the table"
            ]
        },
        {
            "init": "the red block is clear, the orange block is clear, the white block is clear, the hand is empty, the blue block is on top of the yellow block, the white block is on top of the blue block, the red block is on the table, the orange block is on the table and the yellow block is on the table",
            "goal": "the blue block is on top of the yellow block, the orange block is on top of the white block and the white block is on top of the red block",
            "plan": "\nunstack the white block from on top of the blue block\nstack the white block on top of the red block\npick up the orange block\nstack the orange block on top of the white block\n[PLAN END]\n",
            "states": [
                "the blue block is clear, the orange block is clear, the white block is clear, the hand is empty, the blue block is on top of the yellow block, the white block is on top of the red block, the red block is on the table, the orange block is on the table and the yellow block is on the table",
                "the blue block is clear, the orange block is clear, the hand is empty, the blue block is on top of the yellow block, the orange block is on top of the white block, the white block is on top of the red block, the red block is on the table and the yellow block is on the table"
            ]
        }
    ],
    "intro": "I am playing with a set of blocks where I need to arrange the blocks into stacks. Here are the actions I can do\n\nPick up a block\nUnstack a block from on top of another block\nPut down a block\nStack a block on top of another block\n\nI have the following restrictions on my actions:\nI can only pick up or unstack one block at a time.\nI can only pick up or unstack a block if my hand is empty.\nI can only pick up a block if the block is on the table and the block is clear. A block is clear if the block has no other blocks on top of it and if the block is not picked up.\nI can only unstack a block from on top of another block if the block I am unstacking was really on top of the other block.\nI can only unstack a block from on top of another block if the block I am unstacking is clear.\nOnce I pick up or unstack a block, I am holding the block.\nI can only put down a block that I am holding.\nI can only stack a block on top of another block if I am holding the block being stacked.\nI can only stack a block on top of another block if the block onto which I am stacking the block is clear.\nOnce I put down or stack a block, my hand becomes empty.\n\n",
    "world_update_pickup": "I am playing with a set of blocks where I need to arrange the blocks into stacks. Here are the actions I can do \n\nPick up a block \nUnstack a block from on top of another block \nPut down a block \nStack a block on top of another block \n\nI have the following restrictions on my actions:\nI can only pick up or unstack one block at a time. \nI can only pick up or unstack a block if my hand is empty. \nI can only pick up a block if the block is on the table and the block is clear. A block is clear if the block has no other blocks on top of it and if the block is not picked up. \nI can only unstack a block from on top of another block if the block I am unstacking was really on top of the other block. \nI can only unstack a block from on top of another block if the block I am unstacking is clear. Once I pick up or unstack a block, I am holding the block. \nI can only put down a block that I am holding. \nI can only stack a block on top of another block if I am holding the block being stacked. \nI can only stack a block on top of another block if the block onto which I am stacking the block is clear. Once I put down or stack a block, my hand becomes empty.\n\nAfter being given an initial state and an action, give the new state after performing the action.\n\n[SCENARIO 1]\n[STATE 0] I have that, the white block is clear, the cyan block is clear, the brown block is clear, the hand is empty, the white block is on top of the purple block, the purple block is on the table, the cyan block is on the table and the brown block is on the table.\n[ACTION] Pick up the brown block.\n[CHANGE] The hand was empty and is now holding the brown block, the brown block was on the table and is now in the hand, and the brown block is no longer clear.\n[STATE 1] I have that, the white block is clear, the cyan block is clear, the brown block is in the hand, the hand is holding the brown block, the white block is on top of the purple block, the purple block is on the table and the cyan block is on the table.\n\n[SCENARIO 2]\n[STATE 0] I have that, the purple block is clear, the cyan block is clear, the white block is clear, the hand is empty, the white block is on top of the brown block, the purple block is on the table, the cyan block is on the table and the brown block is on the table.\n[ACTION] Pick up the cyan block.\n[CHANGE] The hand was empty and is now holding the cyan block, the cyan block was on the table and is now in the hand, and the cyan block is no longer clear.\n[STATE 1] I have that, the cyan block is in the hand, the white block is clear, the purple block is clear, the hand is holding the cyan block, the white block is on top of the brown block, the purple block is on the table and the brown block is on the table.\n\n[SCENARIO 3]\n[STATE 0] I have that, {}\n[ACTION] {}\n[CHANGE]",
    "world_update_unstack": "I am playing with a set of blocks where I need to arrange the blocks into stacks. Here are the actions I can do \n\nPick up a block \nUnstack a block from on top of another block \nPut down a block \nStack a block on top of another block \n\nI have the following restrictions on my actions:\nI can only pick up or unstack one block at a time. \nI can only pick up or unstack a block if my hand is empty. \nI can only pick up a block if the block is on the table and the block is clear. A block is clear if the block has no other blocks on top of it and if the block is not picked up. \nI can only unstack a block from on top of another block if the block I am unstacking was really on top of the other block. \nI can only unstack a block from on top of another block if the block I am unstacking is clear. Once I pick up or unstack a block, I am holding the block. \nI can only put down a block that I am holding. \nI can only stack a block on top of another block if I am holding the block being stacked. \nI can only stack a block on top of another block if the block onto which I am stacking the block is clear. Once I put down or stack a block, my hand becomes empty.\n\nAfter being given an initial state and an action, give the new state after performing the action.\n\n[SCENARIO 1]\n[STATE 0] I have that, the white block is clear, the cyan block is clear, the brown block is clear, the hand is empty, the white block is on top of the purple block, the purple block is on the table, the cyan block is on the table and the brown block is on the table.\n[ACTION] Unstack the white block from on top of the purple block.\n[CHANGE] The hand was empty and is now holding the white block, the white block was on top of the purple block and is now in the hand, the white block is no longer clear, and the purple block is now clear.\n[STATE 1] I have that, the purple block is clear, the cyan block is clear, the brown block is clear, the hand is holding the white block, the white block is in the hand, the purple block is on the table, the cyan block is on the table and the brown block is on the table.\n\n[SCENARIO 2]\n[STATE 0] I have that, the purple block is clear, the cyan block is clear, the white block is clear, the hand is empty, the cyan block is on top of the brown block, the purple block is on the table, the white block is on the table and the brown block is on the table.\n[ACTION] Unstack the cyan block from on top of the brown block.\n[CHANGE] The hand was empty and is now holding the cyan block, the cyan block was on top of the brown block and is now in the hand, the cyan block is no longer clear, and the brown block is now clear.\n[STATE 1] I have that, the purple block is clear, the brown block is clear, the cyan block is in the hand, the white block is clear, the hand is holding the cyan block, the purple block is on the table, the white block is on the table and the brown block is on the table.\n\n[SCENARIO 3]\n[STATE 0] I have that, {}\n[ACTION] {}\n[CHANGE]",
    "world_update_putdown": "I am playing with a set of blocks where I need to arrange the blocks into stacks. Here are the actions I can do \n\nPick up a block \nUnstack a block from on top of another block \nPut down a block \nStack a block on top of another block \n\nI have the following restrictions on my actions:\nI can only pick up or unstack one block at a time. \nI can only pick up or unstack a block if my hand is empty. \nI can only pick up a block if the block is on the table and the block is clear. A block is clear if the block has no other blocks on top of it and if the block is not picked up. \nI can only unstack a block from on top of another block if the block I am unstacking was really on top of the other block. \nI can only unstack a block from on top of another block if the block I am unstacking is clear. Once I pick up or unstack a block, I am holding the block. \nI can only put down a block that I am holding. \nI can only stack a block on top of another block if I am holding the block being stacked. \nI can only stack a block on top of another block if the block onto which I am stacking the block is clear. Once I put down or stack a block, my hand becomes empty.\n\nAfter being given an initial state and an action, give the new state after performing the action.\n\n[SCENARIO 1]\n[STATE 0] I have that, the white block is clear, the purple block is clear, the cyan block is in the hand, the brown block is clear, the hand is holding the cyan block, the white block is on the table, the purple block is on the table, and the brown block is on the table.\n[ACTION] Put down the cyan block.\n[CHANGE] The hand was holding the cyan block and is now empty, the cyan block was in the hand and is now on the table, and the cyan block is now clear.\n[STATE 1] I have that, the cyan block is clear, the purple block is clear, the white block is clear, the brown block is clear, the hand is empty, the white block is on the table, the purple block is on the table, the cyan block is on the table, and the brown block is on the table.\n\n[SCENARIO 2]\n[STATE 0] I have that, the purple block is clear, the black block is in the hand, the white block is clear, the hand is holding the black block, the white block is on top of the brown block, the purple block is on the table, and the brown block is on the table.\n[ACTION] Put down the black block.\n[CHANGE] The hand was holding the black block and is now empty, the black block was in the hand and is now on the table, and the black block is now clear.\n[STATE 1] I have that, the black block is clear, the purple block is clear, the white block is clear, the hand is empty, the white block is on top of the brown block, the purple block is on the table, the brown block is on the table, and the black block is on the table.\n\n[SCENARIO 3]\n[STATE 0] I have that, {}\n[ACTION] {}\n[CHANGE]",
    "world_update_stack": "I am playing with a set of blocks where I need to arrange the blocks into stacks. Here are the actions I can do \n\nPick up a block \nUnstack a block from on top of another block \nPut down a block \nStack a block on top of another block \n\nI have the following restrictions on my actions:\nI can only pick up or unstack one block at a time. \nI can only pick up or unstack a block if my hand is empty. \nI can only pick up a block if the block is on the table and the block is clear. A block is clear if the block has no other blocks on top of it and if the block is not picked up. \nI can only unstack a block from on top of another block if the block I am unstacking was really on top of the other block. \nI can only unstack a block from on top of another block if the block I am unstacking is clear. Once I pick up or unstack a block, I am holding the block. \nI can only put down a block that I am holding. \nI can only stack a block on top of another block if I am holding the block being stacked. \nI can only stack a block on top of another block if the block onto which I am stacking the block is clear. Once I put down or stack a block, my hand becomes empty.\n\nAfter being given an initial state and an action, give the new state after performing the action.\n\n[SCENARIO 1]\n[STATE 0] I have that, the white block is clear, the purple block is clear, the cyan block is in the hand, the brown block is clear, the hand is holding the cyan block, the white block is on the table, the purple block is on the table, and the brown block is on the table.\n[ACTION] Stack the cyan block on top of the brown block.\n[CHANGE] The hand was holding the cyan block and is now empty, the cyan block was in the hand and is now on top of the brown block, the brown block is no longer clear, and the cyan block is now clear.\n[STATE 1] I have that, the cyan block is clear, the purple block is clear, the white block is clear, the hand is empty, the cyan block is on top of the brown block, the brown block is on the table, the purple block is on the table, and the white block is on the table.\n\n[SCENARIO 2]\n[STATE 0] I have that, the purple block is clear, the black block is in the hand, the white block is clear, the hand is holding the black block, the white block is on top of the brown block, the purple block is on the table, and the brown block is on the table.\n[ACTION] Stack the black block on top of the purple block.\n[CHANGE] The hand was holding the black block and is now empty, the black block was in the hand and is now on top of the purple block, the purple block is no longer clear, and the black block is now clear.\n[STATE 1] I have that, the black block is clear, the white block is clear, the hand is empty, the black block is on top of the purple block, the white block is on top of the brown block, the brown block is on the table, and the purple block is on the table.\n\n[SCENARIO 3]\n[STATE 0] I have that, {}\n[ACTION] {}\n[CHANGE]",
    "self-eval": "I am playing with a set of blocks where I need to arrange the blocks into stacks. Here are the actions I can do\n\nPick up a block\nUnstack a block from on top of another block\nPut down a block\nStack a block on top of another block\n\nI have the following restrictions on my actions:\nI can only pick up or unstack one block at a time.\nI can only pick up or unstack a block if my hand is empty.\nI can only pick up a block if the block is on the table and the block is clear. A block is clear if the block has no other blocks on top of it and if the block is not picked up.\nI can only unstack a block from on top of another block if the block I am unstacking was really on top of the other block.\nI can only unstack a block from on top of another block if the block I am unstacking is clear.\nOnce I pick up or unstack a block, I am holding the block.\nI can only put down a block that I am holding.\nI can only stack a block on top of another block if I am holding the block being stacked.\nI can only stack a block on top of another block if the block onto which I am stacking the block is clear.\nOnce I put down or stack a block, my hand becomes empty.\n\nPlease evaluate whether the given action is a good one under certain conditions.\n\n[STATEMENT]\nAs initial conditions I have that, the red block is clear, the yellow block is clear, the hand is empty, the red block is on top of the blue block, the yellow block is on top of the orange block, the blue block is on the table and the orange block is on the table.\nMy goal is to have that the orange block is on top of the red block.\n[ACTION]\nunstack the red block from on top of the blue block\n[EVALUATION]\nbad\n\n[STATEMENT]\nAs initial conditions I have that, the orange block is in the hand, the yellow block is clear, the hand is holding the orange block, the blue block is on top of the red block, the yellow block is on top of the blue block, and the red block is on the table.\nMy goal is to have have that the yellow block is on top of the orange block.\n[ACTION]\nput down the orange block\n[EVALUATION]\ngood\n\n[STATEMENT]\nAs initial conditions I have that, the orange block is clear, the yellow block is clear, the hand is empty, the blue block is on top of the red block, the orange block is on top of the blue block, the red block is on the table and the yellow block is on the table.\nMy goal is to have that the blue block is on top of the red block and the yellow block is on top of the orange block.\n[ACTION]\npick up the yellow block\n[EVALUATION]\ngood\n\n[STATEMENT]\nAs initial conditions I have that, the orange block is clear, the yellow block is clear, the hand is empty, the blue block is on top of the red block, the orange block is on top of the blue block, the red block is on the table and the yellow block is on the table.\nMy goal is to have that the blue block is on top of the red block and the yellow block is on top of the orange block.\n[ACTION]\npick up the yellow block\n[EVALUATION]\ngood\n\n[STATEMENT]\nAs initial conditions I have that, the blue block is clear, the orange block is in the hand, the red block is clear, the hand is holding the orange block, the red block is on top of the yellow block, the blue block is on the table, and the yellow block is on the table.\nMy goal is to have have that the red block is on top of the yellow block and the orange block is on top of the blue block.\n[ACTION]\nstack the orange block on top of the red block\n[EVALUATION]\nbad\n\n[STATEMENT]\nAs initial conditions I have that, <init_state>\nMy goal is to <goals>\n[ACTION]\n<action>\n[EVALUATION]\n",
    "action_proposals": "I am playing with a set of blocks where I need to arrange the blocks into stacks. Here are the actions I can do \n\nPick up a block \nUnstack a block from on top of another block \nPut down a block \nStack a block on top of another block \n\nI have the following restrictions on my actions:\nI can only pick up or unstack one block at a time. \nI can only pick up or unstack a block if my hand is empty. \nI can only pick up a block if the block is on the table and the block is clear. A block is clear if the block has no other blocks on top of it and if the block is not picked up. \nI can only unstack a block from on top of another block if the block I am unstacking was really on top of the other block. \nI can only unstack a block from on top of another block if the block I am unstacking is clear. Once I pick up or unstack a block, I am holding the block. \nI can only put down a block that I am holding. \nI can only stack a block on top of another block if I am holding the block being stacked. \nI can only stack a block on top of another block if the block onto which I am stacking the block is clear. Once I put down or stack a block, my hand becomes empty.\n\nAfter being given an initial state, propose all possible actions that is valid in the given state.\n\n[SCENARIO 1]\n[STATE] I have that, the white block is clear, the purple block is clear, the cyan block is in the hand, the brown block is clear, the hand is holding the cyan block, the white block is on the table, the purple block is on the table, and the brown block is on the table.\n[ACTION] Stack the cyan block on top of the white block. Stack the cyan block on top of the purple block. Stack the cyan block on top of the brown block. Put down the cyan block\n\n[SCENARIO 2]\n[STATE] I have that, the orange block is clear, the yellow block is clear, the hand is empty, the blue block is on top of the red block, the orange block is on top of the blue block, the red block is on the table and the yellow block is on the table.\n[ACTION] Unstack the orange block from on top of the blue block. Pick up the yellow block. \n\n[SCENARIO 3]\n[STATE] I have that, {}\n[ACTION]",
    "action_proposals_decomp": "I am playing with a set of blocks where I need to arrange the blocks into stacks. Here are the actions I can do \n\nPick up a block \nUnstack a block from on top of another block \nPut down a block \nStack a block on top of another block \n\nI have the following restrictions on my actions:\nI can only pick up or unstack one block at a time. \nI can only pick up or unstack a block if my hand is empty. \nI can only pick up a block if the block is on the table and the block is clear. A block is clear if the block has no other blocks on top of it and if the block is not picked up. \nI can only unstack a block from on top of another block if the block I am unstacking was really on top of the other block. \nI can only unstack a block from on top of another block if the block I am unstacking is clear. Once I pick up or unstack a block, I am holding the block. \nI can only put down a block that I am holding. \nI can only stack a block on top of another block if I am holding the block being stacked. \nI can only stack a block on top of another block if the block onto which I am stacking the block is clear. Once I put down or stack a block, my hand becomes empty.\n\nAfter being given an initial state, proposing all possible actions that is valid.\n\n[SCENARIO 1]\n[STATE] I have that, the white block is clear, the purple block is clear, the cyan block is in the hand, the brown block is clear, the hand is holding the cyan block, the white block is on the table, the purple block is on the table, and the brown block is on the table.\n[REASON] Since cyan block is in the hand, I can only stack the cyan block or put down the cyan block. Since white block is clear, we can stack cyan block on top of the white block. Since the purple block is clear, we can stack the cyan block on top of the purple block. Since the brown block is clear, we can put the cyan block on top of the brown block. \n[ACTION] Stack the cyan block on top of the white block. Stack the cyan block on top of the purple block. Stack the cyan block on top of the brown block. Put down the cyan block\n\n[SCENARIO 2]\n[STATE] I have that, the orange block is clear, the yellow block is clear, the hand is empty, the blue block is on top of the red block, the orange block is on top of the blue block, the red block is on the table and the yellow block is on the table.\n[REASON] Since the hand is empty, I can only unstack a block or pick up a block. Since only the orange block and yellow block is clear, I can only pick up or unstack the yellow block and orange block. Since the yellow block is on the table, I can pick up the yellow block. Since the orange block is on top of the blue block, I can unstack the orange block from on top of the blue block. \n[ACTION] Unstack the orange block from on top of the blue block. Pick up the yellow block. \n\n[SCENARIO 3]\n[STATE] I have that, {}\n[REASON]"
}
\end{lstlisting}

\end{document}